\documentclass[journal]{IEEEtran}
\usepackage{cite}
\ifCLASSINFOpdf
   \usepackage[pdftex]{graphicx}
	 \usepackage{epstopdf}
   \graphicspath{{../pdf/}{../jpeg/}{../eps/}}
   \DeclareGraphicsExtensions{.pdf,.jpeg,.png,.eps}
\else
   \usepackage[dvips]{graphicx}
   \graphicspath{{../eps/}}
   \DeclareGraphicsExtensions{.eps}
\fi
\usepackage[font=scriptsize,skip=0.5pt]{caption} 

\usepackage[cmex10]{amsmath}
\usepackage{amssymb} % Verificar se posso usar esta linha na iEEE
\usepackage{algorithmic}
\usepackage{array}
\usepackage{color}
\usepackage{subfigure} 
\usepackage{fixltx2e}
\usepackage{dblfloatfix}
\usepackage{url}
\usepackage{algorithm}
\usepackage{algorithmic}

\newcommand{\Rn}{\ensuremath{ \mathbb{R}^n }}
\newcommand{\inp}{\ensuremath{ \mathbf{x} }}
\newcommand{\outp}{\ensuremath{ \mathbf{y} }}

\newcommand{\hatalphap}{\ensuremath{ \hat{\alpha}}}
\newcommand{\inpp}{\ensuremath{ x}}
\newcommand{\hatinpp}{\ensuremath{ \hat{x}}}
\newcommand{\hatinp}{\ensuremath{ \hat{\mathbf{x}}}}
\newcommand{\outpp}{\ensuremath{ y}}
\newcommand{\Phip}{\ensuremath{ \Phi}}
\newcommand{\trp}{\ensuremath{ d }}
\newcommand{\tp}{\ensuremath{ y }}
\newcommand{\trset}{\ensuremath{ \mathcal{D} }}
\newcommand{\tset}{\ensuremath{ \mathcal{Y} }}
\newcommand{\sRGC}{\ensuremath{ s }}
\newcommand{\Adiff}{\ensuremath{ A^* }}
\newcommand{\adiff}{\ensuremath{ a^* }}
\newcommand{\Kball}{\ensuremath{ K }}

\newcommand{\numcl}{\ensuremath{ C }}
\newcommand{\icl}{\ensuremath{ k }}
\newcommand{\itcl}{\ensuremath{ l }}
\newcommand{\itmax}{\ensuremath{ L }}
\newcommand{\neig}{\ensuremath{ \mathcal{N} }}

\newcommand{\idecay}{\ensuremath{ I }}

\begin{document}
\title{Geometry-Aware Neighborhood Search for Learning Local Models for Image Reconstruction}

\author{Julio Cesar~Ferreira,~%{Julio Cesar~Ferreira,~\IEEEmembership{Student Member,~IEEE,}
        Elif~Vural,~%,~\IEEEmembership{Fellow,~OSA,}
        and~Christine~Guillemot%,~\IEEEmembership{Life~Fellow,~IEEE}% <-this % stops a space
\thanks{J. C. Ferreira is with the Goiano Federal Institute of Education, Science and Technology, Urutai, 35790-000 Brazil. e-mail: julio.ferreira@ifgoiano.edu.br}% <-this % stops a space
\thanks{E. Vural is with Middle East Technical University, Ankara, 06800 Turkey. e-mail: velif@metu.edu.tr}
\thanks{C. Guillemot is with INRIA, Rennes, 35000 France. e-mail:  christine.guillemot@inria.fr}
\thanks{Most part of the work was performed while the first two authors were in INRIA.}}
\maketitle

\begin{abstract}
Local learning of sparse image models has proven to be very effective to solve inverse problems in many computer vision applications.  To learn such models, the data samples are often clustered using the K-means algorithm with the Euclidean distance as a dissimilarity metric.  However, the Euclidean distance may not always be a good dissimilarity measure for comparing data samples lying on a manifold. In this paper, we propose two algorithms for  determining a local subset of training samples from which a good local model can be computed for reconstructing a given input test sample, where we take into account the underlying geometry of the data. The first algorithm, called Adaptive Geometry-driven Nearest Neighbor search (AGNN), is an adaptive scheme which can be seen as an out-of-sample extension of the replicator graph clustering method for local model learning. The second method, called Geometry-driven Overlapping Clusters (GOC), is a less complex nonadaptive alternative for training subset selection. The proposed AGNN and GOC methods are evaluated in image super-resolution, deblurring and denoising applications and shown to outperform  spectral clustering, soft clustering, and geodesic distance based subset selection in most settings.
\end{abstract}

% \fi
\IEEEpeerreviewmaketitle
%
%
%
%
% Dicas para este paper:
% - colocar no paper o porquê utilizamos patches from LR image instead of HR pathes.
% - colocar duas tabelas no paper: 1 para HF images and another one for usual benchmark.
% - present results for differents scales: 2, 3 and 4 (minimum 2 and 3).
% - make ilustration of the manifolds and another ludic examples an put in the paper.
\section{Introduction}
\label{sec:introduction}

\IEEEPARstart{M}{any} image restoration problems such as superresolution, deblurring, and denoising can be formulated as a linear inverse problem, by modeling the image deformation via a linear system. Such problems are generally ill-posed and the solutions often rely on some a priori information about the image to be reconstructed. Research in the recent years has proven that adopting an appropriate sparse image model can yield quite satisfactory reconstruction qualities. 
Sparse representations are now used to solve inverse problems in many computer vision applications, such as superresolution \cite{Dong13nonlocally}, \cite{Dong11image}, \cite{Yang08image}, \cite{Yang10image}; denoising  \cite{Dong13nonlocally}, \cite{Elad06image}, \cite{Dong11sparsity}; compressive sensing \cite{Donoho06compressed}, \cite{Candes06near}, \cite{Candes06robust}; and deblurring \cite{Dong13nonlocally}, \cite{Dong11image}. While several works assume that the image to be reconstructed has a sparse representation in a large overcomplete dictionary \cite{Yang10image}, \cite{Elad06image}, it has also been observed that representing the data with small, local models (such as subspaces) might have benefits over a single and global model since local models may be more adaptive and capture better the local variations in data characteristics \cite{Dong13nonlocally}, \cite{Dong11image}, \cite{Ni11example}. The image restoration methods in \cite{Dong13nonlocally} and \cite{Dong11image} propose a patch-based processing of images, where the training patches are first clustered and then a principal component analysis (PCA) basis is learned in each cluster. 
The idea of learning adaptive models from groups of similar patches for image restoration has been exploited in several recent works \cite{Salmon14poisson}, \cite{Dabov09bm3d}, \cite{Danielyan10denoising}.

When learning local models, the assessment of the similarity between image patches is of essential importance. Different similarity measures lead to different partitionings of data, which may eventually change the learned models significantly. Many algorithms constructing local models assess similarity based on the Euclidean distance between samples. For example in \cite{Dong13nonlocally} and  \cite{Dong11image} image patches are clustered using the K-means algorithm, where patches having a small Euclidean distance are grouped together to learn a PCA basis. Test patches are then reconstructed under the assumption that they are sparsely representable in this basis. 

\begin{figure*}[!t]
\begin{center}
     \subfigure[]
       {\label{fig:illus_pca_correct}\includegraphics[height=2.5cm]{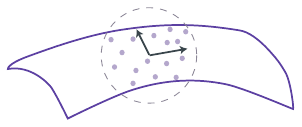}}
   %    \hspace{0.2cm}
     \subfigure[]
       {\label{fig:illus_pca_bend}\includegraphics[height=3cm]{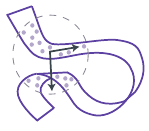}}
   %    \hspace{0.2cm}
     \subfigure[]
       {\label{fig:illus_pca_curve}\includegraphics[height=3.5cm]{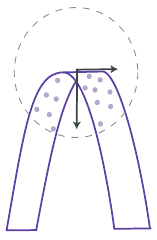}}
 \end{center}
 \caption{PCA basis vectors computed with data sampled from a neighborhood on a manifold. In (a), the two most significant principal directions correspond to tangent directions and PCA computes a local model coherent with the manifold geometry. In (b), PCA fails to recover the tangent space as the manifold bends over itself and the neighborhood size is not selected properly. In (c), as the curvature component is stronger than the tangential components, the subspace spanned by the two most significant PCA basis vectors again fails to approximate the tangent space.}
 \label{fig:illus_pca}
\end{figure*}

However, patches sampled from natural images are highly structured and constitute a low-dimensional subset of the high-dimensional ambient space. In fact, natural image patches are commonly assumed to lie close to a low-dimensional manifold \cite{Lee03the}, \cite{Peyre09manifold}. Similarly, in the deconvolution method proposed in \cite{Ni11example}, image patches are assumed to lie on a large patch manifold, which is decomposed into a collection of locally linear models learned by clustering and computing local PCA bases. The geometric structure of a patch manifold depends very much on the characteristics of the patches constituting it; the manifold is quite nonlinear especially in regions where patches have a rich texture. When evaluating the similarity between patches on a patch manifold, care should be taken especially in high-curvature regions, where Euclidean distance loses its reliability as a dissimilarity measure. 
In other words, in the K-means based setting of \cite{Dong13nonlocally} and  \cite{Dong11image}, one may obtain a good performance only if the local PCA basis agrees with the local geometry of the patch manifold, i.e., the most significant principal directions should correspond to the tangent directions on the patch manifold so that data can be well approximated with a sparse linear combination of only a few basis vectors. While this easily holds in low-curvature regions of the manifold where the manifold is flat, in high-curvature regions, the subspace spanned by the most significant principal vectors computed from the nearest Euclidean-distance neighbors of a reference point may diverge significantly from the tangent space of the manifold if the neighborhood size is not selected properly \cite{Kaslovsky12overcoming}, \cite{Tyagi13tangent}. This is illustrated in Figure \ref{fig:illus_pca}, where the first few significant principal directions fail to approximate the tangent space because the manifold bends over itself as in Figure \ref{fig:illus_pca_bend}, or because the curvature principal components dominate the tangential principal components as in Figure \ref{fig:illus_pca_curve}.

In this work, we focus on image restoration algorithms solving inverse problems based on sparse representations of images in locally learned subspaces, and we present geometry-driven strategies to select subsets of data samples for learning local models. Given a test sample, we address the problem of determining a local subset of the training samples, i.e., a neighborhood of the test sample, from which a good local model can be computed for reconstructing the test sample, where we take into account the underlying geometry of the data. Hence, the idea underlying this work is to compute local models that agree with the low-dimensional intrinsic geometry of data.  Low dimensionality allows sparse representations of data, and the knowledge of sparsity  can be efficiently used  for solving inverse problems in image restoration.

Training subsets for learning local models can be determined in two ways; adaptively or nonadaptively. In adaptive neighborhood selection, a new subset is formed on the fly for each test sample, whereas in nonadaptive neighborhood selection one subset is chosen for each test sample among a collection of training subsets determined beforehand in a learning phase. Adaptive selection has the advantage of flexibility, as the subset formed for a particular test sample fits its characteristics better than a predetermined subset, but the drawback is the higher complexity.
%. In return, the complexity of the reconstruction phase is smaller in nonadaptive subset selection. 
In this work, we study both the adaptive and the nonadaptive settings and propose two different algorithms for geometry-aware local neighborhood selection.

We first present an adaptive scheme, called Adaptive Geometry-driven Nearest Neighbor search (AGNN). Our method is inspired by the Replicator Graph Clustering (RGC) \cite{Donoser13replicator} algorithm and can be regarded as an out-of-sample extension of RGC for local model learning. Given a test sample, the AGNN method computes a diffused affinity measure between each test sample and the training samples in a manner that is coherent with the overall topology of the data graph. The nearest neighbor set is then formed by selecting the training samples that have the highest diffused affinities with the test sample.

The evaluation of the adaptive AGNN method in superresolution experiments 
%suggests that it yields 
shows a quite satisfactory image reconstruction quality. 
%Meanwhile, as AGNN is an adaptive method repeating the neighborhood selection process for each test sample, it may not be optimal for applications where the speed of reconstruction is particularly critical. For this reason, 
We then propose a nonadaptive scheme called Geometry-driven Overlapping Clusters (GOC), which seeks a less complex alternative for training subset selection. The method computes a collection of training subsets in a prior learning phase in the form of overlapping clusters. The overlapping clusters are formed by first initializing the cluster centers and then expanding each cluster around its central sample by following the $K$-nearest neighborhood connections on the data graph. What really determines the performance of the GOC method is the structure of the clusters, driven by the number of neighbors $K$ and the amount of expansion. We propose a geometry-based strategy to set these parameters, by studying the rate of decay of PCA coefficients of data samples in the cluster, thereby characterizing how close the cluster lies to a low-dimensional subspace.

%Note that, in principle, the AGNN method can also benefit from the strategy for setting the neighborhood size proposed in GOC. However, as the subset formation is repeated for each test sample in AGNN, this additional step would increase the complexity of the method and is skipped in our implementation. 

Note that, while the proposed AGNN and GOC algorithms employ similar ideas to those in manifold clustering methods, our study differs from manifold clustering as we do not aim to obtain a partitioning of data.  Instead, given a test sample to be reconstructed, we focus on the selection of a local subset of training data to learn a good local model. We evaluate the performance of our methods in image superresolution, deblurring and denoising applications. The results  show that the proposed similarity assessment strategies can provide performance gains compared to the Euclidean distance, especially for superresolving images with rich texture where patch manifolds are highly nonlinear. When applying the proposed method in the superresolution problem, we select the NCSR algorithm \cite{Dong13nonlocally} as a reference method, which currently leads the state of the art in superresolution. 
%We show that the performance of NCSR is improved when the proposed neighborhood selection methods are used for local model learning instead of the K-means algorithm used in the original method \cite{Dong13nonlocally}.
We first show that the proposed AGNN and GOC methods outperform reference subset selection strategies such as spectral clustering, soft clustering, and geodesic distance based neighborhood selection. Finally, we perform comparative experiments with the NCSR \cite{Dong13nonlocally}, ASDS \cite{Dong11image}, and SPSR \cite{Peleg14a} superresolution algorithms, which suggest that the proposed methods can be successfully applied in superresolution for taking the state of the art one step further. The experiments on image deblurring also confirm these findings, suggesting that the proposed methods perform better than K-means in most images. Meanwhile, we have achieved a marginal performance gain in image denoising applications only at small noise levels.
%The ``averaging out'' of the noise component by using similar noisy examples in a local neighborhood is the key factor that leads to the elimination of the noise, and the geometry-based capture of structural similarities seems to play only a secondary role. 
%Moreover, the geodesic-distance based methods seem to suffer from the deformation of the geometric structure of the patch manifold at high noise levels in the application of denoising.

The rest of the paper is organized as follows. In Section \ref{sec:related_work} we give an overview of manifold-based clustering methods. In Section \ref{sec:problem} we formulate the neighborhood selection problem studied in this paper. In Section \ref{sec:agnn} we discuss the proposed AGNN method. Then in Section \ref{sec:goc} we describe the GOC algorithm. In Section \ref{sec:experiments} we present experimental results, and in Section \ref{sec:conclusion} we conclude.

\section{Clustering on manifolds: related work}
\label{sec:related_work}

As our study has close links with the clustering of low-dimensional data, we now give a brief overview of some clustering methods for data on manifolds. 
The RGC method \cite{Donoser13replicator}, from which the proposed AGNN method has been inspired, first constructs a data graph. An initial affinity matrix is then computed based on the pairwise similarities between data samples. The affinity matrix is iteratively updated such that the affinities between all sample pairs converge to the collective affinities that consider all paths on the data graph. 
Spectral clustering is another well-known algorithm for graph-based clustering  \cite{Shi00normalized}, \cite{Uw01on}. Samples are clustered with respect to a low-dimensional embedding given by the functions of slowest variation on the data graph, which encourages assigning neighboring samples with strong edge weights to the same cluster. The Laplacian eigenmaps method \cite{Belkin03laplacian} builds on the same principle; however, it targets dimensionality reduction. 

Geodesic clustering provides an extension of the $K$-means algorithm to cluster data lying on a manifold, where the Euclidean distance is replaced with the geodesic distance \cite{Asgharbeygi2008}, \cite{Tu2014}. In \cite{Breitenbach2005}, a method is proposed for clustering data lying on a manifold, which extends the graph-based  semi-supervised learning algorithm in \cite{Zhou2004} to a setting with unlabeled data. The diffusion matrix that diffuses known class labels to unlabeled data in \cite{Zhou2004} is interpreted as a diffusion kernel in \cite{Breitenbach2005}, which is then used for determining the similarity between data samples to obtain clusters. The works in \cite{Turaga10nearest}, \cite{Chaudhry10fast} also use the geodesic distance as a dissimilarity measure. They propose methods for embedding the manifold into the tangent spaces of some selected reference points and perform a fast approximate nearest neighbor search on the space of embedding.

While the above algorithms consider all data samples to lie on a single manifold, several other methods model low-dimensional data as samples from multiple manifolds and study the determination of these manifolds. An expectation maximization approach is employed in \cite{Souvenir2005} to partition the data into manifolds. The points on each manifold are then embedded into a lower-dimensional domain. The method in \cite{Elhamifar2011} computes a sparse representation of each data sample in terms of other samples, where high coefficients are encouraged for nearby samples. Once the sparse coefficients are computed, data is grouped into manifolds simply with spectral clustering. The method in \cite{Goh2008} extends several popular nonlinear dimensionality reduction algorithms to the Riemannian setting by replacing the Euclidean distance with the Riemannian distance. It is then shown that, if most data connections lie within the manifolds rather than between them, the proposed Riemannian extensions yield clusters corresponding to different manifolds. 

Finally, the generation of overlapping clusters in GOC is also linked to soft clustering \cite{Bezdek1984}. Rather than strictly partitioning the data into a set of disjoint groups, a membership score is computed between each data sample and each cluster center in soft clustering. The cluster centers are then updated by weighing the samples according to the membership scores. In \cite{Kim07soft}, a manifold extension of soft clustering is proposed, where the membership scores are computed with a geodesic kernel instead of the Euclidean distance.

 
%\hfill September 17, 2014
%\IEEEpubidadjcol
%
%
%
%
\section{Rationale and Problem Formulation}
\label{sec:problem}

In patch-based image processing, one often would like to develop tools that can capture the common structures inherently present in patches and use this information for the efficient treatment of images. One important example is the invariance to geometric transformations. In practical image formation scenarios, different regions of the image are likely to observe the same structure, exposed, however, to different geometric transformations in different parts of the image plane. While most patch-based methods inherently achieve invariance to translations as they extract patches from the image over sliding windows, more complex transformations such as rotations and scale changes are more difficult to handle in evaluating the structural similarities between patches. In addition to geometric transformation models, structural similarities between image patches may be stemming from many other low-dimensional, possibly parametrizable patch models as well. In \cite{Peyre09manifold}, several parametrizable patch manifold models are explored such as oscillating textures and cartoon images. In the treatment or reconstruction of image patches, local models computed from patches sharing the same structure reflect the local geometry of the patch manifold, while the comparison of patch similarities based on Euclidean distance does not necessarily achieve this. In this paper, we propose similarity assessment strategies that better take  structural similarities into account than the simple Euclidean distance in image reconstruction.

Given observed measurements $\outp$, the ill-posed inverse problem can be generally formulated in a Banach space as
\begin{equation}
	\label{eqn:inverse}
	\outp = \Theta \inp+\nu
\end{equation}
where $\Theta$ is a bounded operator, $\inp$ is an unknown data point and $\nu$ is an error term.  In image restoration $\outp$ is the vectorized form of an observed image, $\Theta$ is a degradation matrix, $\inp$ is the vectorized form of the original image, and $\nu$ is an additive noise vector. There are infinitely many possible data points $\inp$ that explain $\outp$; however, image restoration algorithms aim to reconstruct the original image $\inp$ from the given measurements $\outp$, often by using some additional assumptions on $\inp$. 

In image restoration with sparse representations, $\inp$ can be estimated by minimizing the cost function 
\begin{equation}
	\label{eqn:problem}
	\hat \alpha=\underset{\alpha}{\operatorname{arg\,min}} \, \left\{ \left\| \outp -\Theta\Phi\circ\alpha\right\|_{2}^{2} + \lambda \left\| \alpha \right\|_1\right\}
\end{equation}
where $\Phi$ is a  dictionary, $\alpha$ is the sparse representation of $\inp$ in $\Phi$, and $\lambda>0$ is a regularization parameter. It is common to reconstruct images patch by patch and model the patches of $\inp$ as sparsely representable in $\Phi$. Representing the extraction of the $j$-th patch $\inpp_j$ of $\inp$ with a matrix multiplication as $\inpp_j = R_j \inp$, the reconstruction of the overall image $\inp$ can be represented via the operator $\circ$ as shown in \cite{Dong13nonlocally}, \cite{Dong11image}.  If the dictionary $\Phi$ is well-chosen, one can efficiently model the data points $\inp$ using their sparse representations in $\Phi$. Once the sparse coefficient vector $\alpha$ is estimated, one can reconstruct the image $\inp$ as 
%, where $\alpha$ denotes the concatenation of all $\alpha_p$.% and $R_p$.
\begin{equation}
	\label{eqn:reconstruct}
	\hat{\inp} = \Phi\circ \hat\alpha . %=\left( \sum_{p=1}^{N}{R_{p}^{T}R_p} \right)^{-1} \sum_{p=1}^{N}\left({R_{p}^{T}\Phi\alpha_p} \right)
\end{equation}

While a global model is considered in the above problem, several works such as \cite{Dong13nonlocally}, \cite{Dong11image}, \cite{Yu12solving} propose to reconstruct image patches based on sparse representations in local models. In this case, one aims to reconstruct the $j$-th patch $\inpp_j$ of the unknown image $\inp$ from its degraded observation $\tp_j$ by selecting a local model that is suitable for $\tp_j$. The problem in \eqref{eqn:problem} is then reformulated as
\begin{equation}
	\label{eqn:problem_localrec}
	\hatalphap_j=\underset{\alpha_j}{\operatorname{arg\,min}} \, \left\{ \left\|  \outpp_j -\Theta \Phip_j \alpha_j \right\|_{2}^{2} + \lambda \left\| \alpha_j \right\|_1\right\}
\end{equation}
where $\outpp_j$ is the $j$-th patch from the observed image $\outp$, $\Phip_j$ is a local (PCA) basis chosen for the reconstruction of $\outpp_j$, and $\hatalphap_j$ is the coefficient vector. The unknown patch $\inpp_j$ is then reconstructed as $\hatinpp_j= \Phip_j \hatalphap_j$. The optimization problem in \eqref{eqn:problem_localrec} forces the coefficient vector $\hatalphap_j$ to be sparse. Therefore, the accuracy of the reconstructed patch $\hatinpp_j$ in approximating the unknown patch $\inpp_j$ depends on the reliability of the local basis $\Phip_j$, i.e., whether signals are indeed sparsely representable in $\Phip_j$. 

The main idea proposed in this paper is to take into account the manifold structure underlying the data when choosing a neighborhood of training data points to learn a local basis. Our purpose is to develop a dissimilarity measure that is better suited to the local geometry of the data than the Euclidean distance and also to make the neighborhood selection procedure as adaptive as possible to the test samples to be reconstructed.

% We believe that the dictionaries adaptively learned for each test data points $\inp$ using a neighborhood that take account the manifold geometry can overcome the approaches that consider only the Euclidean space.

%The stage of determining the dictionary plays a key role due to the two following  reasons. First, because the choice of the dictionary is already very important to represent the data and second, because the strategy of choice of the dictionary proposed in this paper intrinsically force the sparsity (that correspond to the regularization term $\left\| \alpha \right\|_1$ in Equation \ref{eqn:problem}), together with the central idea in iterative shrinkage algorithm \cite{Daubechies04an}.

Let $	\trset= \left\{\trp_i  \right\}_{i=1}^{m}$ be a set of $m$ training data points $\trp_i \in \mathbb{R}^n$ lying on a manifold $\mathcal{M}$ and let $	\tset =   \left\{\tp_j \right\}_{j=1}^{M}$ be a set of $M$ test data points $\tp_j \in\mathbb{R}^n $. As for the image reconstruction problem in \eqref{eqn:problem_localrec}, each test data point $\tp_j$ corresponds to a degraded image patch, and the training data points in $\trset$ are used to learn the local bases $\Phip_j$. The test samples $\tp_j$ are not expected to lie on the  patch manifold $\mathcal{M}$ formed by the training samples; however, one can assume $\tp_j$ to be close to $\mathcal{M}$ unless the image degradation is very severe.

We then study the following problem. Given an observation $\tp_j \in \tset$ of an unknown image patch $\inpp_j$, we would like to select a subset $S \subset \trset$ of training samples such that the PCA basis $\Phi_j$ computed from $S $ minimizes the reconstruction error $\|  \inpp_j - \hatinpp_j \|$, where  the unknown patch $ \inpp_j$ is reconstructed as  $\hatinpp_j = \Phi_j \hat{\alpha}_j$, and  the sparse coefficient vector is given by
\begin{equation}
	\hat{\alpha}_j=\underset{\alpha_j}{\operatorname{arg\,min}} \, \left\{ \left\|  \tp_j -\Theta \Phi_j \alpha_j \right\|_{2}^{2} + \lambda \left\| \alpha_j \right\|_1\right\}.
\end{equation}
%

% lying in $\mathfrak{M}$ different manifolds $\left\{\mathcal{M}_{\mathfrak{j}}\right\}_{\mathfrak{j}=1}^{\mathfrak{M}}$. 
 %To facilitate the understanding, the matrix with the training data points is presented in Equation \ref{eqn:trainingData}. 		
%\begin{equation}
%	\label{eqn:trainingData}
%	\mathcal{Y}=\left[y_{1}\;\;...\;\;y_{m}\right]
%\end{equation}

%of intrinsic dimensions $\left\{\mathfrak{n}_{\mathfrak{i}}\right\}_{\mathfrak{i}=1}^{\mathfrak{m}}$.
%To facilitate the understanding, the matrix with the test data points is presented in Equation \ref{eqn:testData}. 		
%\begin{equation}
%	\label{eqn:testData}
%	\mathcal{X}=\left[x_{1}\;\;...\;\;x_{M}\right]
%\end{equation}

% of intrinsic dimensions $\left\{\mathfrak{n}_{\mathfrak{j}}\right\}_{\mathfrak{j}=1}^{\mathfrak{M}}$.

Since the nondeformed sample $\inpp_j$ is not known, it is clearly not possible to solve this problem directly. In this work, we propose some constructive solutions to guide the selection of  $S $ by assuming that $\tp_j$ lies close to $\mathcal{M}$. As the manifold $\mathcal{M}$ is not known analytically, we capture the manifold structure of training data $\trset$ by building a similarity graph whose nodes and edges represent the data points and the affinities between them. In Sections \ref{sec:agnn} and \ref{sec:goc} we describe the AGNN and the GOC methods, which respectively propose an adaptive and a nonadaptive solution for training subset selection for local basis learning from the similarity graph.

%In the next section, we propose two strategies to solve the subset selection  problem. The first one (AGNN) adaptively selects the training data points that lie in the same manifold than the test data point diffusing the provided affinities considering the underlying manifold. The second one (GOC) non-adaptively clusters the training data points with respect to underlying manifold structure and overlap cluster.

%We consider the problem of select the more similar data points in $\mathcal{Y}$ to a specific data point $x_j$ according to the underlying manifolds. In order to do this, we build 

%
%
%
%
\section{Adaptive Geometry-Driven Nearest Neighbor Search}
\label{sec:agnn}

%Look for the most representative data points using Euclidean distance has some limitations, as we can see in Figure \ref{fig:illus_pca}. To solve this problem, we propose a strategy that consider the underlying manifold structure.

In this section, we present the Adaptive Geometry-driven Nearest Neighbor search (AGNN) strategy for selecting the nearest neighbors of each test data point within the training data points with respect to an intrinsic manifold structure. Our subset selection method builds on the RGC algorithm \cite{Donoser13replicator}, which targets the clustering of data with respect to the underlying manifold. The RGC method seeks a globally consistent affinity matrix that is the same as its diffused version with respect to the underlying graph topology. However, the RGC method focuses only on the initially available training samples and does not provide a means of handling initially unavailable test samples. We thus present an out-of-sample generalization of RGC and propose a strategy to compute and diffuse the affinities between the test sample and all training samples in a way that is consistent with the data manifold.

\begin{figure}[!t]
	\centering
	\includegraphics[width=2in]{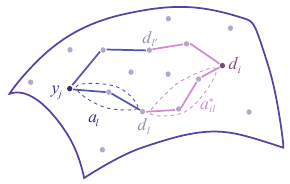}
	\caption{Illustration of AGNN. The affinity between $\tp_j$ and $\trp_l$ is $a_l$, and the affinity between $\trp_l$ and $\trp_i$ is $\adiff_{il}$. The intermediate node $\trp_l$ contributes by the product $a_l \adiff_{il}$ to the overall affinity between $\tp_j$ and $\trp_i$. The sample $\trp_{l'}$ is just another intermediate node like $\trp_l$. Summing the affinities via all possible intermediate nodes (i.e., all training samples), the overall affinity is obtained as in \eqref{eq:agnn_oos_singlep}.}
	\label{fig:illus_agnn}
\end{figure}

In the RGC algorithm, given a set of data points $\trset$, an affinity matrix $A=(a_{il})$ is first computed. The elements $a_{il}$ of $A$ measure the similarity between the data points $\trp_i$ and $\trp_l$. A common similarity measure is the Gaussian kernel 
\begin{equation}
\label{eq:defn_aff_kernel}
a_{il}=\exp{\left(-\frac{\left\| \trp_{i}- \trp_{l}\right\|^2}{n{c_1}^2}\right)}
\end{equation}
where $\left\| \cdot \right\|$ denotes the $\ell_2$-norm on $\mathbb{R}^n$ and $c_1$ is a constant. Then, the initial affinities are updated with respect to the underlying manifold as follows. The affinities are diffused by looking for an $A$ matrix such that each row $A_i$ of $A$ maximizes 
\begin{equation}
\label{eq:RGC}
A_i^T = \arg \max_v (v^T A v).
\end{equation} 
Since the maximization problem on the right hand side of $\eqref{eq:RGC}$ is solved by an eigenvector of $A$, the method seeks an affinity matrix such that the similarities between the data sample $\trp_i$ and all the other samples in $\trset$ (given by the row $A_i$) are proportional to the diffused version of the similarities in $A_i$ over the whole manifold via the product $A A_i^T$; i.e., an affinity matrix is searched such that $A_i^T  \propto A A_i^T$. The optimization problem in \eqref{eq:RGC} is solved with an iterative procedure based on a game theoretical approach to obtain a diffused affinity matrix $\Adiff$. The diffusion of the affinities are constrained to the $\sRGC$ nearest neighbors of each point $\trp_i$.

 %$S\ll m$

In our AGNN method, we first compute and diffuse the affinities of training samples in $\trset$ as proposed in \cite{Donoser13replicator}. This gives us a similarity measure coherent with the global geometry of the manifold. Meanwhile, unlike in RGC, our main purpose is to select a subset $S \subset \trset$ of training samples for a given test sample $\tp_j \in \tset$. We thus need a tool for generalizing the above approach for test samples. 

\begin{algorithm}[t]
\scriptsize
\caption{Adaptive Geometry-driven Nearest Neighbor search (AGNN)}

\begin{algorithmic}[1]

\STATE
\textbf{Input:} \\
$\trset = \left\{\trp_i  \right\}_{i=1}^{m} $: Set of training samples\\
$\tp_j \in \tset $: Test sample \\
$c_1$, $c_2$, $\kappa$: Algorithm parameters

\STATE
\textbf{AGNN Algorithm:}

\STATE
Form affinity matrix $A$ of training samples with respect to \eqref{eq:defn_aff_kernel}.

\STATE
Diffuse the affinities in $A$ to obtain $\Adiff$ as proposed in the RGC method \cite{Donoser13replicator}.

\STATE
Initialize the affinity vector $a$ between test sample $\tp_j$ and the training samples as in \eqref{eq:aff_init_test}.

\STATE
Diffuse the affinities in $a$ to obtain $a^{\star}$ with respect to \eqref{eq:diff_aff_test}. 

\STATE
Determine set $S$  of nearest neighbors of $\tp_j$ by selecting the training samples with the highest affinities as in \eqref{eq:agnn_aff_thres}.

\STATE
\textbf{Output}:\\
$S$ : Set of nearest neighbors of $\tp_j$ in $\trset$. \\

\end{algorithmic}
\label{alg:agnn}
\end{algorithm}
%%%%%%%%%%%%%%%

We propose to compute the affinities between $\tp_j$ and $\trset$ by employing $\Adiff$ as follows. Given a test data point $\tp_j \in \tset$, we first compute an initial affinity vector $a$ whose $i$-th entry
\begin{equation}
\label{eq:aff_init_test}
a_i=\exp\left(-\frac{\left\| \tp_j -\trp_i \right\|^2}{n{c_1}^2}\right)
\end{equation}
measures the similarity between $\tp_j$ and the training sample $\trp_i$. We then update the affinity vector as follows. Denoting the entries of the diffused affinity matrix $\Adiff$ by $\adiff_{il}$, first the product $\adiff_{il} a_l$ should give the component of the overall affinity between $\tp_j$ and $\trp_i$ that is obtained through the sample $\trp_l$: if there is a sample $\trp_l$ that has a high affinity with both $\trp_i$ and $\tp_j$, this means that the affinity between $\trp_i$ and $\tp_j$ should also be high due to the connection established via the intermediate node $\trp_l$ (see the illustration in Figure \ref{fig:illus_agnn}). Note that the formulation in \eqref{eq:RGC} also relies on the same idea. We thus update the affinity vector $a$ such that its $i$-th entry $a_i$ becomes proportional to
\begin{equation}
\label{eq:agnn_oos_singlep}
\sum_{l=1}^m \adiff_{il} \, a_l
\end{equation}
i.e., the total affinity between samples $\trp_i$ and $\tp_j$ obtained through all nodes $\trp_l$ in the training data graph. This suggests that the initial affinities in the vector $a$ should be updated as $\Adiff a$, which corresponds to the diffusion of the affinities on the graph. Repeating this diffusion process $\kappa$ times, we get the diffused affinities of the test sample as
\begin{equation}
\label{eq:diff_aff_test}
a^{\star}=(\Adiff) ^{\kappa}a
\end{equation}
where $a^{\star}_i$ gives the final diffused affinity between $\tp_j$ and $\trp_i$. This generalizes the idea in \eqref{eq:RGC} to initially unavailable data samples; and hence, provides an out-of-sample extension of the diffusion approach in RGC. The parameter $\kappa$ should be chosen in a way to permit a sufficient diffusion of the affinities. However, it should not be too large in order not to diverge too much from the initial affinities in $a$. In our experiments we have observed that $\kappa=2$ gives good results in general.

Once the affinities $a^{\star}$ are computed, the subset $S$ consisting of the nearest neighbors of $\tp_j$ can be obtained as the samples in $\trset$ whose affinities to $\tp_j$ are higher than a threshold
\begin{equation}
\label{eq:agnn_aff_thres}
S=\{ \trp_i \in \trset : a^{\star}_i \geq  c_2  \max_l a^{\star}_l \}
\end{equation}
where $0<c_2< 1$. The samples in $S$ are then used for learning a PCA basis to reconstruct $\tp_j$. The threshold $c_2$ should be chosen sufficiently high to select only the similar patches to the reference patch, however, it should not be selected too high in order to have sufficiently many neighbors necessary for computing a basis. If $S$ contains too few samples, the threshold $c_2$ can be adapted to increase the number of samples or a sufficient number of points with highest affinities can be directly included in $S$. The proposed AGNN method for determining training subsets gets around the problem depicted in Figure \ref{fig:illus_pca_bend}, since points lying at different sides of a manifold twisting onto itself have a small diffused affinity and are not included in the same subset. A summary of the proposed AGNN method is given in Algorithm \ref{alg:agnn}.

%Notice from \eqref{eq:RGC} that the original RGC algorithm looks for an affinity matrix whose rows are its eigenvalues as well. Therefore, when $x$ is taken as a training sample $\trp_i \in \trset$, the affinity vector $a$ is given by the $i$-th row $A_i$ of  diffusion formulated in \eqref{eq:diff_aff_test} provides

%After that, the affinity matrix $A$ computed using the training data points $\trset$ is applied to diffuse the similarities between ${x}$ and $\trset$ with respect to the learned structure on $\trset$, ${\kappa}$ times. To compute the new affinities $a^{\star}$, the equation $a^{\star}=A^{\kappa}a$ is employed, where $a$ is the original affinities and $a^{\star}$ is the diffused affinities between ${x}$ and $\trset$.

%Given a test data point ${x}\notin \trset$,

%After sorting the affinities in $a^{\star}$, we select the number of training data points $K$ solving the equation $K=\max{\left[K^{-}\;K^{\circ}\right]}$, where $K^{-}$ is a minimum value, $K^{\circ}=\max\left\{i\mid a^{\star}_i<c_2\max{a^{\star}}\right\}$, and $c_2$ is a constant. The indices of the selected training data points are given by $\left\{\mathrm{S}\right\}_{i=1}^{K}=\left\{i\mid a^{\star}_i\leq a^{\star}_K\right\}$.

%Finally, PCA bases $\Phi_p$ can be learned using $K$ nearest training data points $\left\{\mathrm{S}\right\}_{i=1}^{K}$ of each test data point ${x}$.% found with respect to $a^{\star}$. %The bases $\Phi$ with size $n\times n$ that corresponds with the $n$ highest eigenvalues are organized like that $\Phi=\left[\phi_1\;\;...\;\;\phi_n\right]^T$.
%
%
%
%
\section{Geometry-Driven Overlapping Clusters}
\label{sec:goc}

As we will see in Section \ref{sec:experiments}, the AGNN method presented in Section \ref{sec:agnn} is efficient in terms of image reconstruction performance. However, it may have a high computational complexity and considerable memory requirements in settings with a large training set $\trset$, as the size of the affinity matrix grows quadratically with the number of training samples and the subset selection is adaptive (repeated for each test sample). For this reason, we propose in this section the Geometry-driven Overlapping Clusters (GOC) method, which provides a computationally less complex solution for obtaining the nearest neighbors of test samples.

The GOC algorithm computes a collection $\{S_\icl \}_{\icl=1}^\numcl$ of subsets  $S_\icl \subset \trset$ of the training data set, which are to be used in local basis computation. Contrary to the AGNN method, the subsets $S_\icl \subset \trset$ are determined only using the training data and are not adapted to the test samples. However, the number  $\numcl$ of subsets should then be sufficiently large to have the desired adaptivity for capturing arbitrary local variations. Due to the large number of subsets, $S_\icl$ are not disjoint in general; hence, can be regarded as overlapping clusters. In the following, we first describe our method for forming the clusters and then propose a strategy to select some parameters that determine the size and the structure of the clusters. 

%In this section, we present , which we propose to select the nearest neighbors training data points for each implicit cluster $k$ with respect to overlap cluster using intrinsic manifold structure. We named ``implicit cluster'' because we do not explain the data points in each cluster $k$. We also propose a strategy to set adaptively the parameters ball size $b$ and the number of levels $l$. 

Given the number of clusters $\numcl$ to be formed, we first determine the central data point $\mu_\icl \in \trset$ of each cluster $S_\icl$. In our implementation, we achieve this by first clustering $\trset$ with the K-means algorithm, and then choosing each $\mu_\icl$ as the point in $\trset$ that has the smallest Euclidean distance to the center of the $\icl$-th cluster given by K-means.

%In order to set the starting training data points, the $k$ central data points $\mu$ among the training data points $\trset\in \mathbb{R}^n$ are selected using the traditional K-means algorithm, where $k$ is the number of clusters. After that, the $y_k\in\trset$ closest to the central points $\mu_k$ are selected using Euclidean distance. 

\begin{figure}[!t]
	\centering
	\includegraphics[width=2in]{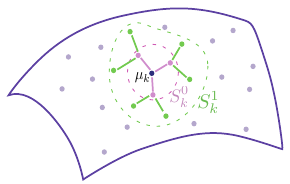}
	\caption{Illustration of the GOC algorithm. The cluster $S_\icl$  around the central sample $\mu_\icl$ is formed gradually. $S_\icl$ is initialized with $S_\icl^0$ containing the $\Kball$ nearest neighbors of $\mu_\icl$ ($\Kball=3$ in the illustration). Then in each iteration $\itcl$, $S_\icl^{\itcl}$ is expanded by adding the nearest neighbors of recently added samples.}
	\label{fig:illus_goc}
\end{figure}

The training data points $ \mu_\icl$ are used as the kickoff for the formation of the clusters $S_\icl$. Given the central sample $\mu_\icl$, the cluster $S_\icl$ is formed iteratively with the GOC algorithm illustrated in Figure \ref{fig:illus_goc} as follows. We first initialize $S_\icl$ as
\begin{equation}
\label{eq:clus_init_goc}
S_\icl^0 = \neig_\Kball (\mu_\icl)
\end{equation}
where $\neig_\Kball (\mu_\icl)$ denotes the set of the $\Kball$-nearest neighbors of $\mu_\icl$ in $\trset$ with respect to the Euclidean distance. Then in each iteration $\itcl$, we update the cluster $S_\icl^\itcl$ as
\begin{equation}
S_\icl^\itcl =S_\icl^{\itcl-1} \cup  \bigcup_{\trp_i \in S_\icl^{\itcl-1} } \neig_\Kball (\trp_i)  
\end{equation}
by including all samples in the previous iteration as well as their $\Kball$-nearest neighbors. Hence, the clusters are gradually expanded by following the nearest neighborhood connections on the data graph. This procedure is repeated for $\itmax$ iterations so that the final set of clusters is given by
\begin{equation}
\label{eq:defn_clust_final}
\{ S_\icl \}_{\icl=1}^\numcl = \{ S_\icl^\itmax \}_{\icl=1}^\numcl .
\end{equation}
The expansion of the clusters is in a similar spirit to the affinity diffusion principle of AGNN; however, is computationally much less complex.

%for each starting training data point $y_k$, we find all training data points $y$ within a b-NN initial ball with ball size $b$, i.e. we find the $b$ training data points that has the smallest Euclidean distance $d\left(y-y_k\right)$. Next, for each training data point in the current ball, we find again a new b-NN ball. We do this $l$ times, always using the same ball size $b$ for each starting data point $y_k$.

%After achieve a predefined number of levels $l$, we merge the indices $\left\{\mathrm{S}\right\}_{\left(l=1,\;i=1\right)}^{\left(L,\;K\right)}$ for $l$ balls obtained the final indices $\left\{\mathrm{S}\right\}_{i=1}^{K}$. The indices $\left\{\mathrm{S}\right\}_{i=1}^{K}$ and centroids $y_k$ are stored for each cluster $k$.

%Finally, PCA bases $\Phi_k$ can be learned using $K$ nearest training data points $\left\{\mathrm{S}\right\}_{i=1}^{K}$ for each implicit cluster ${k}$.

In the simple strategy presented in this section, we have two important parameters to set, which essentially influence the performance of learning: the number of iterations $\itmax$ and the number of samples $\Kball$ in each small neighborhood. In the following, we propose an algorithm to adaptively set these parameters based on the local geometry of data. Our method is based on the observation that the samples in each cluster will eventually be used to learn a local subspace that provides an approximation of the local tangent space of the manifold. Therefore,  $S_\icl$ should lie close to a low-dimensional subspace in $\Rn$, so that nearby test samples  can be assumed to have a sparse representation in the basis $\Phi_\icl$ computed from $S_\icl$. We characterize the concentration of the samples in $S_\icl$ around  a low-dimensional subspace by the decay of the coefficients of the samples in the local PCA basis.

%the local basis $\Phi_\icl$ computed in each cluster $S_\icl$ should approximate the local tangent space. Unfortunately, the local tangent space of the manifold is not known in advance. Nevertheless, one can optimize the parameters $\Kball$ and $\itmax$ such that 

We omit the cluster index $\icl$ for a moment to simplify the notation and consider the formation of a certain cluster $S=S_\icl$. With a slight abuse of notation, let  $S^{\itmax,\Kball} $ stand for the cluster $S$ that is computed by the algorithm described above with parameters $\itmax$ and $\Kball$. Let $\Phi=\left[\phi_1\;\;...\;\;\phi_n\right]$ be the PCA basis computed with the samples in $S$, where  the principal vectors $\phi_1, \dots, \phi_n \in \Rn$ are sorted with respect to the decreasing order of the absolute values of their corresponding eigenvalues. For a training point $\trp_i \in S$, let $\overline \trp_i = \trp_i - \eta_S $ denote the shifted version of $\trp_i$ where $\eta_S=|S|^{-1} \sum_{\trp_i \in S} \trp_i$ is the centroid of cluster $S$. We define
\begin{equation}
	\label{eqn:funclb}
	\begin{split}	
	\idecay(\itmax, \Kball)
	 =\min \big \{  \iota\ |\
	 &\sum_{q=1}^{\iota}
	 {\sum_{\trp_i \in S^{\itmax,\Kball}  }{\left\langle \phi_q,  \overline \trp_i \right\rangle^2}} \\
	 &\geq
	  c_3\sum_{q=1}^{n}
	  {\sum_{\trp_i \in S^{\itmax,\Kball} } {\left\langle \phi_q , \overline \trp_i \right\rangle^2}} 
	  \big \}
	  \end{split}
\end{equation}
%
%\begin{equation}
%	\label{eqn:funclb}	
%	\idecay(\itmax, \Kball)
%	 =\min_{\iota} \big \{  
%	 \sum_{q=1}^{\iota}
%	 {\sum_{\trp_i \in S^{\itmax,\Kball}  }{\left\langle \phi_q,  \overline \trp_i \right\rangle^2}} 
%	 \geq
%	  c_3\sum_{q=1}^{n}
%	  {\sum_{\trp_i \in S^{\itmax,\Kball} } {\left\langle \phi_q , \overline \trp_i \right\rangle^2}} 
%	  \big \}	
%\end{equation}
%
which gives the smallest number of principal vectors to generate a subspace that captures a given proportion $c_3$ of the total energy of the samples in $S$, where $0<c_3<1$. We propose to set the parameters $\itmax$, $\Kball $ by minimizing the function $\idecay(\itmax, \Kball)$, which gives a measure of the concentration of the energy of $S$ around a low-dimensional subspace. However, in the case that $S$ contains $m \leq n$ samples where $n$ is the dimension of the ambient space, the subspace spanned by the first $m-1$ principal vectors always captures all of the energy in $S$; therefore $\idecay(\itmax, \Kball)$ takes a relatively small value; i.e., $\idecay(\itmax, \Kball) \leq m-1$.  In order not to bias the algorithm towards reducing the size of the clusters as a result of this, a normalization of the function $\idecay(\itmax, \Kball)$ is required. We define 
%
%\begin{equation} 
%\label{eq:defn_decayfun_goc}
%\tilde \idecay(\itmax, \Kball) =  \frac{ \idecay(\itmax, \Kball) }
%{\min \{ \, |S^{\itmax,\Kball}| -1, \, n \}}
%\end{equation} 
%
\begin{equation} 
\label{eq:defn_decayfun_goc}
\tilde \idecay(\itmax, \Kball) =   \idecay(\itmax, \Kball)  / 
{\min \{ \, |S^{\itmax,\Kball}| -1, \, n \}}
\end{equation} 
where $| \cdot |$ denotes the cardinality of a set. The denominator $\min \{ \, |S^{\itmax,\Kball}| -1, \, n \}$ of the above expression gives the maximum possible value of $\idecay(\itmax, \Kball)$ in cluster $S^{\itmax,\Kball}$. Hence, the normalization of the coefficient decay function by its maximum value prevents the bias towards small clusters.

\begin{algorithm}[t]
\scriptsize
\caption{Geometry-driven Overlapping Clusters (GOC)}

\begin{algorithmic}[1]

\STATE
\textbf{Input:} \\
$\trset = \left\{\trp_i  \right\}_{i=1}^{m} $: Set of training samples\\
$\numcl$: Number of clusters \\
$c_3$: Algorithm parameter

\STATE
\textbf{GOC Algorithm:}

\STATE
Determine cluster centers $\mu_\icl$ of all $\numcl$ clusters (possibly with the K-means algorithm). 

\FOR{$\icl=1, \cdots, \numcl$}

\STATE
Fix parameter $\itmax'=\itmax_0$ at an initial value $\itmax_0$. 

\FOR{$\Kball'=1, \cdots, \Kball_{max}$}

\STATE
Form cluster $S_\icl = S^{\itmax_0,\Kball'} $ as described in \eqref{eq:clus_init_goc}-\eqref{eq:defn_clust_final}.

\STATE
Evaluate decay rate function $\tilde \idecay(\itmax_0, \Kball') $ given in \eqref{eq:defn_decayfun_goc}. 

\ENDFOR

\STATE
Set $\Kball$ as the $\Kball'$ value that minimizes  $\tilde \idecay(\itmax_0, \Kball') $.

\FOR{$\itmax'=1, \cdots, \itmax_{max}$}

\STATE
Form cluster $S_\icl = S^{\itmax',\Kball} $ as described in \eqref{eq:clus_init_goc}-\eqref{eq:defn_clust_final}.

\STATE
Evaluate decay rate function $\tilde \idecay(\itmax', \Kball) $ given by \eqref{eq:defn_decayfun_goc}. 

\ENDFOR

\STATE
Set $\itmax$ as the $\itmax'$ value that minimizes  $\tilde \idecay(\itmax', \Kball) $.

\STATE
Determine cluster $S_\icl$ as $ S^{\itmax,\Kball}$ with the optimized parameters.

\ENDFOR

\STATE
\textbf{Output}:\\
$\{ S_\icl \}_{\icl=1}^{\numcl}$ : Set of overlapping clusters in $\trset$. \\

\end{algorithmic}
\label{alg:goc}
\end{algorithm}
%%%%%%%%%%%%%%%

We can finally formulate the selection of $\itmax$, $\Kball $  as
%
%\begin{equation}
%	\label{eqn:lb}
% (\itmax, \Kball) = \underset{(\itmax', \Kball') \in \Lambda}{\operatorname{arg\,min}}
%  \,\tilde \idecay(\itmax', \Kball')
%\end{equation}
%
$
 (\itmax, \Kball) = \arg\min_{(\itmax', \Kball') \in \Lambda}
  \,\tilde \idecay(\itmax', \Kball')
$
where $\Lambda$ is a bounded parameter domain. This optimization problem is not easy to solve exactly. One can possibly evaluate the values of $\tilde \idecay(\itmax, \Kball)$ on a two-dimensional grid in the parameter domain. However, in order to reduce the computation cost, we approximately minimize the objective by optimizing one of the parameters and fixing the other in each iteration. We first fix the number of iterations $\itmax$ at an initial value and optimize the number of neighbors $\Kball$. Then, updating and fixing $\Kball$, we optimize $\itmax$. 

The computation of the parameters $\itmax$ and $\Kball$ with the above procedure determines the clusters as in \eqref{eq:defn_clust_final}. The samples in each cluster $ S_\icl $ are then used for computing a local basis $\Phi_\icl$. The proposed GOC method is summarized in Algorithm \ref{alg:goc}. Since the proposed GOC method determines the clusters not only with respect to the connectivity of the data samples on the graph, but also by adjusting the size of the clusters with respect to the local geometry, it provides a solution for both of the problems described in Figures \ref{fig:illus_pca_bend} and \ref{fig:illus_pca_curve}.

In the proposed GOC method, contrary to AGNN, we need to define a strategy to select the PCA basis that best fits a given test patch. Given a test patch $\tp_j$, we propose to select a basis $\Phi_\icl$ by taking into account the distance between $\tp_j$ and the centroid $\mu_\icl$ of the cluster $S_\icl$ (corresponding to $\Phi_\icl$), as well as the agreement between $\tp_j$ and the principal directions in $\Phi_\icl$. Let $\Phi_\icl^r = [\phi_1 \dots \phi_r]$ denote the submatrix of $\Phi_\icl$ consisting of the first $r$ principal vectors, which give the directions that determine the main orientation of the cluster. We then choose the basis $\Phi_\icl$ that minimizes 
\begin{equation}
	\label{eqn:selectD}
	\icl=\underset{\icl'}{\operatorname{arg\,min}} \, \left\{ \left\| \tp_j -\mu_{\icl'}  \right\|_2-\gamma \left\| (\Phi_\icl^r )^T \frac{ \tp_j-\mu_{\icl'}}{\left\| \tp_j -  \mu_{\icl'} \right\|_2}\right\|_2\right\}
\end{equation}
where $\gamma>0$ is a weight parameter. While the first term above minimizes the distance to the centroid of the cluster, the second term maximizes the correlation between the relative patch position $\tp_j-\mu_{\icl'}$  and the most significant principal directions. Once the basis index $\icl$ is determined as above, the test patch $\tp_j$ is reconstructed based on a sparse representation in $\Phi_\icl$.

\section{Experiments}
\label{sec:experiments}

We verify the performance of our proposed methods with extensive experiments  
on image restoration based on sparse representations. In Section \ref{ssec:exp_struc_sim} we first present an experiment where we evaluate the performance of the proposed neighborhood selection strategies in capturing the structural similarities of images.  Then in Sections \ref{ssec:exp_superres}, \ref{ssec:exp_deblur}, and \ref{ssec:exp_denoise} we test our algorithms respectively in superresolution, deblurring, and denoising applications.

\subsection{Transformation-invariant patch similarity analysis}
\label{ssec:exp_struc_sim}

\begin{figure}[!t]
\centering
\includegraphics[width=3.5in]{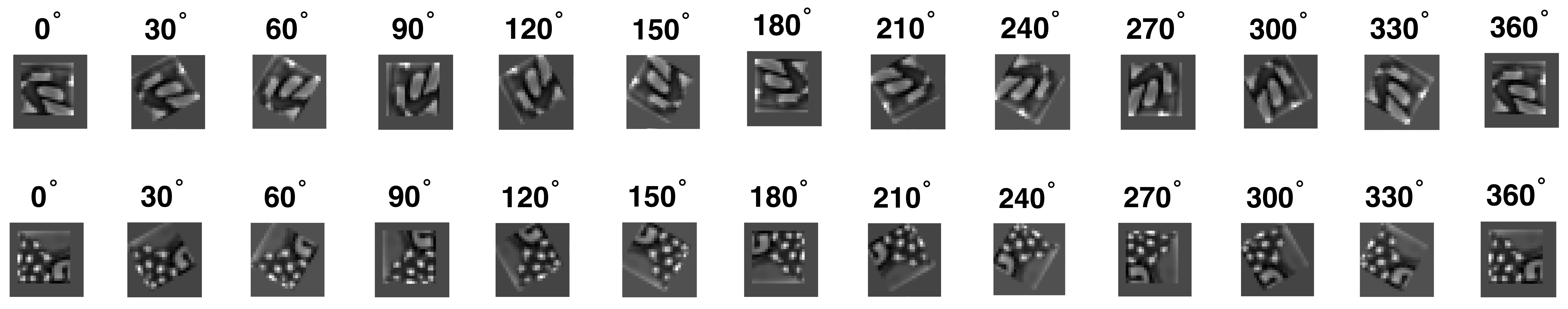}
\caption{Two of the reference patches and their rotated versions used in the experiment}
\label{fig:illus_patch_strsim}
\end{figure}

Natural images often contain different observations of the same structure in different regions of the image. Patches that share a common structure may be generated from the same reference pattern with respect to a transformation model that can possibly be parameterized with a few parameters. One example to parametrizable transformation models is geometric transformations. In this section, we evaluate the performance of the proposed AGNN strategy in capturing structural similarities between image patches in a transformation-invariant way. We generate a collection of patches of size $10 \times 10 $ pixels, by taking a small set of reference patches and applying  geometric transformations consisting of a rotation with different angles to each reference patch to obtain a set of geometrically transformed versions of it. Figure \ref{fig:illus_patch_strsim} shows two reference patches and some of their rotated versions. The data set used in the experiment is generated from 10 reference patches, which are rotated at intervals of 5 degrees.

\begin{figure}[!t]
\centering
\includegraphics[width=2in]{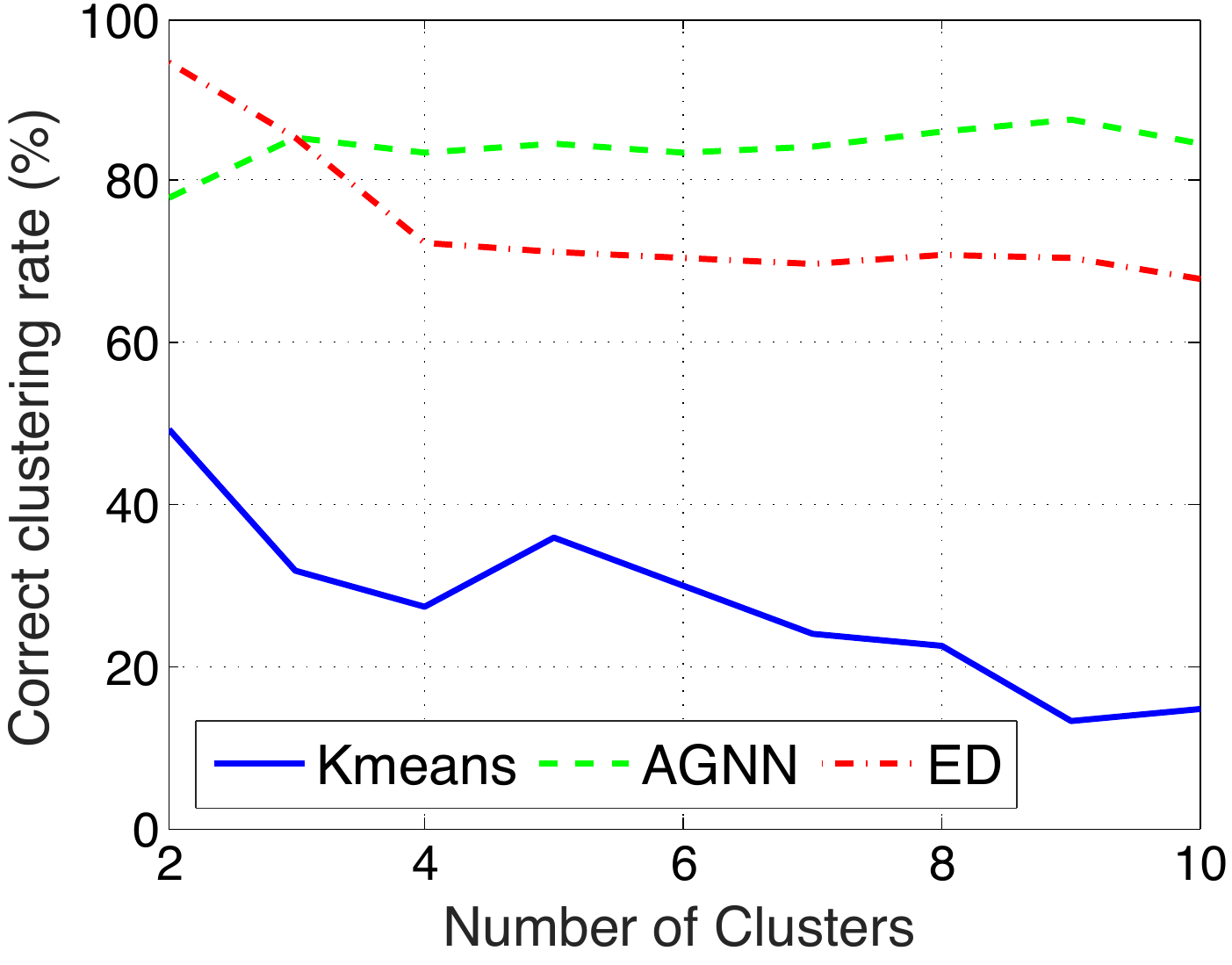}
\caption{Percentage of patches correctly included in the clusters}
\label{fig:corr_cluster_strsim}
\end{figure}

In order to evaluate the performance of transformation-invariant similarity assessment, we look for the nearest neighbors of each patch in the whole collection and identify the ``correct'' neighbors as the ones sharing the same structure, i.e., the patches generated from the same reference patch. Three nearest neighbor selection strategies are tested in the experiment, which are AGNN, neighbor selection with respect to Euclidean distance, and K-means clustering. In AGNN, the neighborhood size that gives the best algorithm performance is used. The Euclidean distance uses the same neighborhood size as AGNN, and the number of clusters in K-means is set as the true number of clusters, i.e., the number of reference patches generating the data set.  The correct clustering rates are shown in Figure \ref{fig:corr_cluster_strsim}, which are the percentage of patches that are correctly present in a cluster (each neighborhood is considered as a cluster in AGNN and Euclidean distance). The horizontal axis shows the number of clusters (i.e., number of reference patches) used in different repetitions of the experiment. It can be observed that the AGNN method yields the best transformation-invariant similarity assessment performance. Contrary to methods based on simple Euclidean distance, AGNN measures the similarity of two patches by tracing all paths on the manifold joining them. Therefore, it is capable of following the gradual transformations of structures on the patch manifold and thus identifying structural similarities of patches in a transformation-invariant manner.

\subsection{Image superresolution}
\label{ssec:exp_superres}

In this section, we demonstrate the benefits of our neighborhood selection strategies in the context of the NCSR algorithm \cite{Dong13nonlocally}, which leads to state-of-the-art performance in image superresolution.

The NCSR algorithm \cite{Dong13nonlocally} is an image restoration method that reconstructs image patches by selecting a model among a set of local PCA bases. This strategy exploits the image nonlocal self-similarity to obtain estimates of the sparse coding coefficients of the observed image. The method first clusters training patches with the K-means algorithm and then adopts the adaptive sparse domain selection strategy proposed in \cite{Dong11image} to learn a local PCA basis for each cluster from the estimated high-resolution (HR) images. After the patches are coded, the NCSR objective function is optimized with the Iterative Shrinkage Thresholding (IST) algorithm proposed in \cite{Daubechies04an}. The clustering of training patches with the K-means algorithm in \cite{Dong13nonlocally} is based on adopting the Euclidean distance as a dissimilarity measure. The purpose of our experiments is then to show that the proposed geometry-based nearest neighbor selection methods can be used for improving the performance of an image reconstruction algorithm such as NCSR.

%The method is based on learning local PCA bases with a set of training patches and reconstructing test patches with the local bases, where the training patches are initialized as the test patches in the beginning of the algorithm and then gradually updated as test patches are superresolved. For more details about the NCSR algorithm, please see \cite{Dong13nonlocally}.

\begin{figure}[!t]
\begin{center}
\setcounter{subfigure}{0}   
\subfigure{\includegraphics[width=0.18\linewidth]{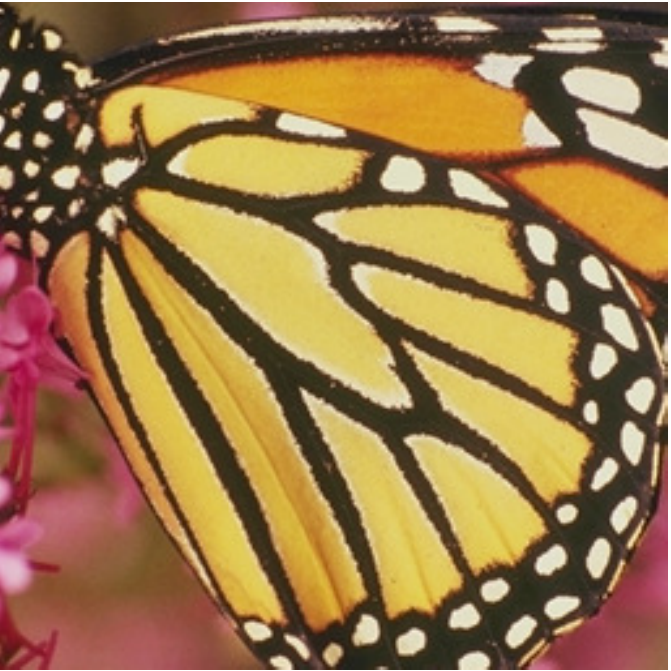}}   \subfigure{\includegraphics[width=0.18\linewidth]{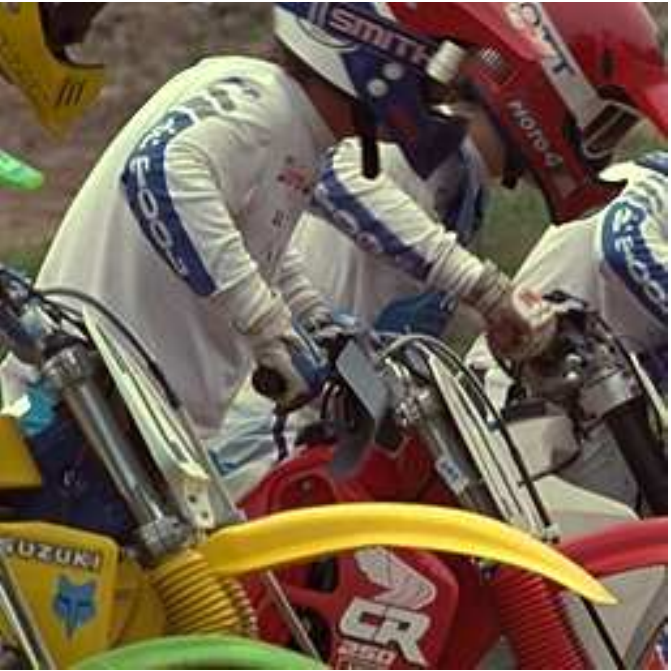}}
   \subfigure{\includegraphics[width=0.18\linewidth]{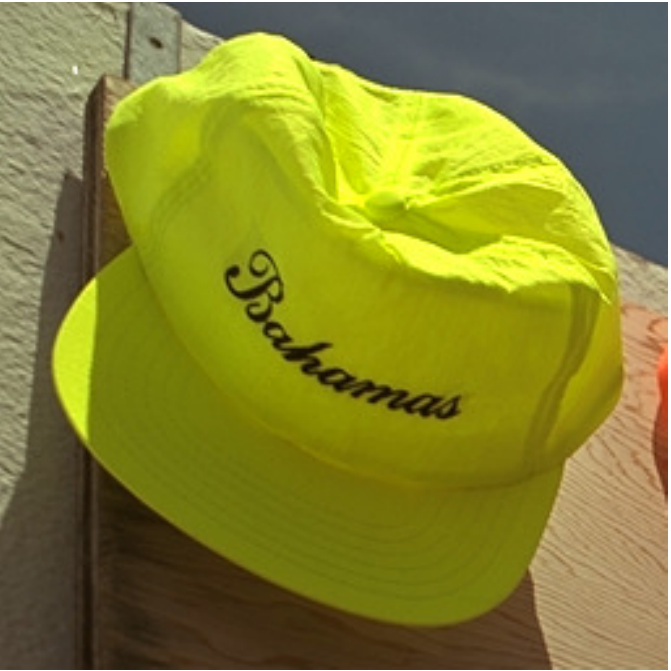}}
   \subfigure{\includegraphics[width=0.18\linewidth]{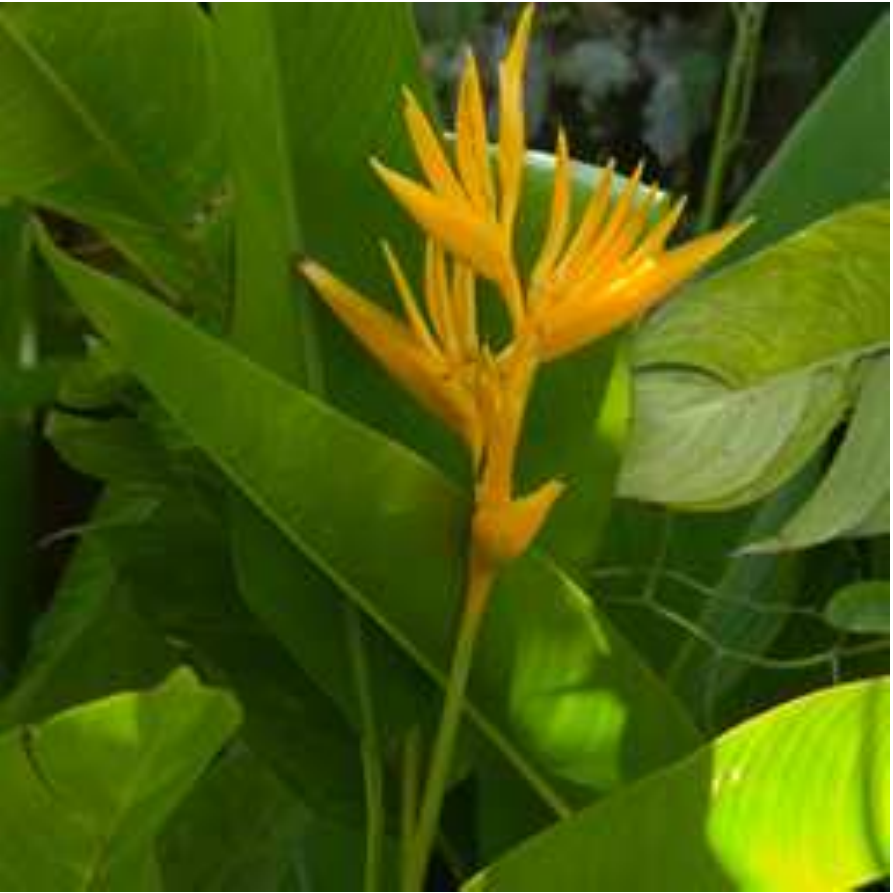}}
   \subfigure{\includegraphics[width=0.18\linewidth]{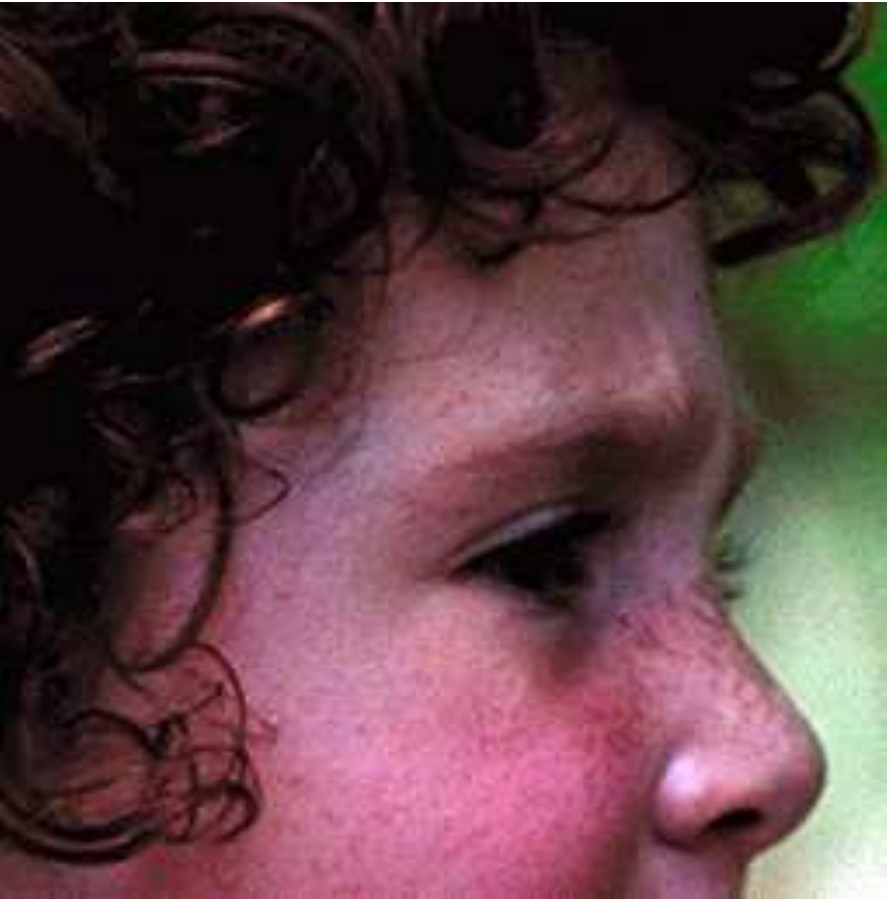}} \\ 
\setcounter{subfigure}{0}   
\subfigure{\includegraphics[width=0.18\linewidth]{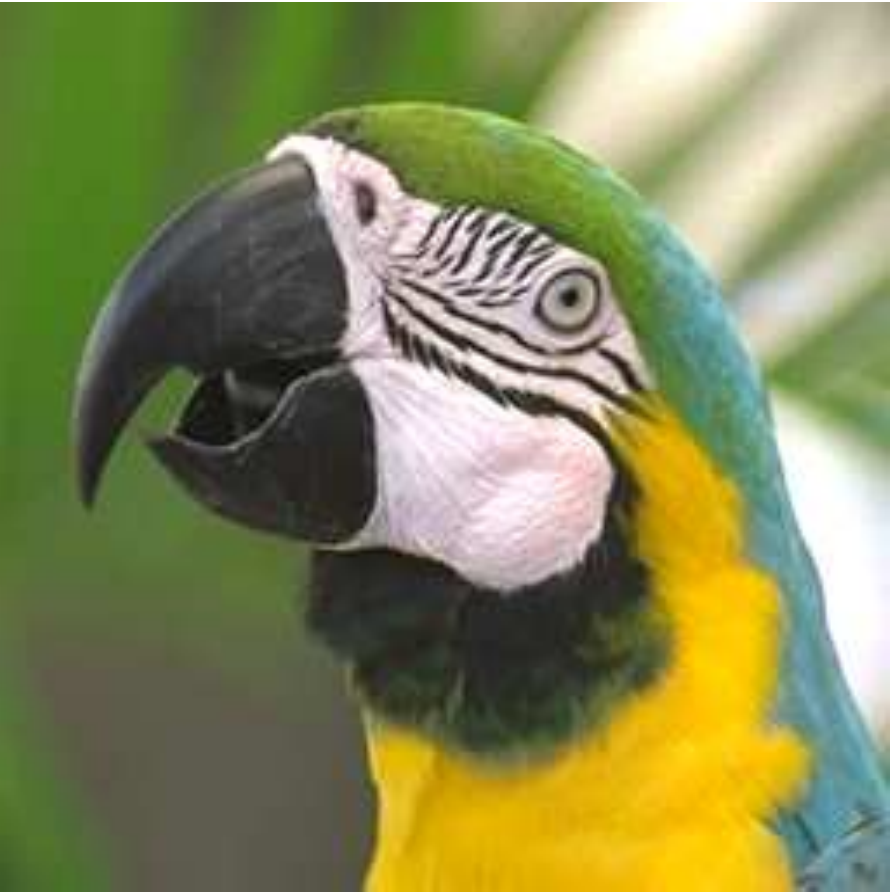}}   \subfigure{\includegraphics[width=0.18\linewidth]{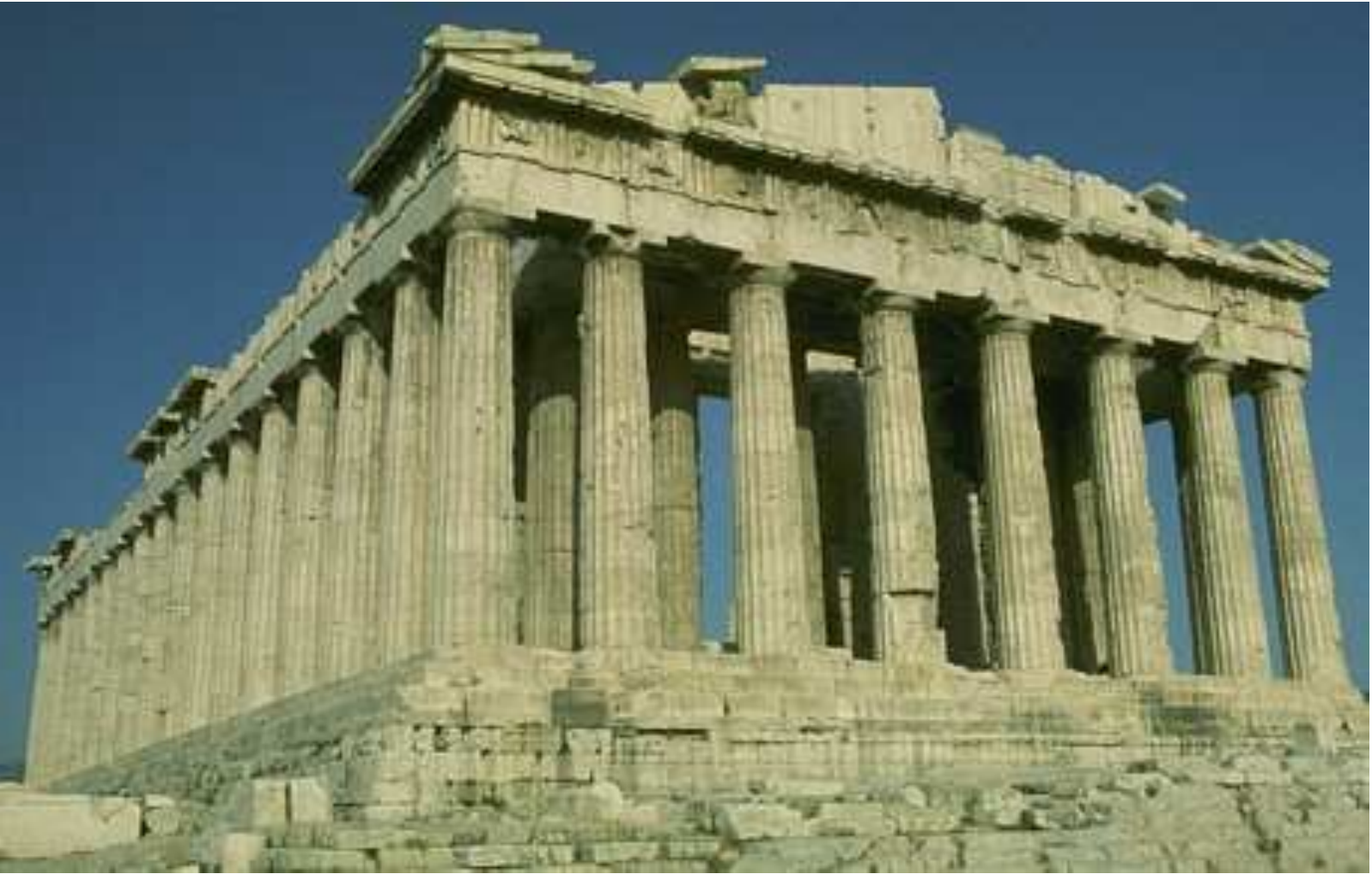}}
   \subfigure{\includegraphics[width=0.18\linewidth]{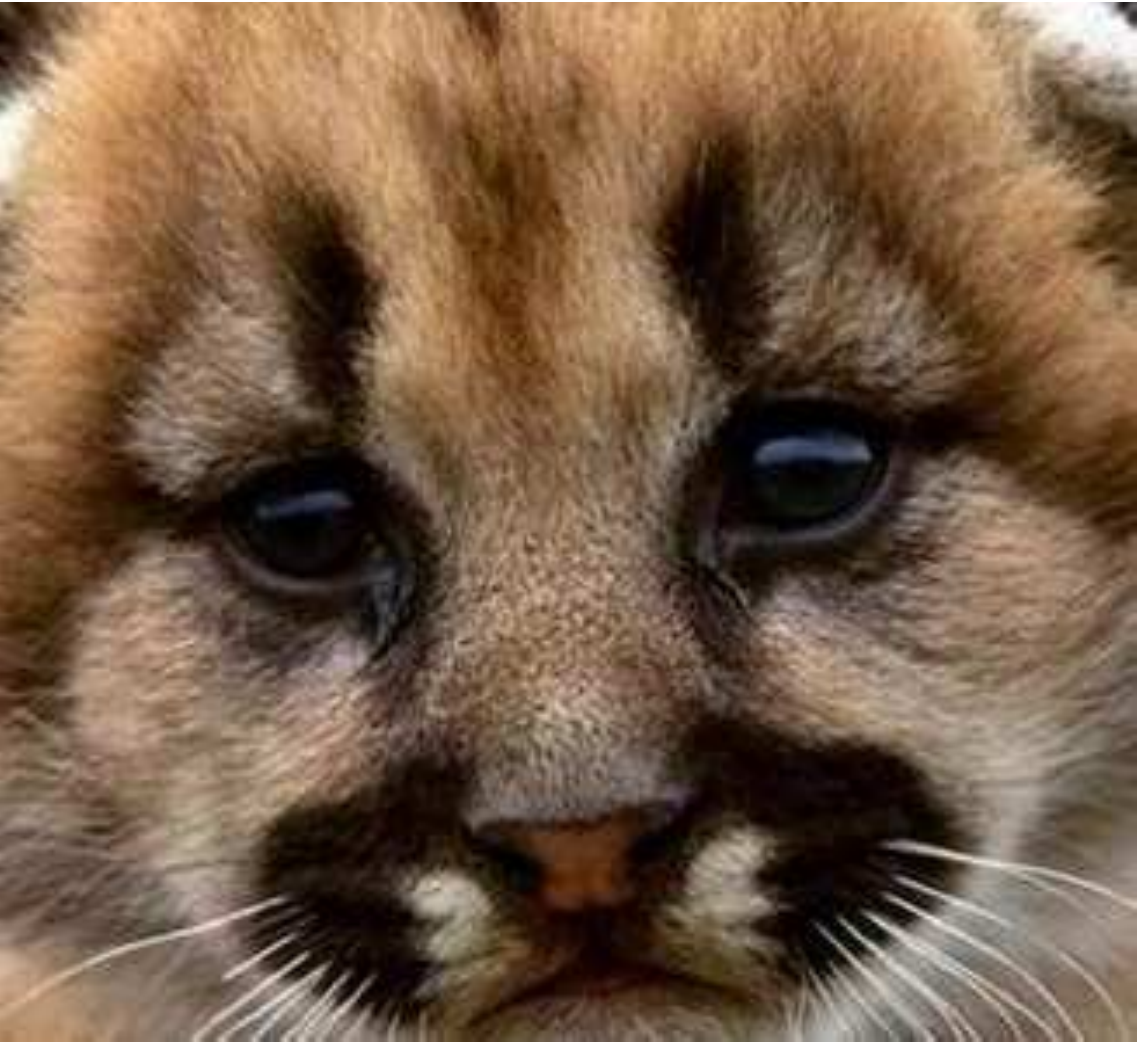}}
   \subfigure{\includegraphics[width=0.18\linewidth]{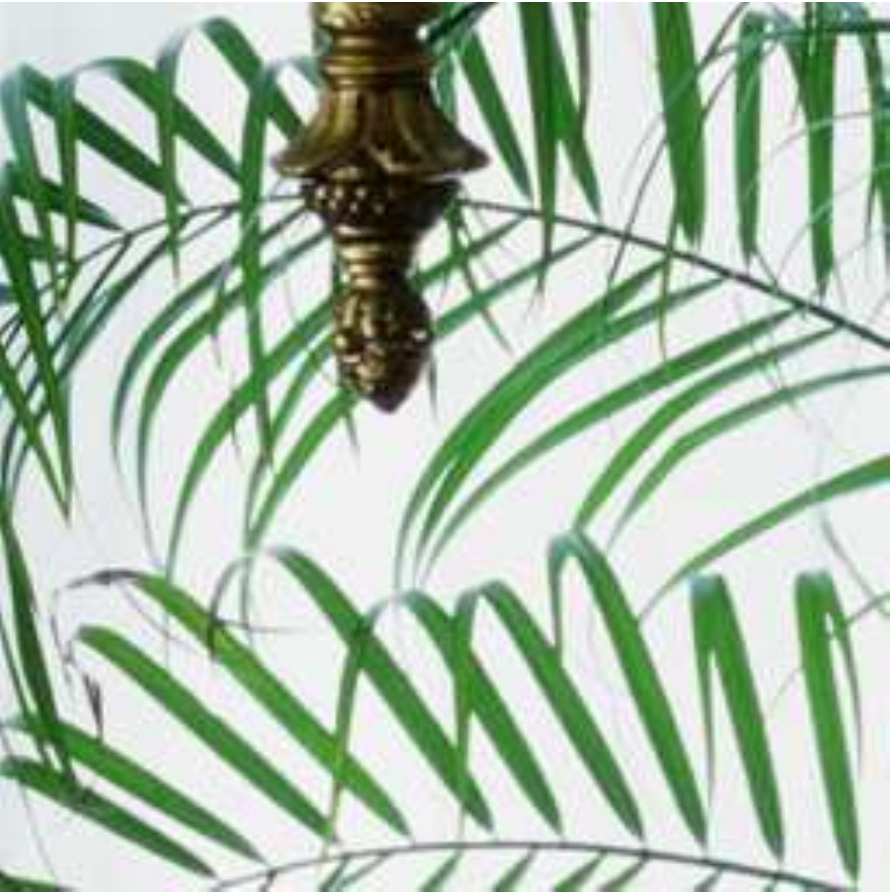}}
   \subfigure{\includegraphics[width=0.18\linewidth]{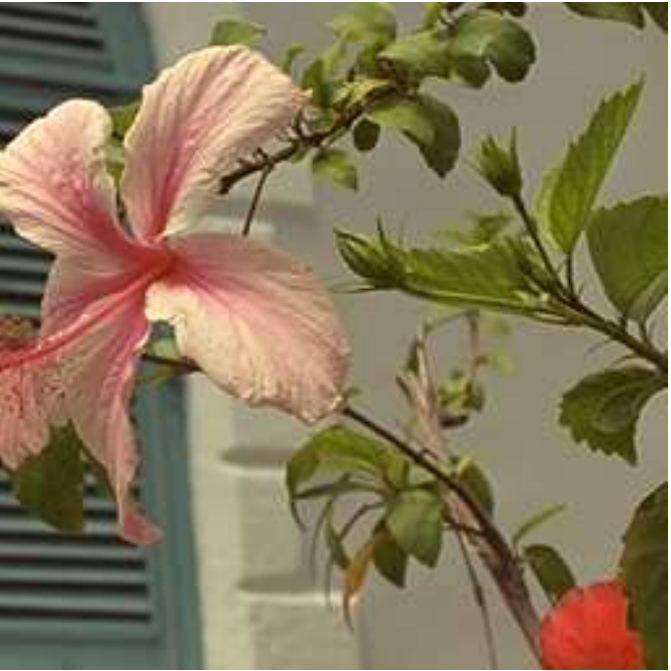}}      
\end{center}
  \caption{Test images for super-resolution: Butterfly, Bike, Hat, Plants, Girl, Parrot, Parthenon, Raccoon, Leaves, Flower.}
\label{fig:images}
\end{figure}

We now describe the details of our experimental setting for the superresolution problem. In the inverse problem $\outp=\Theta\inp+\nu$ in \eqref{eqn:inverse}, $\inp$ and $\outp$ denote respectively the lexicographical representations of the unknown image $X$ and the degraded image $Y$. The degradation matrix $\Theta=DH$ is composed of a down-sampling operator $D$ with a scale factor of $q=3$ and a Gaussian filter $H$ of size $7 \times 7$ with a standard deviation of $1.6$, and $\nu$ is an additive noise. We aim to recover the unknown image vector $\inp$ from the observed image vector $\outp$. We evaluate the proposed algorithms on the $10$ images presented in Figure \ref{fig:images}, which differ in their frequency characteristics and content. For color images, we apply the single image SR algorithm only on the luminance channel and we compute the PSNR and SSIM \cite{Wang04image} only on the luminance channel for coherence. Besides PSNR and SSIM, the visual quality of the images is also used as a comparison metric.

%The unknown image $X$ is of size $\sqrt{N}\times\sqrt{N}$, the degraded image $Y$ is of size $\frac{\sqrt{N}}{\scale}\times\frac{\sqrt{N}}{\scale}$, where $\scale$ is a downsampling scale factor,

%%%%% FIGURE: USED IMAGES

Overlapping patches of size  $6 \times 6$ are used in the experiments. The original NCSR algorithm initializes the training set $\trset$ by extracting patches from several images in the scale space of the HR image. However, in our implementation we initialize the set of training patches by extracting them only from the low-resolution image; i.e., the $m$ initial training patches $\trp_i \in \mathbb{R}^n$ in $\trset= \left\{\trp_i  \right\}_{i=1}^{m}$ are extracted from the observed low-resolution (LR) image vector $\outp$. We learn online PCA bases using the training patches in $\trset$ with the proposed AGNN and GOC methods. In the original NCSR method, in every $P$ iterations of the IST algorithm, the training set $\trset$ is updated by extracting the training patches from the current version of the reconstructed image $\hatinp$ and the PCA bases are updated as well by repeating the neighborhood selection with the updated training data. In our experiments, we use the same training patches $\trset$ for the whole algorithm.

%The superresolved test image $\hatinp$ is estimated iteratively. As in the original NCSR, in every $P$ iterations of the IST algorithm, the set $\tset$ of test patches is also updated such that $\tset = \{ \tp_j \} = \left\{\hatinpp_j \right\}_{j=1}^{M}$ are extracted from the current estimation $\hatinp$ of the HR image vector. The $M$ test patches $\hatinpp_j \in \mathbb{R}^n$ have the same size as the training patches. Since the HR image vector is not known, in the beginning of the algorithm, $\hatinp$ is initialized by applying a bicubic interpolation on the LR image vector $\outp$. The updates of the bases and test sets are repeated $\xi$ times during the whole algorithm, such that the total number of iterations is given by $T=\xi P$. We use the same parameters as in \cite{Dong13nonlocally} in the IST algorithm for solving the $l_1$-minimization problem.

In Section \ref{ssec:agnn_goc}, we evaluate our methods AGNN and GOC by comparing their performance to some other clustering or nearest neighbor selection strategies in superresolution. In Section \ref{ssec:state_art}, we provide comparative experiments with several widely used superresolution algorithms and show that our proposed manifold-based neighborhood selection techniques can be used for improving the state of the art in superresolution.

In Sections \ref{ssec:agnn_goc} and \ref{ssec:state_art}, the results for the algorithms in comparison are obtained by using the software packages made publicly available by the corresponding authors \footnote{We would like to thank the authors of \cite{Dong13nonlocally}, \cite{Dong11image}, \cite{Donoser13replicator}, \cite{Peleg14a},  \cite{Shi00normalized}, \cite{Bezdek1984} and \cite{Dijkstra59a} for making their software packages publicly available.}.

\subsubsection{Performance Evaluation of AGNN and GOC}
\label{ssec:agnn_goc}

We compare the proposed AGNN and GOC methods  with 4 different clustering algorithms; namely,  the K-means algorithm (Kmeans), Fuzzy C-means clustering algorithm (FCM) \cite{Bezdek1984}, Spectral Clustering (SC) \cite{Shi00normalized}, Replicator Graph Clustering (RGC) \cite{Donoser13replicator}; and also with K-NN search using geodesic distance (GeoD). Among the clustering methods,  Kmeans and FCM employ the Euclidean distance as a dissimilarity measure, while SC and RGC are graph-based methods that consider the manifold structure of data. When testing these four methods, we cluster the training patches and compute a PCA basis for each cluster. Then, given a test patch, the basis of the cluster whose centroid has the smallest distance to the test patch is selected as done in the original NCSR algorithm where K-means is used. In the GeoD method, each test patch is reconstructed with the PCA basis computed from its nearest neighbors with respect to the geodesic distance numerically computed with Dijkstra's algorithm \cite{Dijkstra59a}. The idea of nearest neighbor selection with respect to the geodesic distance is also in the core of the methods proposed in \cite{Turaga10nearest} and \cite{Chaudhry10fast}. Note that the four reference clustering methods and GOC provide nonadaptive solutions for training subset selection, while the GeoD and the AGNN methods are adaptive.
 %All experiments were performed on an Intel Core i7-3687U 2.10GHz laptop PC under MatLab R2012b environment.

%%%% FIGURE 5 : VISUAL QUALITY EVALUATION
\begin{figure*}[!ht]
\begin{center}
\setcounter{subfigure}{0}   
\subfigure[LR image]{\includegraphics[width=0.1\linewidth]{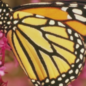}}   \subfigure[Original HR image]{\includegraphics[width=0.2\linewidth]{butterfly.pdf}}
   \subfigure[NCSR-Kmeans]{\includegraphics[width=0.2\linewidth]{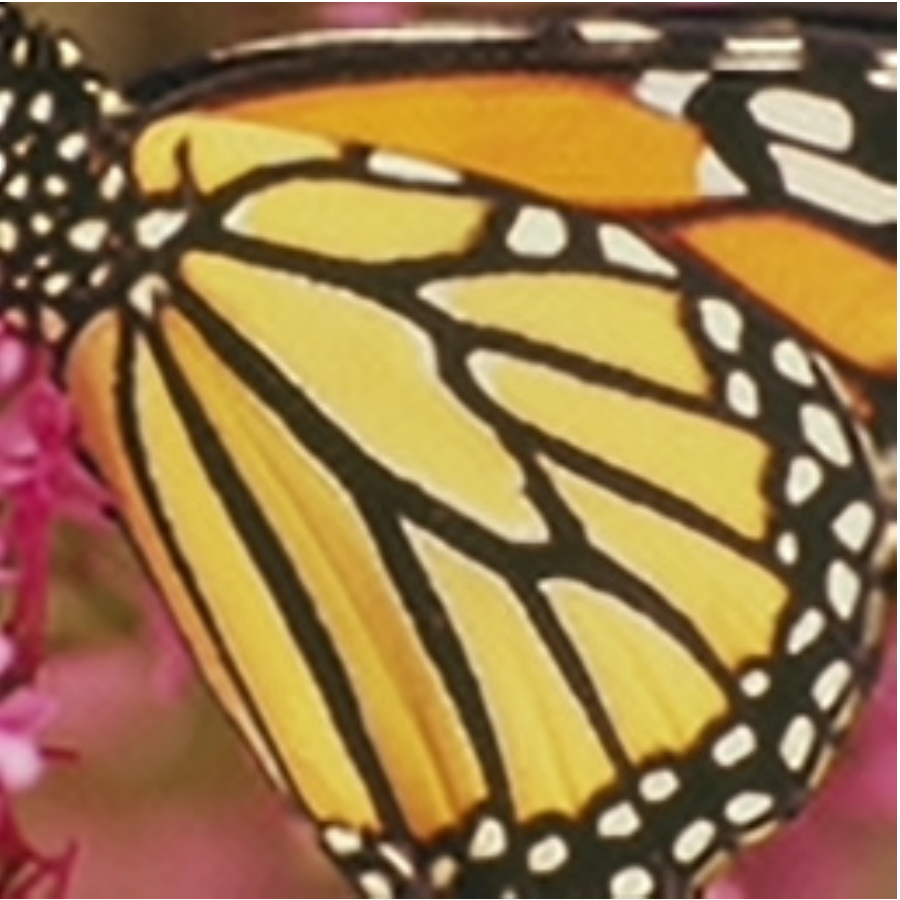}}
   \subfigure[NCSR-AGNN]{\includegraphics[width=0.2\linewidth]{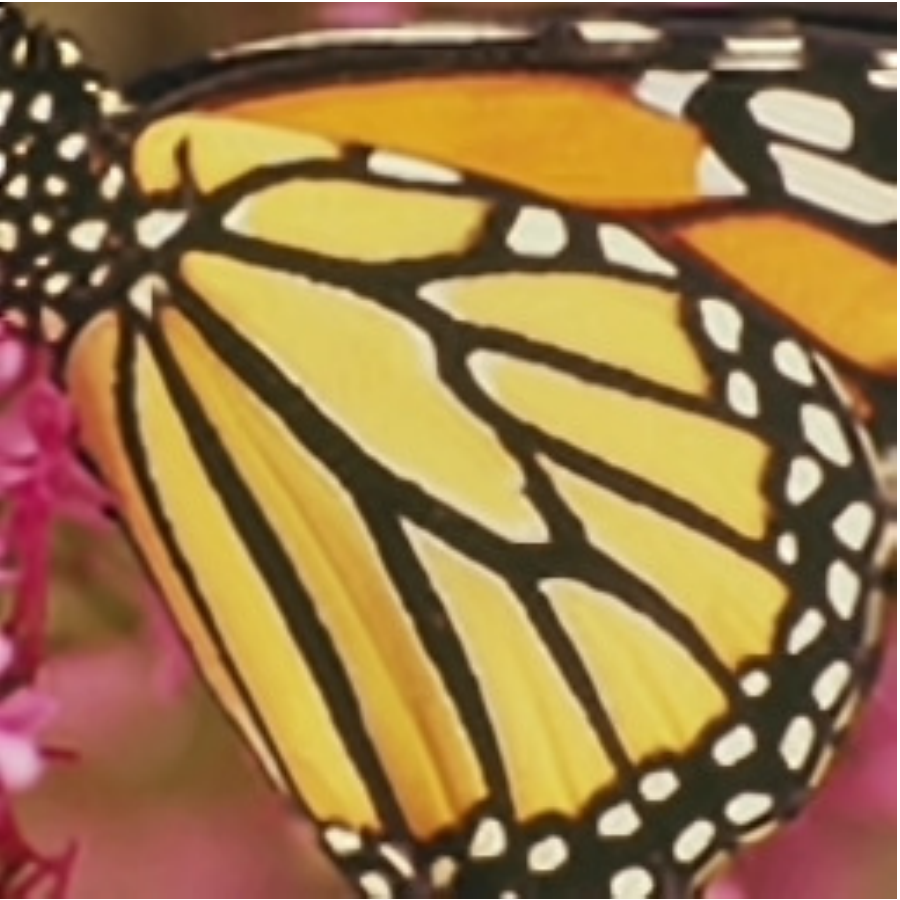}}
   \subfigure[NCSR-GOC]{\includegraphics[width=0.2\linewidth]{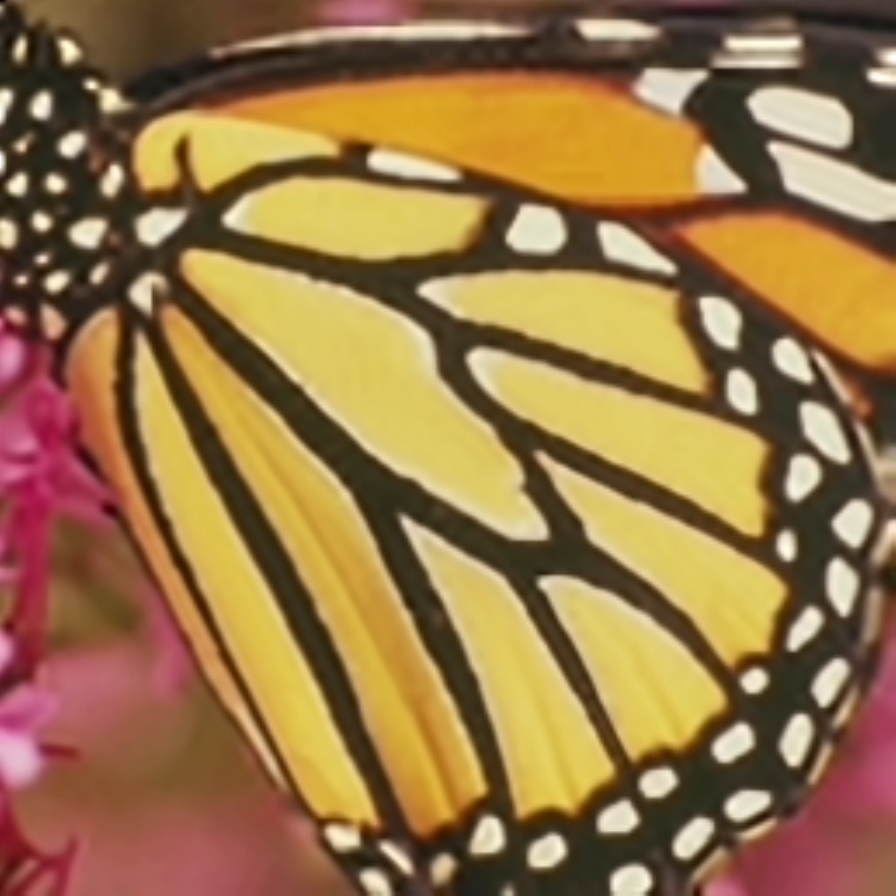}}
 %  \subfigure{\includegraphics[width=0.2\linewidth]{eps/zoomButterfly.eps}} \\ 
  \subfigure[Original HR close-up]{\includegraphics[width=0.2\linewidth]{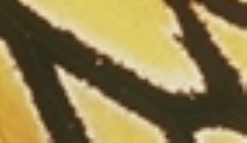}}
  \subfigure[NCSR-Kmeans close-up]{\includegraphics[width=0.2\linewidth]{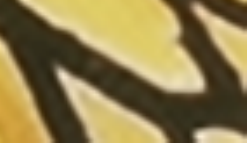}}
  \subfigure[NCSR-AGNN close-up]{\includegraphics[width=0.2\linewidth]{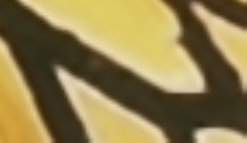}}
  \subfigure[NCSR-GOC close-up]{\includegraphics[width=0.2\linewidth]{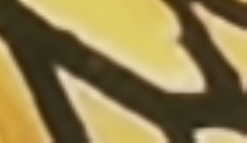}}\\
  \end{center}
  \caption{Comparison of SR results ($\times 3$). It can be observed that NCSR-AGNN and NCSR-GOC reconstruct edges with a higher contrast than NCSR-Kmeans.  Artifacts visible with NCSR-Kmeans (e.g., the oscillatory phantom bands perpendicular to the black stripes on the butterfly's wing) are significantly reduced with NCSR-AGNN and NCSR-GOC.}
\label{fig:visual_results_butterfly}
  \end{figure*}
\begin{figure*}[!ht]
\begin{center}
\subfigure[LR image]{\includegraphics[width=0.1\linewidth]{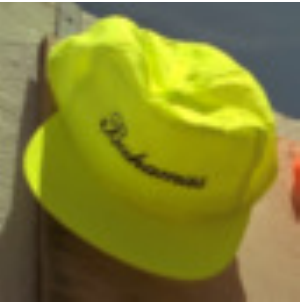}}   \subfigure[Original HR image]{\includegraphics[width=0.2\linewidth]{hat.pdf}}
   \subfigure[NCSR-Kmeans]{\includegraphics[width=0.2\linewidth]{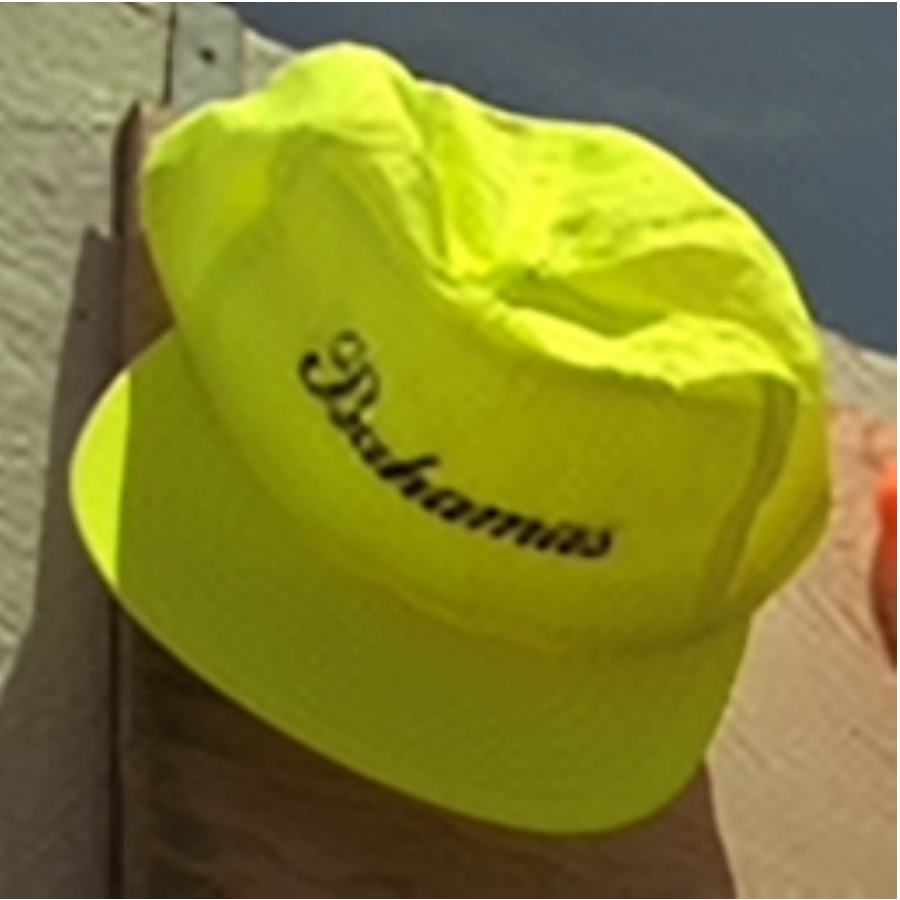}}
   \subfigure[NCSR-AGNN]{\includegraphics[width=0.2\linewidth]{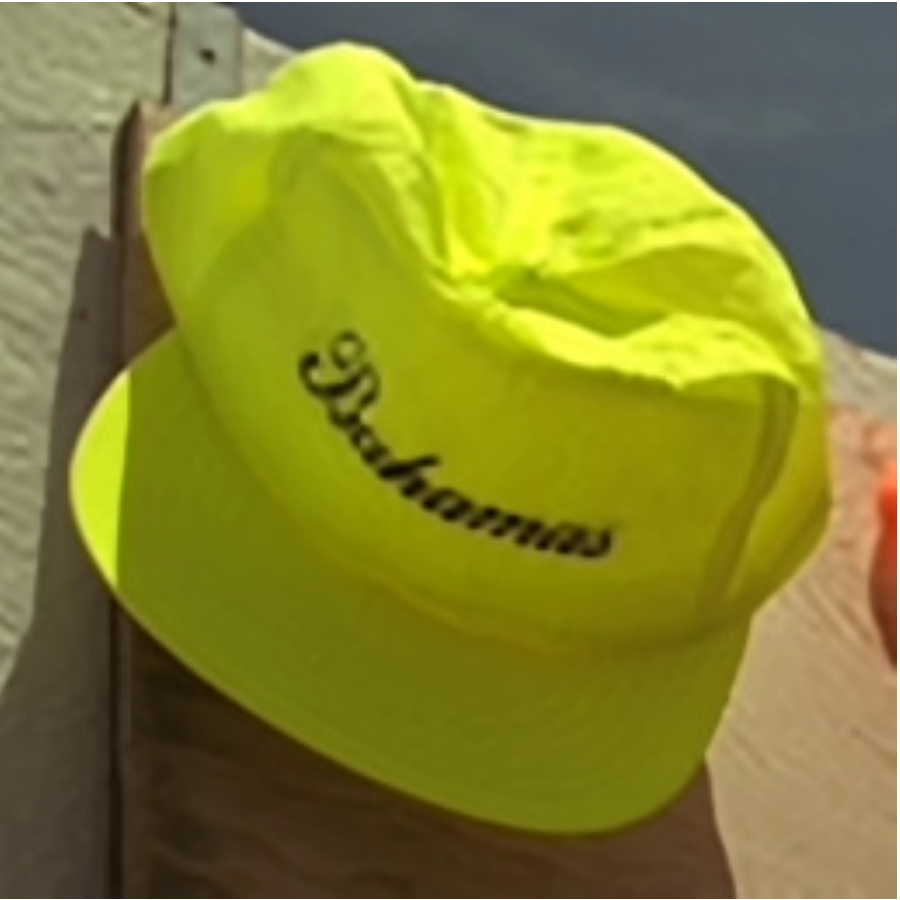}}
   \subfigure[NCSR-GOC]{\includegraphics[width=0.2\linewidth]{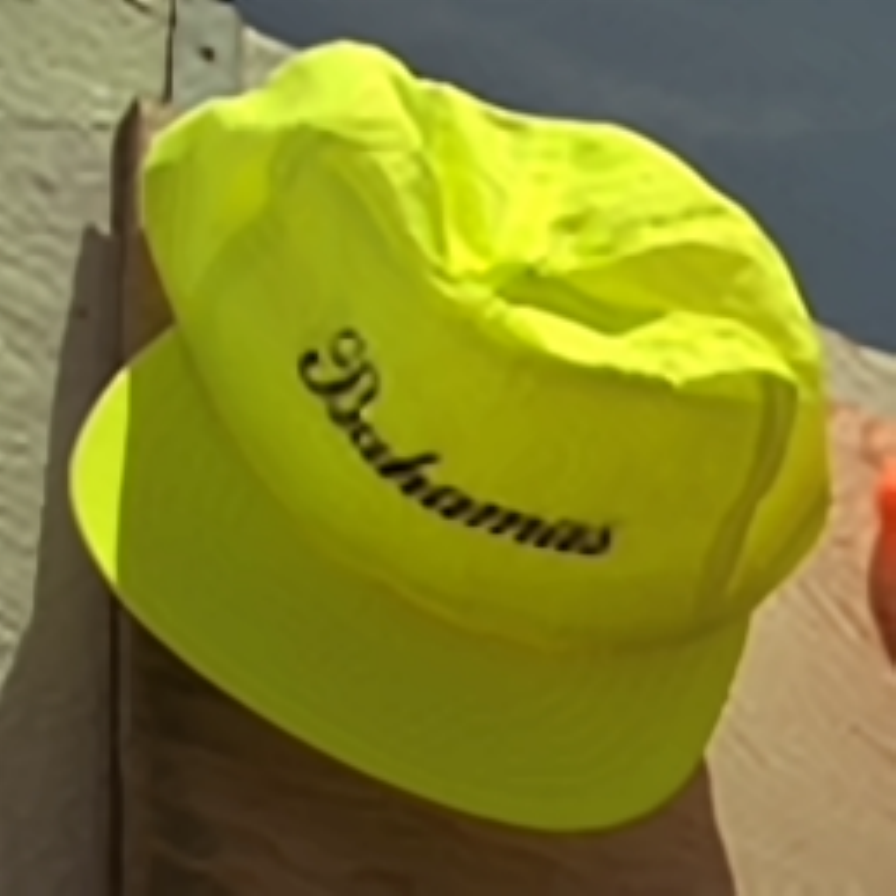}}
%   \subfigure[Close up of (b), (c), and (d)]{\includegraphics[width=0.2\linewidth]{eps/zoomHat.eps}} 
   \subfigure[Original HR close-up]{\includegraphics[width=0.2\linewidth]{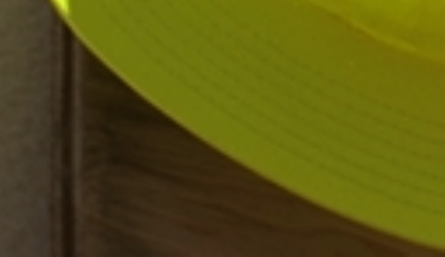}}
  \subfigure[NCSR-Kmeans close-up]{\includegraphics[width=0.2\linewidth]{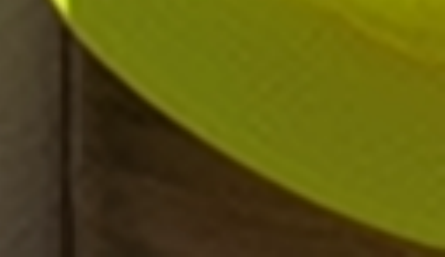}}
  \subfigure[NCSR-AGNN close-up]{\includegraphics[width=0.2\linewidth]{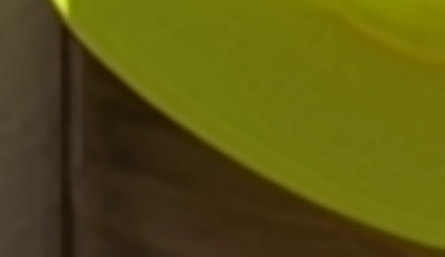}}
  \subfigure[NCSR-GOC close-up]{\includegraphics[width=0.2\linewidth]{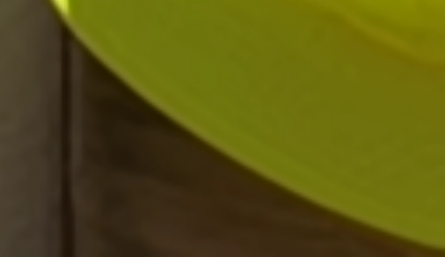}}\\
\end{center}
  \caption{Comparison of SR results ($\times 3$). NCSR-Kmeans produces artifacts such as the checkerboard-like noise patterns visible on plain regions of the cap, which are prevented by NCSR-AGNN or NCSR-GOC.}
\label{fig:visual_results_hat}
\end{figure*}

\begin{table*}[!t]%use {table*} to use 1 column
\scriptsize
\centering
\caption{PSNR (top row, in $\mathrm{d}\mathrm{B}$) and SSIM (bottom row) results for the luminance components of super-resolved HR images for different clustering or neighborhood selection approaches: Spectral Clustering (SC) \cite{Shi00normalized}; Fuzzy C-means clustering algorithm (FCM) \cite{Bezdek1984}; K-means clustering (Kmeans); Replicator Graph Clustering (RGC) \cite{Donoser13replicator}; kNN search with Dijkstra Algorithm (GeoD) \cite{Dijkstra59a}; and our methods GOC and AGNN. The methods are ordered according to the average PSNR values (from the lowest to the highest).}
\label{tbl:resultsAGNN}
\begin{IEEEeqnarraybox}[\IEEEeqnarraystrutmode\IEEEeqnarraystrutsizeadd{2pt}{0pt}]{x/r/Vx/r/x/r/x/r/x/r/x/r/x/r/x/r/x/r/x/r/x/r/x/r/x/r}
\IEEEeqnarraydblrulerowcut\\
&\hfill\raisebox{-8pt}[0pt][0pt]{\mbox{Images}}\hfill&&\IEEEeqnarraymulticol{23}{v}{}%
\IEEEeqnarraystrutsize{0pt}{0pt}\\
&&&&\hfill\mbox{Butterfly}\hfill&&\hfill\mbox{Bike}\hfill&&\hfill\mbox{Hat}\hfill&&\hfill\mbox{Plants}\hfill&&\hfill\mbox{Girl}\hfill&&\hfill\mbox{Parrot}\hfill&&\hfill\mbox{Parthenon}\hfill&&\hfill\mbox{Raccoon}\hfill&&\hfill\mbox{Leaves}\hfill&&\hfill\mbox{Flower}\hfill&&\hfill\mbox{Average}\hfill&\IEEEeqnarraystrutsizeadd{0pt}{2pt}\\
\IEEEeqnarraydblrulerowcut\\
&\hfill\raisebox{-15pt}[0pt][0pt]{\mbox{SC \cite{Shi00normalized}}}\hfill&&\IEEEeqnarraymulticol{23}{v}{}%
\IEEEeqnarraystrutsize{0pt}{0pt}\\
&&&&\hfill\mbox{28.15}\hfill&&\hfill\mbox{24.73}\hfill&&\hfill\mbox{31.28}\hfill&&\hfill\mbox{33.98}\hfill&&\hfill\mbox{33.65}\hfill&&\hfill\mbox{30.45}\hfill&&\hfill\mbox{27.19}\hfill&&\hfill\mbox{29.24}\hfill&&\hfill\mbox{27.50}\hfill&&\hfill\mbox{29.45}\hfill&&\hfill\mbox{29.56}\hfill&\IEEEeqnarraystrutsizeadd{0pt}{2pt}\\
&&&&\hfill\mbox{0.9193}\hfill&&\hfill\mbox{0.8026}\hfill&&\hfill\mbox{0.8723}\hfill&&\hfill\mbox{0.9198}\hfill&&\hfill\mbox{0.8255}\hfill&&\hfill\mbox{0.9170}\hfill&&\hfill\mbox{0.7509}\hfill&&\hfill\mbox{0.7659}\hfill&&\hfill\mbox{0.9242}\hfill&&\hfill\mbox{0.8567}\hfill&&\hfill\mbox{0.8554}\hfill&\IEEEeqnarraystrutsizeadd{0pt}{2pt}\\
\hline
&\hfill\raisebox{-15pt}[0pt][0pt]{\mbox{FCM \cite{Bezdek1984}}}\hfill&&\IEEEeqnarraymulticol{23}{v}{}%
\IEEEeqnarraystrutsize{0pt}{0pt}\\
&&&&\hfill\mbox{28.20}\hfill&&\hfill\mbox{24.76}\hfill&&\hfill\mbox{31.25}\hfill&&\hfill\mbox{33.99}\hfill&&\hfill\mbox{33.65}\hfill&&\hfill\mbox{30.47}\hfill&&\hfill\mbox{27.25}\hfill&&\hfill\mbox{29.25}\hfill&&\hfill\mbox{27.68}\hfill&&\hfill\mbox{29.50}\hfill&&\hfill\mbox{29.60}\hfill&\IEEEeqnarraystrutsizeadd{0pt}{2pt}\\
&&&&\hfill\mbox{0.9205}\hfill&&\hfill\mbox{0.8040}\hfill&&\hfill\mbox{0.8726}\hfill&&\hfill\mbox{0.9205}\hfill&&\hfill\mbox{0.8256}\hfill&&\hfill\mbox{0.9174}\hfill&&\hfill\mbox{0.7531}\hfill&&\hfill\mbox{0.7663}\hfill&&\hfill\mbox{0.9271}\hfill&&\hfill\mbox{0.8575}\hfill&&\hfill\mbox{0.8565}\hfill&\IEEEeqnarraystrutsizeadd{0pt}{2pt}\\
\hline
&\hfill\raisebox{-15pt}[0pt][0pt]{\mbox{Kmeans}}\hfill&&\IEEEeqnarraymulticol{23}{v}{}%
\IEEEeqnarraystrutsize{0pt}{0pt}\\
&&&&\hfill\mbox{28.14}\hfill&&\hfill\mbox{24.79}\hfill&&\hfill\mbox{31.31}\hfill&&\hfill\mbox{34.07}\hfill&&\hfill\mbox{33.64}\hfill&&\hfill\mbox{30.53}\hfill&&\hfill\mbox{27.20}\hfill&&\hfill\mbox{\textbf{29.28}}\hfill&&\hfill\mbox{27.67}\hfill&&\hfill\mbox{29.47}\hfill&&\hfill\mbox{29.61}\hfill&\IEEEeqnarraystrutsizeadd{0pt}{2pt}\\
&&&&\hfill\mbox{0.9204}\hfill&&\hfill\mbox{0.8050}\hfill&&\hfill\mbox{0.8730}\hfill&&\hfill\mbox{0.9213}\hfill&&\hfill\mbox{0.8254}\hfill&&\hfill\mbox{0.9178}\hfill&&\hfill\mbox{0.7517}\hfill&&\hfill\mbox{0.7668}\hfill&&\hfill\mbox{0.9265}\hfill&&\hfill\mbox{0.8567}\hfill&&\hfill\mbox{0.8565}\hfill&\IEEEeqnarraystrutsizeadd{0pt}{2pt}\\
\hline
&\hfill\raisebox{-15pt}[0pt][0pt]{\mbox{RGC \cite{Donoser13replicator}}}\hfill&&\IEEEeqnarraymulticol{23}{v}{}%
\IEEEeqnarraystrutsize{0pt}{0pt}\\
&&&&\hfill\mbox{28.45}\hfill&&\hfill\mbox{24.80}\hfill&&\hfill\mbox{31.37}\hfill&&\hfill\mbox{\textbf{34.20}}\hfill&&\hfill\mbox{33.65}\hfill&&\hfill\mbox{30.57}\hfill&&\hfill\mbox{27.22}\hfill&&\hfill\mbox{29.27}\hfill&&\hfill\mbox{27.90}\hfill&&\hfill\mbox{29.50}\hfill&&\hfill\mbox{29.69}\hfill&\IEEEeqnarraystrutsizeadd{0pt}{2pt}\\
&&&&\hfill\mbox{0.9234}\hfill&&\hfill\mbox{0.8061}\hfill&&\hfill\mbox{0.8739}\hfill&&\hfill\mbox{0.9219}\hfill&&\hfill\mbox{0.8254}\hfill&&\hfill\mbox{0.9181}\hfill&&\hfill\mbox{0.7525}\hfill&&\hfill\mbox{0.7658}\hfill&&\hfill\mbox{0.9317}\hfill&&\hfill\mbox{0.8576}\hfill&&\hfill\mbox{0.8576}\hfill&\IEEEeqnarraystrutsizeadd{0pt}{2pt}\\
\hline
&\hfill\raisebox{-15pt}[0pt][0pt]{\mbox{GeoD \cite{Dijkstra59a}}}\hfill&&\IEEEeqnarraymulticol{23}{v}{}%
\IEEEeqnarraystrutsize{0pt}{0pt}\\
&&&&\hfill\mbox{28.61}\hfill&&\hfill\mbox{24.82}\hfill&&\hfill\mbox{31.42}\hfill&&\hfill\mbox{34.16}\hfill&&\hfill\mbox{33.63}\hfill&&\hfill\mbox{30.44}\hfill&&\hfill\mbox{27.24}\hfill&&\hfill\mbox{29.25}\hfill&&\hfill\mbox{27.98}\hfill&&\hfill\mbox{29.54}\hfill&&\hfill\mbox{29.71}\hfill&\IEEEeqnarraystrutsizeadd{0pt}{2pt}\\
&&&&\hfill\mbox{0.9257}\hfill&&\hfill\mbox{0.8070}\hfill&&\hfill\mbox{0.8746}\hfill&&\hfill\mbox{0.9219}\hfill&&\hfill\mbox{0.8250}\hfill&&\hfill\mbox{0.9178}\hfill&&\hfill\mbox{0.7530}\hfill&&\hfill\mbox{0.7650}\hfill&&\hfill\mbox{0.9323}\hfill&&\hfill\mbox{0.8587}\hfill&&\hfill\mbox{0.8581}\hfill&\IEEEeqnarraystrutsizeadd{0pt}{2pt}\\
\hline
&\hfill\raisebox{-15pt}[0pt][0pt]{\mbox{avGOC}}\hfill&&\IEEEeqnarraymulticol{23}{v}{}%
\IEEEeqnarraystrutsize{0pt}{0pt}\\
&&&&\hfill\mbox{28.34}\hfill&&\hfill\mbox{24.85}\hfill&&\hfill\mbox{31.42}\hfill&&\hfill\mbox{34.17}\hfill&&\hfill\mbox{33.66}\hfill&&\hfill\mbox{30.68}\hfill&&\hfill\mbox{27.23}\hfill&&\hfill\mbox{\textbf{29.28}}\hfill&&\hfill\mbox{27.89}\hfill&&\hfill\mbox{29.55}\hfill&&\hfill\mbox{29.71}\hfill&\IEEEeqnarraystrutsizeadd{0pt}{2pt}\\
&&&&\hfill\mbox{0.9222}\hfill&&\hfill\mbox{0.8076}\hfill&&\hfill\mbox{0.8747}\hfill&&\hfill\mbox{0.9224}\hfill&&\hfill\mbox{0.8258}\hfill&&\hfill\mbox{0.9191}\hfill&&\hfill\mbox{0.7528}\hfill&&\hfill\mbox{0.7668}\hfill&&\hfill\mbox{0.9317}\hfill&&\hfill\mbox{0.8591}\hfill&&\hfill\mbox{0.8582}\hfill&\IEEEeqnarraystrutsizeadd{0pt}{2pt}\\
\hline
&\hfill\raisebox{-15pt}[0pt][0pt]{\mbox{aGOC}}\hfill&&\IEEEeqnarraymulticol{23}{v}{}%
\IEEEeqnarraystrutsize{0pt}{0pt}\\
&&&&\hfill\mbox{28.46}\hfill&&\hfill\mbox{24.85}\hfill&&\hfill\mbox{31.44}\hfill&&\hfill\mbox{34.18}\hfill&&\hfill\mbox{33.65}\hfill&&\hfill\mbox{30.63}\hfill&&\hfill\mbox{27.23}\hfill&&\hfill\mbox{29.27}\hfill&&\hfill\mbox{27.92}\hfill&&\hfill\mbox{29.54}\hfill&&\hfill\mbox{29.72}\hfill&\IEEEeqnarraystrutsizeadd{0pt}{2pt}\\
&&&&\hfill\mbox{0.9239}\hfill&&\hfill\mbox{0.8082}\hfill&&\hfill\mbox{0.8744}\hfill&&\hfill\mbox{\textbf{0.9227}}\hfill&&\hfill\mbox{0.8257}\hfill&&\hfill\mbox{0.9187}\hfill&&\hfill\mbox{0.7530}\hfill&&\hfill\mbox{0.7663}\hfill&&\hfill\mbox{\textbf{0.9324}}\hfill&&\hfill\mbox{0.8588}\hfill&&\hfill\mbox{0.8584}\hfill&\IEEEeqnarraystrutsizeadd{0pt}{2pt}\\
\hline
&\hfill\raisebox{-15pt}[0pt][0pt]{\mbox{mGOC}}\hfill&&\IEEEeqnarraymulticol{23}{v}{}%
\IEEEeqnarraystrutsize{0pt}{0pt}\\
&&&&\hfill\mbox{28.54}\hfill&&\hfill\mbox{\textbf{24.90}}\hfill&&\hfill\mbox{31.43}\hfill&&\hfill\mbox{\textbf{34.20}}\hfill&&\hfill\mbox{\textbf{33.67}}\hfill&&\hfill\mbox{\textbf{30.71}}\hfill&&\hfill\mbox{27.25}\hfill&&\hfill\mbox{\textbf{29.28}}\hfill&&\hfill\mbox{27.95}\hfill&&\hfill\mbox{29.55}\hfill&&\hfill\mbox{29.75}\hfill&\IEEEeqnarraystrutsizeadd{0pt}{2pt}\\
&&&&\hfill\mbox{0.9251}\hfill&&\hfill\mbox{\textbf{0.8085}}\hfill&&\hfill\mbox{0.8748}\hfill&&\hfill\mbox{0.9222}\hfill&&\hfill\mbox{\textbf{0.8261}}\hfill&&\hfill\mbox{\textbf{0.9192}}\hfill&&\hfill\mbox{0.7530}\hfill&&\hfill\mbox{\textbf{0.7671}}\hfill&&\hfill\mbox{\textbf{0.9324}}\hfill&&\hfill\mbox{0.8593}\hfill&&\hfill\mbox{0.8588}\hfill&\IEEEeqnarraystrutsizeadd{0pt}{2pt}\\
\hline
&\hfill\raisebox{-15pt}[0pt][0pt]{\mbox{AGNN}}\hfill&&\IEEEeqnarraymulticol{23}{v}{}%
\IEEEeqnarraystrutsize{0pt}{0pt}\\
&&&&\hfill\mbox{\textbf{28.78}}\hfill&&\hfill\mbox{24.87}\hfill&&\hfill\mbox{\textbf{31.46}}\hfill&&\hfill\mbox{34.16}\hfill&&\hfill\mbox{\textbf{33.67}}\hfill&&\hfill\mbox{30.60}\hfill&&\hfill\mbox{\textbf{27.29}}\hfill&&\hfill\mbox{29.26}\hfill&&\hfill\mbox{\textbf{28.01}}\hfill&&\hfill\mbox{\textbf{29.61}}\hfill&&\hfill\mbox{\textbf{29.77}}\hfill&\IEEEeqnarraystrutsizeadd{0pt}{2pt}\\
&&&&\hfill\mbox{\textbf{0.9266}}\hfill&&\hfill\mbox{0.8081}\hfill&&\hfill\mbox{\textbf{0.8749}}\hfill&&\hfill\mbox{0.9218}\hfill&&\hfill\mbox{0.8260}\hfill&&\hfill\mbox{0.9188}\hfill&&\hfill\mbox{\textbf{0.7540}}\hfill&&\hfill\mbox{0.7661}\hfill&&\hfill\mbox{\textbf{0.9324}}\hfill&&\hfill\mbox{\textbf{0.8601}}\hfill&&\hfill\mbox{\textbf{0.8589}}\hfill&\IEEEeqnarraystrutsizeadd{0pt}{2pt}\\
\IEEEeqnarraydblrulerowcut\\
\end{IEEEeqnarraybox}
\end{table*}

The parameters of the AGNN algorithm are set as  $\sRGC=35$ (number of nearest neighbors in the diffusion stage of RGC \cite{Donoser13replicator}), $\kappa=2$ (number of iterations for diffusing the affinity matrix),  $c_1=10$ (Gaussian kernel scale), and $c_2=0.9$ (affinity threshold). The parameters of the GOC algorithm are set as $\numcl=64$ (number of clusters), $c_3 = 0.5$ (threshold defining the decay function), $\gamma=150$, and $r=8$ (parameters for selecting a PCA basis for each test patch). The number of clusters in the other four clustering methods in comparison are also set to the same value as $\numcl=64$. The size of the clusters with the FCM algorithm are selected roughly the same as the cluster sizes computed with K-means. The total number of iterations and the number of PCA basis updates are chosen as $1000$ and $4$ in the NCSR algorithm. All the general parameters for the NCSR algorithm are selected as Dong et al. \cite{Dong13nonlocally}. In this way, we can maintain consistency in the comparison of the methods related to NCSR algorithm.

We evaluate the GOC algorithm in three different settings. In the first setting  the cluster size parameters $\itmax$ and $\Kball$ are estimated adaptively for each cluster with the strategy proposed in Algorithm \ref{alg:goc}, which is denoted as aGOC. In the second setting, denoted avGOC, the parameters $\itmax$ and $\Kball$ are not adapted to each cluster; all clusters are formed with the same parameter values, where $\itmax$ and $\Kball$ are computed by minimizing the average value of coefficient decay function $\tilde \idecay(\itmax, \Kball) $ over all clusters of the same image. The parameters are thus adapted to the images, but not to the individual clusters of patches of an image. Finally, in the third setting, denoted mGOC, the parameters $\itmax$ and $\Kball$ are manually entered and used for all clusters of the same image. The parameter values provided to the algorithm for each image are set as the best values  obtained with an exhaustive search. Therefore, mGOC can be considered as an oracle setting.

The results are presented in Figure \ref{fig:visual_results_butterfly}, Figure \ref{fig:visual_results_hat}, and Table \ref{tbl:resultsAGNN}. Figures \ref{fig:visual_results_butterfly} and \ref{fig:visual_results_hat} provide a visual comparison between the image reconstruction qualities obtained with the K-means clustering algorithm and the proposed AGNN and GOC methods for the Butterfly and the Hat images. It is observed that AGNN and GOC produce sharper edges than K-means. Moreover, the visual artifacts produced by K-means such as the phantom perpendicular bands on the black stripes of the butterfly and the checkerboard-like noise patterns on the cap are significantly reduced with AGNN and GOC. The efficiency of the proposed methods for removing these artifacts can be explained as follows. When image patches are clustered with the K-means algorithm, the similarity between patches is measured with the Euclidean distance. Therefore, when reconstructing a test patch, the algorithm tends to use a basis computed with patches that have similar intensity values. The nonuniformity of the pixel intensities along the black stripes of the LR Butterfly image thus propagates to the reconstructed HR image as well, which produces the phantom bands on the wing (due to the too low resolution, the black stripes on the LR image contain periodically appearing clear pixels contaminated by the yellow plain regions on the wing). Similarly, in the Hat image, the clusters used in learning a basis for reconstructing the plain regions on the cap contain also patches extracted from the wall, which have a similar intensity with the cap. This reproduces the shadowy patterns of the wall also on the cap. On the other hand, the AGNN method groups together patches that have a connection on the data graph. As the patches are extracted with overlapping windows shifting by one pixel, AGNN and GOC may have a stronger tendency than K-means for favoring patches from nearby or similar regions on the image that all share a common structure, which is also confirmed by the experiment in Section \ref{ssec:exp_struc_sim}. The proposed methods yield local bases better fitted to the characteristics of patches, therefore, less artifacts are observed.

In Table \ref{tbl:resultsAGNN} the performance of the compared clustering methods are measured with the PSNR and the SSIM metrics. Graph-based methods are generally seen to  yield a better performance than methods based on Euclidean distance. This confirms the intuition that motivates our study; when selecting neighborhoods for learning local models, the geometry of the data should be respected. As far as the average performance is concerned, the AGNN method gives the highest reconstruction quality and is followed by the GOC method. The performance difference between AGNN and GOC can be justified with the fact that the training subset selection is adaptive to the test patches in AGNN, while GOC is a nonadaptive method that offers a less complex solution. In particular, with a non-optimized implementation of our algorithms, we have observed that GOC has roughly the same computation time as K-means, while the computation time of AGNN is around three times K-means and GOC in the tested images on an Intel Core i5 2.6GHz under the Matlab R2015a programming environment, as shown in Table \ref{tbl:resultsTime}. K-means and GOC in the tested images. After the proposed AGNN and GOC methods, GeoD gives the best average performance. While this adaptive method ensures a good reconstruction quality, it requires the computation of the geodesic distance between each test patch and all training patches. Therefore, it is computationally very complex. Although several works such as \cite{Turaga10nearest} and \cite{Chaudhry10fast} provide solutions for fast approximations of the geodesic distance, we observe that in terms of reconstruction quality AGNN performs better than GeoD in most images. This suggests that using a globally consistent affinity measure optimized with respect to the entire graph topology provides a more refined and precise similarity metric than the geodesic distance, which only takes into account the shortest paths between samples.

Concerning the performances of the clustering methods on the individual images, an important conclusion is that geometry-based methods yield a better performance especially for images that contain patches of rich texture. The AGNN and GOC methods provide a performance gain of respectively $0.64$ dB and $0.4$ dB over K-means (used in the original NCSR method) for the Butterfly image. Meanwhile, all clustering methods give similar reconstruction qualities for the Girl image. This discrepancy can be explained with the difference in the characteristics of the patch manifolds of these two images. The patches of the Butterfly image contain high-frequency textures; therefore,  the patch manifold has a large curvature (see, e.g., \cite{Vural10curvature} for a study of the relation between the manifold curvature and the image characteristics). Consequently, the proposed methods adapted to the local geometry of the manifold perform better on this image. On the other hand, the Girl image mostly contains weakly textured low-frequency patches, which generate a rather flat patch manifold of small curvature. The Euclidean distance is more reliable as a dissimilarity measure on flat manifolds compared to curved manifolds as it gets closer to the geodesic distance. Hence, the performance gain of geometry-based methods over K-means is much smaller on the Girl image compared to Butterfly.

Next, the comparison of the three modes of the GOC algorithm shows that aGOC and avGOC yield reconstruction qualities that are close to that of the oracle method mGOC. This suggests that setting the parameters $\itmax$ and $\Kball$ with respect to the PCA coefficient decay rates as proposed in Algorithm \ref{alg:goc} provides an efficient strategy for the automatic determination of cluster sizes. While the average performances of aGOC and avGOC are quite close, interestingly, aGOC performs better than avGOC on Butterfly and Leaves. Both of these two images contain patches of quite varying characteristics, e.g., highly textured regions formed by repetitive edges as well as weakly textured regions. As the structures of the patches change significantly among different clusters in these images, optimizing the cluster size parameters individually for each cluster in aGOC has an advantage over using common parameters in avGOC. %A small cluster may be preferable to a large cluster in regions where the patch manifold has high curvature, whereas a relatively large cluster may work better where the patch manifold is flat. This observation provides interesting evidence for the intuition that respecting the geometry of data may improve the performance of image reconstruction.

%This work cannot be compared with approach that consider bicubic filter as blur kernel instead of Gaussian filter, \cite{Yang10image}, \cite{He13beta}, and \cite{Bevilacqua14single}.

\begin{table*}[!t]%use {table*} to use 1 column
\scriptsize
\centering
\caption{PSNR  (top row, in $\mathrm{d}\mathrm{B}$) and SSIM (bottom row) results for the luminance components of super-resolved HR images for different superresolution algorithms: Bicubic Interpolation; SPSR (Peleg et al.) \cite{Peleg14a}; ASDS (Dong et al.) \cite{Dong11image}; NCSR (Dong et al.) \cite{Dong13nonlocally}; NCSR with proposed GOC; NCSR with proposed AGNN. The methods are ordered according to the average PSNR values (from the lowest to the highest).}
\label{tbl:resultAplication}
\begin{IEEEeqnarraybox}[\IEEEeqnarraystrutmode\IEEEeqnarraystrutsizeadd{2pt}{0pt}]{x/r/Vx/r/x/r/x/r/x/r/x/r/v/r/v/r/x/r/x/r/x/r/x/r/v/r/x/r}
\IEEEeqnarraydblrulerowcut\\
&\hfill\raisebox{-8pt}[0pt][0pt]{\mbox{Images}}\hfill&&\IEEEeqnarraymulticol{25}{v}{}%
\IEEEeqnarraystrutsize{0pt}{0pt}\\
&&&&\hfill\mbox{Butterfly}\hfill&&\hfill\mbox{Bike}\hfill&&\hfill\mbox{Hat}\hfill&&\hfill\mbox{Plants}\hfill&&\hfill\mbox{Leaves}\hfill&&\hfill\mbox{\textit{Average}}\hfill&&\hfill\mbox{Parrot}\hfill&&\hfill\mbox{Parthenon}\hfill&&\hfill\mbox{Raccoon}\hfill&&\hfill\mbox{Girl}\hfill&&\hfill\mbox{Flower}\hfill&&\hfill\mbox{\textit{Average}}\hfill&\IEEEeqnarraystrutsizeadd{0pt}{2pt}\\
\hline
&\hfill\raisebox{-15pt}[0pt][0pt]{\mbox{Bicubic}}\hfill&&\IEEEeqnarraymulticol{25}{v}{}%
\IEEEeqnarraystrutsize{0pt}{0pt}\\
&&&&\hfill\mbox{22.41}\hfill&&\hfill\mbox{21.77}\hfill&&\hfill\mbox{28.22}\hfill&&\hfill\mbox{29.69}\hfill&&\hfill\mbox{21.73}\hfill&&\hfill\mbox{\textit{24.76}}\hfill&&\hfill\mbox{26.54}\hfill&&\hfill\mbox{25.20}\hfill&&\hfill\mbox{27.54}\hfill&&\hfill\mbox{31.65}\hfill&&\hfill\mbox{26.16}\hfill&&\hfill\mbox{\textit{27.42}}\hfill&\IEEEeqnarraystrutsizeadd{0pt}{2pt}\\
&&&&\hfill\mbox{0.7705}\hfill&&\hfill\mbox{0.6299}\hfill&&\hfill\mbox{0.8056}\hfill&&\hfill\mbox{0.8286}\hfill&&\hfill\mbox{0.7302}\hfill&&\hfill\mbox{\textit{0.7530}}\hfill&&\hfill\mbox{0.8493}\hfill&&\hfill\mbox{0.6528}\hfill&&\hfill\mbox{0.6737}\hfill&&\hfill\mbox{0.7671}\hfill&&\hfill\mbox{0.7295}\hfill&&\hfill\mbox{\textit{0.7345}}\hfill&\IEEEeqnarraystrutsizeadd{0pt}{2pt}\\
\hline
&\hfill\raisebox{-15pt}[0pt][0pt]{\mbox{SPSR \cite{Peleg14a}}}\hfill&&\IEEEeqnarraymulticol{25}{v}{}%
\IEEEeqnarraystrutsize{0pt}{0pt}\\
&&&&\hfill\mbox{26.74}\hfill&&\hfill\mbox{24.31}\hfill&&\hfill\mbox{30.84}\hfill&&\hfill\mbox{32.83}\hfill&&\hfill\mbox{25.84}\hfill&&\hfill\mbox{\textit{28.11}}\hfill&&\hfill\mbox{29.68}\hfill&&\hfill\mbox{26.77}\hfill&&\hfill\mbox{29.00}\hfill&&\hfill\mbox{33.40}\hfill&&\hfill\mbox{28.89}\hfill&&\hfill\mbox{\textit{29.55}}\hfill&\IEEEeqnarraystrutsizeadd{0pt}{2pt}\\
&&&&\hfill\mbox{0.8973}\hfill&&\hfill\mbox{0.7830}\hfill&&\hfill\mbox{0.8674}\hfill&&\hfill\mbox{0.9036}\hfill&&\hfill\mbox{0.8892}\hfill&&\hfill\mbox{\textit{0.8681}}\hfill&&\hfill\mbox{0.9089}\hfill&&\hfill\mbox{0.7310}\hfill&&\hfill\mbox{0.7562}\hfill&&\hfill\mbox{0.8211}\hfill&&\hfill\mbox{0.8415}\hfill&&\hfill\mbox{\textit{0.8117}}\hfill&\IEEEeqnarraystrutsizeadd{0pt}{2pt}\\
\hline
&\hfill\raisebox{-15pt}[0pt][0pt]{\mbox{ASDS \cite{Dong11image}}}\hfill&&\IEEEeqnarraymulticol{25}{v}{}%
\IEEEeqnarraystrutsize{0pt}{0pt}\\
&&&&\hfill\mbox{27.34}\hfill&&\hfill\mbox{24.62}\hfill&&\hfill\mbox{30.93}\hfill&&\hfill\mbox{33.47}\hfill&&\hfill\mbox{26.80}\hfill&&\hfill\mbox{\textit{28.63}}\hfill&&\hfill\mbox{30.00}\hfill&&\hfill\mbox{26.83}\hfill&&\hfill\mbox{29.24}\hfill&&\hfill\mbox{33.53}\hfill&&\hfill\mbox{29.19}\hfill&&\hfill\mbox{\textit{29.76}}\hfill&\IEEEeqnarraystrutsizeadd{0pt}{2pt}\\
&&&&\hfill\mbox{0.9047}\hfill&&\hfill\mbox{0.7962}\hfill&&\hfill\mbox{0.8706}\hfill&&\hfill\mbox{0.9095}\hfill&&\hfill\mbox{0.9058}\hfill&&\hfill\mbox{\textit{0.8774}}\hfill&&\hfill\mbox{0.9093}\hfill&&\hfill\mbox{0.7349}\hfill&&\hfill\mbox{0.7677}\hfill&&\hfill\mbox{0.8242}\hfill&&\hfill\mbox{0.8480}\hfill&&\hfill\mbox{\textit{0.8168}}\hfill&\IEEEeqnarraystrutsizeadd{0pt}{2pt}\\
\hline
&\hfill\raisebox{-15pt}[0pt][0pt]{\mbox{NCSR \cite{Dong13nonlocally}}}\hfill&&\IEEEeqnarraymulticol{25}{v}{}%
\IEEEeqnarraystrutsize{0pt}{0pt}\\
&&&&\hfill\mbox{28.07}\hfill&&\hfill\mbox{24.74}\hfill&&\hfill\mbox{31.29}\hfill&&\hfill\mbox{34.05}\hfill&&\hfill\mbox{27.46}\hfill&&\hfill\mbox{\textit{29.12}}\hfill&&\hfill\mbox{30.49}\hfill&&\hfill\mbox{27.18}\hfill&&\hfill\mbox{29.27}\hfill&&\hfill\mbox{33.66}\hfill&&\hfill\mbox{29.50}\hfill&&\hfill\mbox{\textit{30.02}}\hfill&\IEEEeqnarraystrutsizeadd{0pt}{2pt}\\
&&&&\hfill\mbox{0.9156}\hfill&&\hfill\mbox{0.8031}\hfill&&\hfill\mbox{0.8704}\hfill&&\hfill\mbox{0.9188}\hfill&&\hfill\mbox{0.9219}\hfill&&\hfill\mbox{\textit{0.8860}}\hfill&&\hfill\mbox{0.9147}\hfill&&\hfill\mbox{0.7510}\hfill&&\hfill\mbox{\textbf{0.7707}}\hfill&&\hfill\mbox{\textbf{0.8276}}\hfill&&\hfill\mbox{0.8563}\hfill&&\hfill\mbox{\textit{0.8241}}\hfill&\IEEEeqnarraystrutsizeadd{0pt}{2pt}\\
%
%\hline
%&\hfill\raisebox{-15pt}[0pt][0pt]{\mbox{SE-ASDS}}\hfill&&\IEEEeqnarraymulticol{23}{v}{}%
%\IEEEeqnarraystrutsize{0pt}{0pt}\\
%&&&&\hfill\mbox{28.48}\hfill&&\hfill\mbox{24.97}\hfill&&\hfill\mbox{31.53}\hfill&&\hfill\mbox{34.17}\hfill&&\hfill\mbox{33.56}\hfill&&\hfill\mbox{30.29}\hfill&&\hfill\mbox{27.05}\hfill&&\hfill\mbox{29.27}\hfill&&\hfill\mbox{27.69}\hfill&&\hfill\mbox{29.29}\hfill&&\hfill\mbox{29.63}\hfill&\IEEEeqnarraystrutsizeadd{0pt}{2pt}\\
%&&&&\hfill\mbox{0.9236}\hfill&&\hfill\mbox{0.8098}\hfill&&\hfill\mbox{0.8805}\hfill&&\hfill\mbox{0.9163}\hfill&&\hfill\mbox{0.8252}\hfill&&\hfill\mbox{0.9136}\hfill&&\hfill\mbox{0.7446}\hfill&&\hfill\mbox{0.7686}\hfill&&\hfill\mbox{0.9261}\hfill&&\hfill\mbox{0.8511}\hfill&&\hfill\mbox{0.8559}\hfill&\IEEEeqnarraystrutsizeadd{0pt}{2pt}\\
%
\hline
&\hfill\raisebox{-15pt}[0pt][0pt]{\mbox{NCSR-GOC}}\hfill&&\IEEEeqnarraymulticol{25}{v}{}%
\IEEEeqnarraystrutsize{0pt}{0pt}\\
&&&&\hfill\mbox{28.47}\hfill&&\hfill\mbox{24.85}\hfill&&\hfill\mbox{31.44}\hfill&&\hfill\mbox{34.16}\hfill&&\hfill\mbox{28.05}\hfill&&\hfill\mbox{\textit{29.39}}\hfill&&\hfill\mbox{\textbf{30.71}}\hfill&&\hfill\mbox{27.23}\hfill&&\hfill\mbox{\textbf{29.28}}\hfill&&\hfill\mbox{33.65}\hfill&&\hfill\mbox{29.58}\hfill&&\hfill\mbox{\textbf{\textit{30.09}}}\hfill&\IEEEeqnarraystrutsizeadd{0pt}{2pt}\\
&&&&\hfill\mbox{0.9241}\hfill&&\hfill\mbox{\textbf{0.8084}}\hfill&&\hfill\mbox{0.8747}\hfill&&\hfill\mbox{\textbf{0.9232}}\hfill&&\hfill\mbox{\textbf{0.9339}}\hfill&&\hfill\mbox{\textit{0.8929}}\hfill&&\hfill\mbox{\textbf{0.9192}}\hfill&&\hfill\mbox{0.7526}\hfill&&\hfill\mbox{0.7666}\hfill&&\hfill\mbox{0.8257}\hfill&&\hfill\mbox{0.8600}\hfill&&\hfill\mbox{\textit{0.8248}}\hfill&\IEEEeqnarraystrutsizeadd{0pt}{2pt}\\
\hline
&\hfill\raisebox{-15pt}[0pt][0pt]{\mbox{NCSR-AGNN}}\hfill&&\IEEEeqnarraymulticol{25}{v}{}%
\IEEEeqnarraystrutsize{0pt}{0pt}\\
&&&&\hfill\mbox{\textbf{28.81}}\hfill&&\hfill\mbox{\textbf{24.86}}\hfill&&\hfill\mbox{\textbf{31.47}}\hfill&&\hfill\mbox{\textbf{34.19}}\hfill&&\hfill\mbox{\textbf{28.06}}\hfill&&\hfill\mbox{\textit{\textbf{29.48}}}\hfill&&\hfill\mbox{30.60}\hfill&&\hfill\mbox{\textbf{27.30}}\hfill&&\hfill\mbox{29.27}\hfill&&\hfill\mbox{\textbf{33.67}}\hfill&&\hfill\mbox{\textbf{29.60}}\hfill&&\hfill\mbox{\textit{\textbf{30.09}}}\hfill&\IEEEeqnarraystrutsizeadd{0pt}{2pt}\\
&&&&\hfill\mbox{\textbf{0.9273}}\hfill&&\hfill\mbox{0.8080}\hfill&&\hfill\mbox{\textbf{0.8755}}\hfill&&\hfill\mbox{0.9223}\hfill&&\hfill\mbox{0.9332}\hfill&&\hfill\mbox{\textit{\textbf{0.8933}}}\hfill&&\hfill\mbox{0.9189}\hfill&&\hfill\mbox{\textbf{0.7546}}\hfill&&\hfill\mbox{0.7662}\hfill&&\hfill\mbox{0.8261}\hfill&&\hfill\mbox{\textbf{0.8601}}\hfill&&\hfill\mbox{\textit{\textbf{0.8252}}}\hfill&\IEEEeqnarraystrutsizeadd{0pt}{2pt}\\
\IEEEeqnarraydblrulerowcut\\
\end{IEEEeqnarraybox}
\end{table*}

\begin{table*}[!t]%use {table*} to use 1 column
\scriptsize
\centering
\caption{Running times for the luminance components of super-resolved HR images for different super-resolution algorithms: NCSR (Dong et al.) \cite{Dong13nonlocally}; NCSR with proposed GOC; NCSR with proposed AGNN.}
\label{tbl:resultsTime}
\begin{IEEEeqnarraybox}[\IEEEeqnarraystrutmode\IEEEeqnarraystrutsizeadd{2pt}{0pt}]{x/r/Vx/r/x/r/x/r/x/r/x/r/x/r/x/r/x/r/x/r/x/r/x/r/x/r}
\IEEEeqnarraydblrulerowcut\\
&\hfill\raisebox{-8pt}[0pt][0pt]{\mbox{Images}}\hfill&&\IEEEeqnarraymulticol{23}{v}{}%
\IEEEeqnarraystrutsize{0pt}{0pt}\\
&&&&\hfill\mbox{Butterfly}\hfill&&\hfill\mbox{Bike}\hfill&&\hfill\mbox{Hat}\hfill&&\hfill\mbox{Plants}\hfill&&\hfill\mbox{Leaves}\hfill&&\hfill\mbox{Parrot}\hfill&&\hfill\mbox{Parthenon}\hfill&&\hfill\mbox{Raccoon}\hfill&&\hfill\mbox{Girl}\hfill&&\hfill\mbox{Flower}\hfill&&\hfill\mbox{\textit{Average}}\hfill&\IEEEeqnarraystrutsizeadd{0pt}{2pt}\\
\hline
&\hfill\raisebox{-8pt}[0pt][0pt]{\mbox{NCSR \cite{Dong13nonlocally}}}\hfill&&\IEEEeqnarraymulticol{23}{v}{}%
\IEEEeqnarraystrutsize{0pt}{0pt}\\
&&&&\hfill\mbox{261}\hfill&&\hfill\mbox{229}\hfill&&\hfill\mbox{213}\hfill&&\hfill\mbox{229}\hfill&&\hfill\mbox{233}\hfill&&\hfill\mbox{220}\hfill&&\hfill\mbox{481}\hfill&&\hfill\mbox{362}\hfill&&\hfill\mbox{213}\hfill&&\hfill\mbox{226}\hfill&&\hfill\mbox{\textit{267}}\hfill&\IEEEeqnarraystrutsizeadd{0pt}{2pt}\\
\hline
&\hfill\raisebox{-8pt}[0pt][0pt]{\mbox{NCSR-GOC}}\hfill&&\IEEEeqnarraymulticol{23}{v}{}%
\IEEEeqnarraystrutsize{0pt}{0pt}\\
&&&&\hfill\mbox{271}\hfill&&\hfill\mbox{266}\hfill&&\hfill\mbox{253}\hfill&&\hfill\mbox{261}\hfill&&\hfill\mbox{278}\hfill&&\hfill\mbox{{256}}\hfill&&\hfill\mbox{518}\hfill&&\hfill\mbox{{383}}\hfill&&\hfill\mbox{246}\hfill&&\hfill\mbox{264}\hfill&&\hfill\mbox{{\textit{299}}}\hfill&\IEEEeqnarraystrutsizeadd{0pt}{2pt}\\
\hline
&\hfill\raisebox{-8pt}[0pt][0pt]{\mbox{NCSR-AGNN}}\hfill&&\IEEEeqnarraymulticol{23}{v}{}%
\IEEEeqnarraystrutsize{0pt}{0pt}\\
&&&&\hfill\mbox{{960}}\hfill&&\hfill\mbox{{1039}}\hfill&&\hfill\mbox{{467}}\hfill&&\hfill\mbox{{578}}\hfill&&\hfill\mbox{{1146}}\hfill&&\hfill\mbox{505}\hfill&&\hfill\mbox{{2541}}\hfill&&\hfill\mbox{1637}\hfill&&\hfill\mbox{{416}}\hfill&&\hfill\mbox{{830}}\hfill&&\hfill\mbox{\textit{{1012}}}\hfill&\IEEEeqnarraystrutsizeadd{0pt}{2pt}\\
\IEEEeqnarraydblrulerowcut\\
\end{IEEEeqnarraybox}
\end{table*}

\subsubsection{Improvements over the State of the Art in Superresolution}
\label{ssec:state_art}

In this section, we present an experimental comparison of several popular superresolution algorithms; namely, the bicubic interpolation algorithm, ASDS \cite{Dong11image}, SPSR \cite{Peleg14a}, and NCSR \cite{Dong13nonlocally}. We evaluate the performance of the NCSR algorithm under three different settings where the local bases are computed with K-means, AGNN, and GOC. The GOC method is used as in Algorithm \ref{alg:goc} (denoted as aGOC in the previous experiments). 

The experiments are conducted on the same images as in the previous set of experiments. The total number of iterations and the number of PCA basis updates of NCSR are selected respectively as $960$ and  $6$, while the other parameters are chosen as before. The results  presented in Table \ref{tbl:resultAplication} show that the  state of the art in superresolution is led by the NCSR method \cite{Dong13nonlocally}. The performance of NCSR is improved when it is coupled with the AGNN and GOC strategies for selecting local models. In Table \ref{tbl:resultAplication} the images are divided into two categories as those with high-frequency and low-frequency content. The average PSNR and SSIM metrics are reported in both groups. It can be observed that the advantage of the proposed neighborhood selection strategies over K-means is especially significant for high-frequency images. In images with low-frequency content, K-means gives the same performance as the proposed methods. As the patch manifold gets flatter, clusters obtained with K-means and the proposed methods get similar. Hence, we may conclude that the proposed geometry-based neighborhood selection methods can be successfully used for improving the state of the art in image superresolution, whose efficacy is especially observable for sharp images rich in high-frequency texture.

\subsection{Image deblurring}
\label{ssec:exp_deblur}

\begin{figure}[!t]
\begin{center}
\setcounter{subfigure}{0}   
\subfigure{\includegraphics[width=0.18\linewidth]{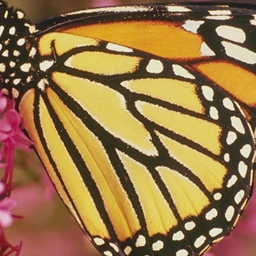}}   \subfigure{\includegraphics[width=0.18\linewidth]{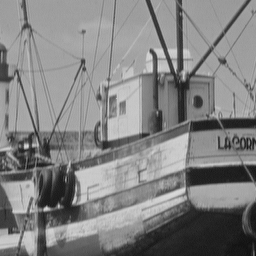}}
   \subfigure{\includegraphics[width=0.18\linewidth]{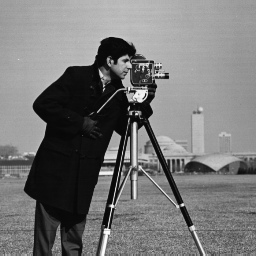}}
   \subfigure{\includegraphics[width=0.18\linewidth]{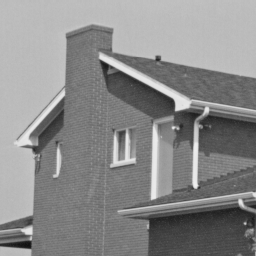}}
   \subfigure{\includegraphics[width=0.18\linewidth]{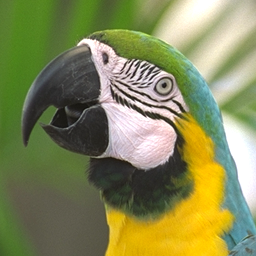}} \\ 
\setcounter{subfigure}{0}   
\subfigure{\includegraphics[width=0.18\linewidth]{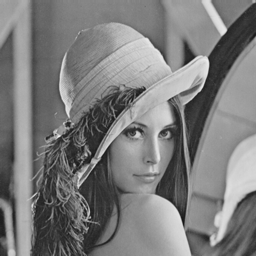}}   \subfigure{\includegraphics[width=0.18\linewidth]{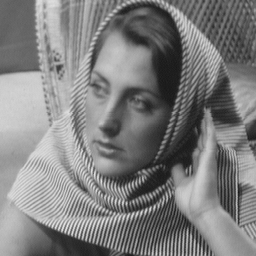}}  
\subfigure{\includegraphics[width=0.18\linewidth]{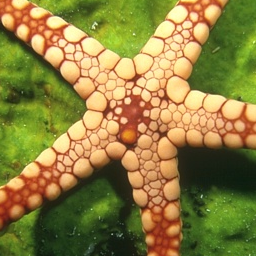}}   
\subfigure{\includegraphics[width=0.18\linewidth]{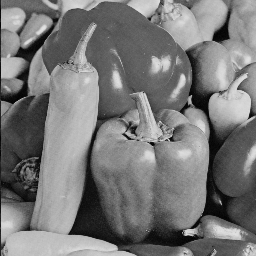}}   
\subfigure{\includegraphics[width=0.18\linewidth]{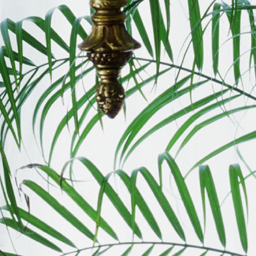}}    
\end{center}
  \caption{Test images for deblurring: Butterfly, Boats, Cameraman, House, Parrot, Lena, Barbara, Starfish, Peppers, Leaves.}
\label{fig:img_used_deblur}
\end{figure}

We now evaluate our method in the image deblurring application. Unlike the case in superresolution, the images to be deblurred have a normal resolution, which leads to a large number of patches for large images. In this case GOC has an advantage over AGNN in terms of complexity and memory requirements. Thus, it is more interesting to study the performance of the GOC algorithm in deblurring.  We compare GOC with the K-means clustering algorithm within the framework of the NCSR method \cite{Dong13nonlocally}. The algorithms are tested on the images shown in Figure \ref{fig:img_used_deblur}. Two blurring kernels are used, which are a uniform blur kernel of size $9\times 9$ pixels and a Gaussian blur kernel of standard deviation  $1.6$ pixels. Along with the blurring, the images are also corrupted with an additive white Gaussian noise of standard deviation $\sqrt{2}$. The parameters of GOC are set as $\numcl=64$ (number of clusters), $c_3 = 0.5$ (threshold defining the decay function), $\gamma=150$, and $r=8$ (parameters for selecting a PCA basis for each test patch). All the general parameters for the NCSR algorithm are selected as Dong et al. \cite{Dong13nonlocally} in order to maintain the consistency.

The PSNR and FSIM \cite{LinZhang11fsim} measures of the reconstruction qualities are presented in Table \ref{tbl:resultDeblur}. The results obtained with the image restoration algorithms FISTA (Portilla et al.) \cite{Portilla09image}, $l_0$-SPAR (Irani et al.) \cite{Irani93motion}, IDD-BM3D (Danielyan et al.) \cite{Danielyan12bm3d}, and ASDS (Dong et al.) \cite{Dong11image} reported in \cite{Dong13nonlocally} for the same experiments are also given for the purpose of comparison. The results show that the proposed GOC algorithm can be effectively used for improving the image reconstruction quality of the NCSR method in deblurring applications. The GOC method either outperforms the K-means clustering algorithm or yields a quite close performance when coupled with NCSR. Moreover, one can observe that the best average PSNR value is given by the proposed method, whose benefits are especially observable for images with significant high-frequency components such as Butterfly, Cameraman, and Leaves.

%%%%%% FIGURE: USED IMAGES IN DEBLURRING

%% DEBLURRING RESULTS WITH UNIFORM KERNEL with only FSIM
\begin{table*}[!t]%use {table*} to use 1 column
\scriptsize
\centering
\caption{PSNR  (top row, in $\mathrm{d}\mathrm{B}$) and FSIM (bottom row) results for the luminance components of deblurred images for different deblurring algorithms for uniform blur kernel and Gaussian blur kernel of standard deviation $1.6$ pixels: NCSR (Dong et al.) \cite{Dong13nonlocally}; NCSR with proposed GOC; FISTA (Portilla et al.) \cite{Portilla09image}; $l_0$-SPAR (Irani et al.) \cite{Irani93motion}; IDD-BM3D (Danielyan et al.) \cite{Danielyan12bm3d}, ASDS (Dong et al.) \cite{Dong11image}. The methods are ordered according to the average PSNR values (from the lowest to the highest).}
%\caption{PSNR  (top row, in $\mathrm{d}\mathrm{B}$) and FSIM (bottom row) results for the luminance components of deblurred images for different deblurring algorithms for uniform blur kernel and Gaussian blur kernel: NCSR (Dong et al.) \cite{Dong13nonlocally}; NCSR with subsampled patches;  NCSR with proposed GOC; FISTA (Portilla et al.) \cite{Portilla09image}; $l_0$-SPAR (Irani et al.) \cite{Irani93motion}; IDD-BM3D (Danielyan et al.) \cite{Danielyan12bm3d}, ASDS (Dong et al.) \cite{Dong11image}. The methods are ordered according to the average PSNR values (from the lowest to the highest).}
\label{tbl:resultDeblur}
\begin{IEEEeqnarraybox}[\IEEEeqnarraystrutmode\IEEEeqnarraystrutsizeadd{2pt}{0pt}]{x/r/Vx/r/x/r/x/r/x/r/x/r/x/r/x/r/x/r/x/r/x/r/x/r/x/r}
\IEEEeqnarraydblrulerowcut\\
&\hfill\raisebox{-8pt}[0pt][0pt]{\mbox{Images}}\hfill&&\IEEEeqnarraymulticol{23}{v}{}%
\IEEEeqnarraystrutsize{0pt}{0pt}\\
&&&&\hfill\mbox{Butterfly}\hfill&&\hfill\mbox{Boats}\hfill&&\hfill\mbox{C. Man}\hfill&&\hfill\mbox{House}\hfill&&\hfill\mbox{Parrot}\hfill&&\hfill\mbox{Lena}\hfill&&\hfill\mbox{Barbara}\hfill&&\hfill\mbox{Starfish}\hfill&&\hfill\mbox{Peppers}\hfill&&\hfill\mbox{Leaves}\hfill&&\hfill\mbox{Average}\hfill&\IEEEeqnarraystrutsizeadd{0pt}{2pt}\\
\IEEEeqnarraydblrulerowcut\\
&\hfill\raisebox{-15pt}[0pt][0pt]{\mbox{}}\hfill&&\IEEEeqnarraymulticol{23}{v}{}%
\IEEEeqnarraystrutsize{0pt}{0pt}\\
&&&&\hfill\mbox{\textbf{}}\hfill&&\hfill\mbox{\textbf{}}\hfill&&\hfill\mbox{}\hfill&&\hfill\mbox{}\hfill&&\hfill\mbox{}\hfill&&\hfill\mbox{Uniform}\hfill&&\hfill\mbox{}\hfill&&\hfill\mbox{}\hfill&&\hfill\mbox{\textbf{}}\hfill&&\hfill\mbox{\textbf{}}\hfill&&\hfill\mbox{\textbf{}}\hfill&\IEEEeqnarraystrutsizeadd{0pt}{2pt}\\
\hline
&\hfill\raisebox{-15pt}[0pt][0pt]{\mbox{FISTA \cite{Portilla09image}}}\hfill&&\IEEEeqnarraymulticol{23}{v}{}%
\IEEEeqnarraystrutsize{0pt}{0pt}\\
&&&&\hfill\mbox{28.37}\hfill&&\hfill\mbox{29.04}\hfill&&\hfill\mbox{26.82}\hfill&&\hfill\mbox{31.99}\hfill&&\hfill\mbox{29.11}\hfill&&\hfill\mbox{28.33}\hfill&&\hfill\mbox{25.75}\hfill&&\hfill\mbox{27.75}\hfill&&\hfill\mbox{28.43}\hfill&&\hfill\mbox{26.49}\hfill&&\hfill\mbox{28.21}\hfill&\IEEEeqnarraystrutsizeadd{0pt}{2pt}\\
&&&&\hfill\mbox{0.9119}\hfill&&\hfill\mbox{0.8858}\hfill&&\hfill\mbox{0.8627}\hfill&&\hfill\mbox{0.9017}\hfill&&\hfill\mbox{0.9002}\hfill&&\hfill\mbox{0.8798}\hfill&&\hfill\mbox{0.8375}\hfill&&\hfill\mbox{0.8775}\hfill&&\hfill\mbox{0.8813}\hfill&&\hfill\mbox{0.8958}\hfill&&\hfill\mbox{0.8834}\hfill&\IEEEeqnarraystrutsizeadd{0pt}{2pt}\\
\hline
&\hfill\raisebox{-15pt}[0pt][0pt]{\mbox{$l_0$-SPAR \cite{Irani93motion}}}\hfill&&\IEEEeqnarraymulticol{23}{v}{}%
\IEEEeqnarraystrutsize{0pt}{0pt}\\
&&&&\hfill\mbox{27.10}\hfill&&\hfill\mbox{29.86}\hfill&&\hfill\mbox{26.97}\hfill&&\hfill\mbox{32.98}\hfill&&\hfill\mbox{29.34}\hfill&&\hfill\mbox{28.72}\hfill&&\hfill\mbox{26.42}\hfill&&\hfill\mbox{28.11}\hfill&&\hfill\mbox{28.66}\hfill&&\hfill\mbox{26.30}\hfill&&\hfill\mbox{28.44}\hfill&\IEEEeqnarraystrutsizeadd{0pt}{2pt}\\
&&&&\hfill\mbox{0.8879}\hfill&&\hfill\mbox{0.9094}\hfill&&\hfill\mbox{0.8689}\hfill&&\hfill\mbox{0.9225}\hfill&&\hfill\mbox{0.9262}\hfill&&\hfill\mbox{0.9063}\hfill&&\hfill\mbox{0.8691}\hfill&&\hfill\mbox{0.8951}\hfill&&\hfill\mbox{0.9066}\hfill&&\hfill\mbox{0.8776}\hfill&&\hfill\mbox{0.8970}\hfill&\IEEEeqnarraystrutsizeadd{0pt}{2pt}\\
\hline
&\hfill\raisebox{-15pt}[0pt][0pt]{\mbox{ASDS \cite{Dong11image}}}\hfill&&\IEEEeqnarraymulticol{23}{v}{}%
\IEEEeqnarraystrutsize{0pt}{0pt}\\
&&&&\hfill\mbox{28.70}\hfill&&\hfill\mbox{30.80}\hfill&&\hfill\mbox{28.08}\hfill&&\hfill\mbox{34.03}\hfill&&\hfill\mbox{31.22}\hfill&&\hfill\mbox{29.92}\hfill&&\hfill\mbox{27.86}\hfill&&\hfill\mbox{29.72}\hfill&&\hfill\mbox{29.48}\hfill&&\hfill\mbox{28.59}\hfill&&\hfill\mbox{29.84}\hfill&\IEEEeqnarraystrutsizeadd{0pt}{2pt}\\
&&&&\hfill\mbox{0.9053}\hfill&&\hfill\mbox{0.9236}\hfill&&\hfill\mbox{0.8950}\hfill&&\hfill\mbox{0.9337}\hfill&&\hfill\mbox{0.9306}\hfill&&\hfill\mbox{\textbf{0.9256}}\hfill&&\hfill\mbox{0.9088}\hfill&&\hfill\mbox{0.9208}\hfill&&\hfill\mbox{0.9203}\hfill&&\hfill\mbox{0.9075}\hfill&&\hfill\mbox{0.9171}\hfill&\IEEEeqnarraystrutsizeadd{0pt}{2pt}\\
\hline
&\hfill\raisebox{-15pt}[0pt][0pt]{\mbox{IDD-BM3D \cite{Danielyan12bm3d}}}\hfill&&\IEEEeqnarraymulticol{23}{v}{}%
\IEEEeqnarraystrutsize{0pt}{0pt}\\
&&&&\hfill\mbox{29.21}\hfill&&\hfill\mbox{\textbf{31.20}}\hfill&&\hfill\mbox{28.56}\hfill&&\hfill\mbox{\textbf{34.44}}\hfill&&\hfill\mbox{31.06}\hfill&&\hfill\mbox{29.70}\hfill&&\hfill\mbox{27.98}\hfill&&\hfill\mbox{29.48}\hfill&&\hfill\mbox{29.62}\hfill&&\hfill\mbox{29.38}\hfill&&\hfill\mbox{30.06}\hfill&\IEEEeqnarraystrutsizeadd{0pt}{2pt}\\
&&&&\hfill\mbox{0.9287}\hfill&&\hfill\mbox{0.9304}\hfill&&\hfill\mbox{0.9007}\hfill&&\hfill\mbox{0.9369}\hfill&&\hfill\mbox{0.9364}\hfill&&\hfill\mbox{0.9197}\hfill&&\hfill\mbox{0.9014}\hfill&&\hfill\mbox{0.9167}\hfill&&\hfill\mbox{0.9200}\hfill&&\hfill\mbox{0.9295}\hfill&&\hfill\mbox{0.9220}\hfill&\IEEEeqnarraystrutsizeadd{0pt}{2pt}\\
\hline
&\hfill\raisebox{-15pt}[0pt][0pt]{\mbox{NCSR \cite{Dong13nonlocally}}}\hfill&&\IEEEeqnarraymulticol{23}{v}{}%
\IEEEeqnarraystrutsize{0pt}{0pt}\\
&&&&\hfill\mbox{29.73}\hfill&&\hfill\mbox{31.04}\hfill&&\hfill\mbox{28.61}\hfill&&\hfill\mbox{34.26}\hfill&&\hfill\mbox{31.98}\hfill&&\hfill\mbox{29.95}\hfill&&\hfill\mbox{\textbf{28.07}}\hfill&&\hfill\mbox{\textbf{30.29}}\hfill&&\hfill\mbox{29.62}\hfill&&\hfill\mbox{30.01}\hfill&&\hfill\mbox{30.36}\hfill&\IEEEeqnarraystrutsizeadd{0pt}{2pt}\\
&&&&\hfill\mbox{0.9277}\hfill&&\hfill\mbox{0.9294}\hfill&&\hfill\mbox{0.9021}\hfill&&\hfill\mbox{\textbf{0.9409}}\hfill&&\hfill\mbox{0.9412}\hfill&&\hfill\mbox{0.9252}\hfill&&\hfill\mbox{\textbf{0.9113}}\hfill&&\hfill\mbox{\textbf{0.9274}}\hfill&&\hfill\mbox{0.9215}\hfill&&\hfill\mbox{0.9329}\hfill&&\hfill\mbox{0.9260}\hfill&\IEEEeqnarraystrutsizeadd{0pt}{2pt}\\
%
%\hline
%&\hfill\raisebox{-15pt}[0pt][0pt]{\mbox{NCSR Sub}}\hfill&&\IEEEeqnarraymulticol{23}{v}{}%
%\IEEEeqnarraystrutsize{0pt}{0pt}\\
%&&&&\hfill\mbox{29.80}\hfill&&\hfill\mbox{31.10}\hfill&&\hfill\mbox{28.62}\hfill&&\hfill\mbox{34.30}\hfill&&\hfill\mbox{31.93}\hfill&&\hfill\mbox{\textbf{30.02}}\hfill&&\hfill\mbox{27.85}\hfill&&\hfill\mbox{30.15}\hfill&&\hfill\mbox{\textbf{29.80}}\hfill&&\hfill\mbox{30.07}\hfill&&\hfill\mbox{\textbf{30.36}}\hfill&\IEEEeqnarraystrutsizeadd{0pt}{2pt}\\
%&&&&\hfill\mbox{0.9298}\hfill&&\hfill\mbox{0.9313}\hfill&&\hfill\mbox{0.9042}\hfill&&\hfill\mbox{0.9406}\hfill&&\hfill\mbox{0.9404}\hfill&&\hfill\mbox{0.9258}\hfill&&\hfill\mbox{0.9058}\hfill&&\hfill\mbox{0.9258}\hfill&&\hfill\mbox{0.9245}\hfill&&\hfill\mbox{0.9348}\hfill&&\hfill\mbox{0.9263}\hfill&\IEEEeqnarraystrutsizeadd{0pt}{2pt}\\
%
\hline
&\hfill\raisebox{-15pt}[0pt][0pt]{\mbox{NCSR-GOC}}\hfill&&\IEEEeqnarraymulticol{23}{v}{}%
\IEEEeqnarraystrutsize{0pt}{0pt}\\
&&&&\hfill\mbox{\textbf{29.98}}\hfill&&\hfill\mbox{31.03}\hfill&&\hfill\mbox{\textbf{28.67}}\hfill&&\hfill\mbox{34.31}\hfill&&\hfill\mbox{\textbf{32.06}}\hfill&&\hfill\mbox{\textbf{30.04}}\hfill&&\hfill\mbox{27.92}\hfill&&\hfill\mbox{30.18}\hfill&&\hfill\mbox{\textbf{29.84}}\hfill&&\hfill\mbox{\textbf{30.29}}\hfill&&\hfill\mbox{\textbf{30.43}}\hfill&\IEEEeqnarraystrutsizeadd{0pt}{2pt}\\
&&&&\hfill\mbox{\textbf{0.9332}}\hfill&&\hfill\mbox{\textbf{0.9316}}\hfill&&\hfill\mbox{\textbf{0.9059}}\hfill&&\hfill\mbox{0.9396}\hfill&&\hfill\mbox{\textbf{0.9414}}\hfill&&\hfill\mbox{0.9254}\hfill&&\hfill\mbox{0.9071}\hfill&&\hfill\mbox{0.9260}\hfill&&\hfill\mbox{\textbf{0.9251}}\hfill&&\hfill\mbox{\textbf{0.9371}}\hfill&&\hfill\mbox{\textbf{0.9272}}\hfill&\IEEEeqnarraystrutsizeadd{0pt}{2pt}\\
\hline
&\hfill\raisebox{-15pt}[0pt][0pt]{\mbox{}}\hfill&&\IEEEeqnarraymulticol{2}{v}{}%
\IEEEeqnarraystrutsize{0pt}{0pt}\\
&&&&\hfill\mbox{\textbf{}}\hfill&&\hfill\mbox{}\hfill&&\hfill\mbox{\textbf{}}\hfill&&\hfill\mbox{}\hfill&&\hfill\mbox{}\hfill&&\hfill\mbox{Gaussian}\hfill&&\hfill\mbox{}\hfill&&\hfill\mbox{}\hfill&&\hfill\mbox{\textbf{}}\hfill&&\hfill\mbox{\textbf{}}\hfill&&\hfill\mbox{\textbf{}}\hfill&\IEEEeqnarraystrutsizeadd{0pt}{2pt}\\
%&&&&\hfill\mbox{\textbf{HHHHHHH}}\hfill&&\hfill\mbox{\textbf{jjjj}}\hfill&\IEEEeqnarraystrutsizeadd{0pt}{2pt}\\
%
\hline
&\hfill\raisebox{-15pt}[0pt][0pt]{\mbox{FISTA \cite{Portilla09image}}}\hfill&&\IEEEeqnarraymulticol{23}{v}{}%
\IEEEeqnarraystrutsize{0pt}{0pt}\\
&&&&\hfill\mbox{30.36}\hfill&&\hfill\mbox{29.36}\hfill&&\hfill\mbox{26.81}\hfill&&\hfill\mbox{31.50}\hfill&&\hfill\mbox{31.23}\hfill&&\hfill\mbox{29.47}\hfill&&\hfill\mbox{25.03}\hfill&&\hfill\mbox{29.65}\hfill&&\hfill\mbox{29.42}\hfill&&\hfill\mbox{29.36}\hfill&&\hfill\mbox{29.22}\hfill&\IEEEeqnarraystrutsizeadd{0pt}{2pt}\\
&&&&\hfill\mbox{0.9452}\hfill&&\hfill\mbox{0.9024}\hfill&&\hfill\mbox{0.8845}\hfill&&\hfill\mbox{0.8968}\hfill&&\hfill\mbox{0.9290}\hfill&&\hfill\mbox{0.9011}\hfill&&\hfill\mbox{0.8415}\hfill&&\hfill\mbox{0.9256}\hfill&&\hfill\mbox{0.9057}\hfill&&\hfill\mbox{0.9393}\hfill&&\hfill\mbox{0.9071}\hfill&\IEEEeqnarraystrutsizeadd{0pt}{2pt}\\
\hline
&\hfill\raisebox{-15pt}[0pt][0pt]{\mbox{ASDS \cite{Dong11image}}}\hfill&&\IEEEeqnarraymulticol{23}{v}{}%
\IEEEeqnarraystrutsize{0pt}{0pt}\\
&&&&\hfill\mbox{29.83}\hfill&&\hfill\mbox{30.27}\hfill&&\hfill\mbox{27.29}\hfill&&\hfill\mbox{31.87}\hfill&&\hfill\mbox{32.93}\hfill&&\hfill\mbox{30.36}\hfill&&\hfill\mbox{27.05}\hfill&&\hfill\mbox{31.91}\hfill&&\hfill\mbox{28.95}\hfill&&\hfill\mbox{30.62}\hfill&&\hfill\mbox{30.11}\hfill&\IEEEeqnarraystrutsizeadd{0pt}{2pt}\\
&&&&\hfill\mbox{0.9126}\hfill&&\hfill\mbox{0.9064}\hfill&&\hfill\mbox{0.8637}\hfill&&\hfill\mbox{0.8978}\hfill&&\hfill\mbox{0.9576}\hfill&&\hfill\mbox{0.9058}\hfill&&\hfill\mbox{0.8881}\hfill&&\hfill\mbox{0.9491}\hfill&&\hfill\mbox{0.9039}\hfill&&\hfill\mbox{0.9304}\hfill&&\hfill\mbox{0.9115}\hfill&\IEEEeqnarraystrutsizeadd{0pt}{2pt}\\
\hline
&\hfill\raisebox{-15pt}[0pt][0pt]{\mbox{IDD-BM3D \cite{Danielyan12bm3d}}}\hfill&&\IEEEeqnarraymulticol{23}{v}{}%
\IEEEeqnarraystrutsize{0pt}{0pt}\\
&&&&\hfill\mbox{30.73}\hfill&&\hfill\mbox{\textbf{31.68}}\hfill&&\hfill\mbox{28.17}\hfill&&\hfill\mbox{\textbf{34.08}}\hfill&&\hfill\mbox{32.89}\hfill&&\hfill\mbox{\textbf{31.45}}\hfill&&\hfill\mbox{27.19}\hfill&&\hfill\mbox{31.66}\hfill&&\hfill\mbox{29.99}\hfill&&\hfill\mbox{31.40}\hfill&&\hfill\mbox{30.92}\hfill&\IEEEeqnarraystrutsizeadd{0pt}{2pt}\\
&&&&\hfill\mbox{0.9442}\hfill&&\hfill\mbox{\textbf{0.9426}}\hfill&&\hfill\mbox{0.9136}\hfill&&\hfill\mbox{0.9359}\hfill&&\hfill\mbox{0.9561}\hfill&&\hfill\mbox{\textbf{0.9430}}\hfill&&\hfill\mbox{0.8986}\hfill&&\hfill\mbox{0.9496}\hfill&&\hfill\mbox{0.9373}\hfill&&\hfill\mbox{0.9512}\hfill&&\hfill\mbox{0.9372}\hfill&\IEEEeqnarraystrutsizeadd{0pt}{2pt}\\
\hline
&\hfill\raisebox{-15pt}[0pt][0pt]{\mbox{NCSR \cite{Dong13nonlocally}}}\hfill&&\IEEEeqnarraymulticol{23}{v}{}%
\IEEEeqnarraystrutsize{0pt}{0pt}\\
&&&&\hfill\mbox{30.84}\hfill&&\hfill\mbox{31.37}\hfill&&\hfill\mbox{28.27}\hfill&&\hfill\mbox{33.69}\hfill&&\hfill\mbox{33.40}\hfill&&\hfill\mbox{31.17}\hfill&&\hfill\mbox{\textbf{28.02}}\hfill&&\hfill\mbox{32.23}\hfill&&\hfill\mbox{30.01}\hfill&&\hfill\mbox{31.62}\hfill&&\hfill\mbox{31.06}\hfill&\IEEEeqnarraystrutsizeadd{0pt}{2pt}\\
&&&&\hfill\mbox{0.9379}\hfill&&\hfill\mbox{0.9348}\hfill&&\hfill\mbox{0.9044}\hfill&&\hfill\mbox{0.9339}\hfill&&\hfill\mbox{0.9589}\hfill&&\hfill\mbox{0.9360}\hfill&&\hfill\mbox{\textbf{0.9108}}\hfill&&\hfill\mbox{0.9533}\hfill&&\hfill\mbox{0.9300}\hfill&&\hfill\mbox{0.9514}\hfill&&\hfill\mbox{0.9351}\hfill&\IEEEeqnarraystrutsizeadd{0pt}{2pt}\\
%
%\hline
%&\hfill\raisebox{-15pt}[0pt][0pt]{\mbox{NCSR Sub}}\hfill&&\IEEEeqnarraymulticol{23}{v}{}%
%\IEEEeqnarraystrutsize{0pt}{0pt}\\
%&&&&\hfill\mbox{31.20}\hfill&&\hfill\mbox{31.50}\hfill&&\hfill\mbox{\textbf{28.49}}\hfill&&\hfill\mbox{33.81}\hfill&&\hfill\mbox{33.43}\hfill&&\hfill\mbox{31.30}\hfill&&\hfill\mbox{27.30}\hfill&&\hfill\mbox{\textbf{32.36}}\hfill&&\hfill\mbox{30.25}\hfill&&\hfill\mbox{31.79}\hfill&&\hfill\mbox{31.14}\hfill&\IEEEeqnarraystrutsizeadd{0pt}{2pt}\\
%&&&&\hfill\mbox{0.9465}\hfill&&\hfill\mbox{0.9409}\hfill&&\hfill\mbox{0.9150}\hfill&&\hfill\mbox{\textbf{0.9377}}\hfill&&\hfill\mbox{0.9593}\hfill&&\hfill\mbox{0.9425}\hfill&&\hfill\mbox{0.8991}\hfill&&\hfill\mbox{\textbf{0.9555}}\hfill&&\hfill\mbox{0.9388}\hfill&&\hfill\mbox{0.9554}\hfill&&\hfill\mbox{0.9391}\hfill&\IEEEeqnarraystrutsizeadd{0pt}{2pt}\\
%
\hline
&\hfill\raisebox{-15pt}[0pt][0pt]{\mbox{NCSR-GOC}}\hfill&&\IEEEeqnarraymulticol{23}{v}{}%
\IEEEeqnarraystrutsize{0pt}{0pt}\\
&&&&\hfill\mbox{\textbf{31.32}}\hfill&&\hfill\mbox{31.48}\hfill&&\hfill\mbox{\textbf{28.44}}\hfill&&\hfill\mbox{33.80}\hfill&&\hfill\mbox{\textbf{33.45}}\hfill&&\hfill\mbox{31.28}\hfill&&\hfill\mbox{27.45}\hfill&&\hfill\mbox{\textbf{32.27}}\hfill&&\hfill\mbox{\textbf{30.27}}\hfill&&\hfill\mbox{\textbf{32.04}}\hfill&&\hfill\mbox{\textbf{31.18}}\hfill&\IEEEeqnarraystrutsizeadd{0pt}{2pt}\\
&&&&\hfill\mbox{\textbf{0.9486}}\hfill&&\hfill\mbox{0.9413}\hfill&&\hfill\mbox{\textbf{0.9153}}\hfill&&\hfill\mbox{\textbf{0.9375}}\hfill&&\hfill\mbox{\textbf{0.9594}}\hfill&&\hfill\mbox{0.9429}\hfill&&\hfill\mbox{0.9014}\hfill&&\hfill\mbox{\textbf{0.9554}}\hfill&&\hfill\mbox{\textbf{0.9389}}\hfill&&\hfill\mbox{\textbf{0.9587}}\hfill&&\hfill\mbox{\textbf{0.9399}}\hfill&\IEEEeqnarraystrutsizeadd{0pt}{2pt}\\
\IEEEeqnarraydblrulerowcut\\
\end{IEEEeqnarraybox}
\end{table*}

\subsection{Image denoising}
\label{ssec:exp_denoise}

\begin{figure}[!t]
\begin{center}
\setcounter{subfigure}{0}   
\subfigure{\includegraphics[width=0.15\linewidth]{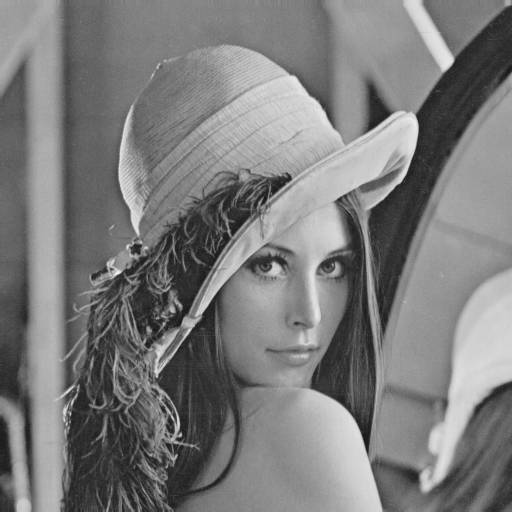}}   \subfigure{\includegraphics[width=0.15\linewidth]{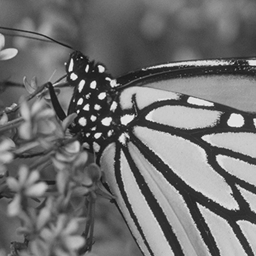}}
   \subfigure{\includegraphics[width=0.15\linewidth]{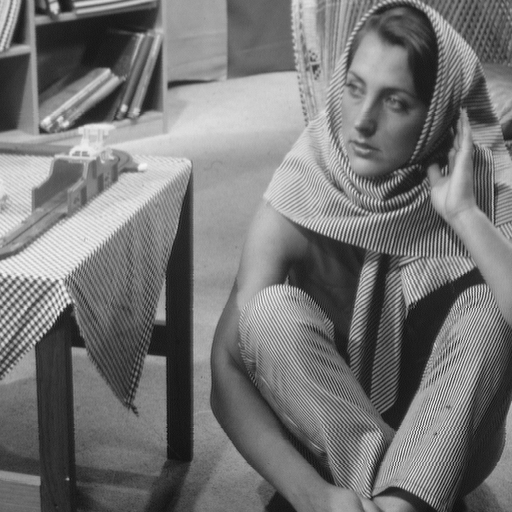}}
   \subfigure{\includegraphics[width=0.15\linewidth]{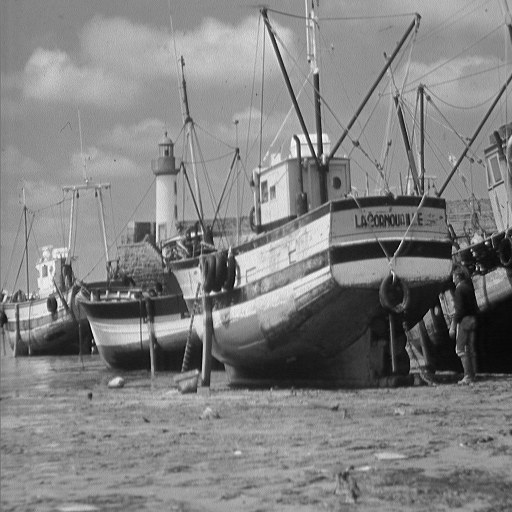}}
   \subfigure{\includegraphics[width=0.15\linewidth]{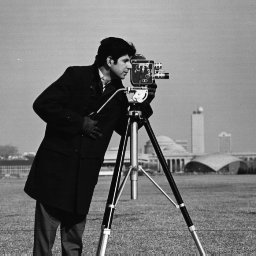}}
   \subfigure{\includegraphics[width=0.15\linewidth]{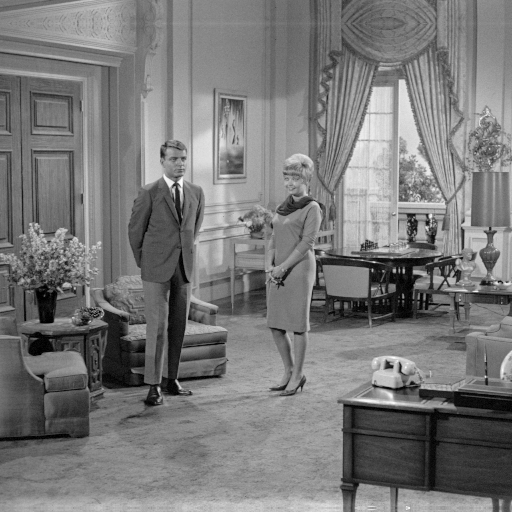}} \\ 
\setcounter{subfigure}{0}   
\subfigure{\includegraphics[width=0.15\linewidth]{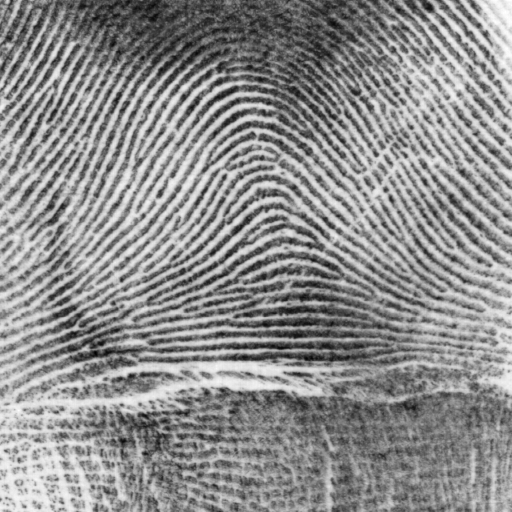}}   \subfigure{\includegraphics[width=0.15\linewidth]{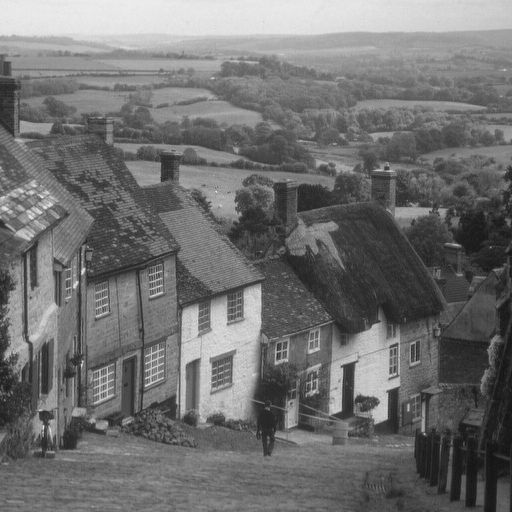}}  
\subfigure{\includegraphics[width=0.15\linewidth]{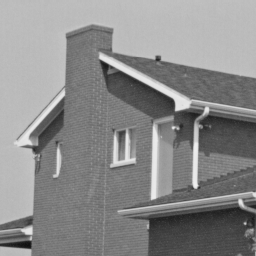}}   
\subfigure{\includegraphics[width=0.15\linewidth]{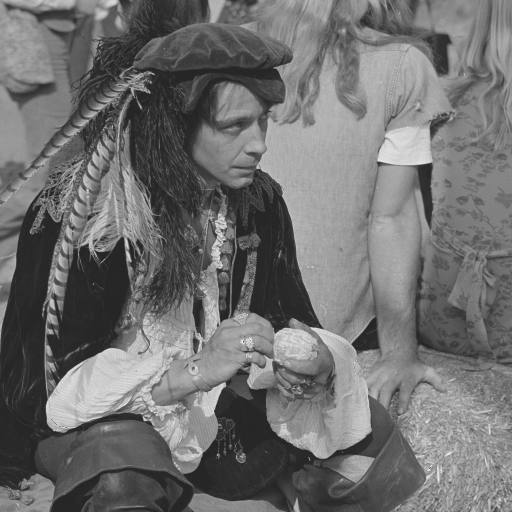}} 
\subfigure{\includegraphics[width=0.15\linewidth]{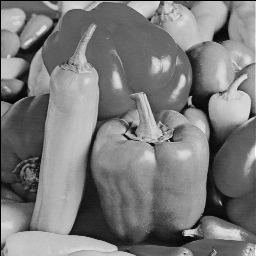}}     
\subfigure{\includegraphics[width=0.15\linewidth]{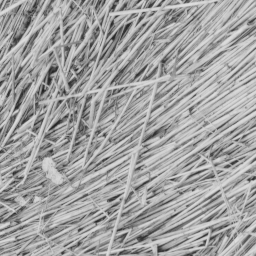}}    
\end{center}
  \caption{Test images for denoising: Lena, Monarch, Barbara, Boat, Cameraman (C. Man), Couple, Fingerprint (F. Print), Hill, House, Man, Peppers, Straw.}
\label{fig:img_used_denois}
\end{figure}

We now evaluate our method in the image denoising application. Since the deformation of the patch manifold geometry due to noise poses a challenge on geometry-based similarity assessment between patches, we use the AGNN method in the experiments in this section, which usually has a better reconstruction quality than GOC. We compare AGNN with K-means within the framework of the NCSR method \cite{Dong13nonlocally}. The algorithms are tested on the images shown in Figure \ref{fig:img_used_denois}. The images are corrupted with additive white Gaussian noise at different noise levels with standard deviation $\sigma = \left[5\;\; 10\;\; 15\;\; 20 \;\; 50\;\; 100\right]$. The parameters of AGNN are set as  $\sRGC=35$ (number of nearest neighbors in the diffusion stage of RGC \cite{Donoser13replicator}), $\kappa=2$ (number of iterations for diffusing the affinities),  $c_1=10$ (Gaussian kernel scale), and $c_2=0.9$ (affinity threshold). All the general parameters for NCSR are selected as Dong et al.~\cite{Dong13nonlocally} in order to maintain the consistency in the comparison.

%SAPCA-BM3D \cite{Katkovnik10from}
The PSNR measures of the reconstruction qualities are presented in Table \ref{tbl:results_denoisingAGNN}. The results obtained with the image denoising algorithms SAPCA-BM3D \cite{Katkovnik10from}; LSSC \cite{Mairal09non}; EPLL \cite{Zoran11from}; NCSR \cite{Dong13nonlocally} reported in \cite{Dong13nonlocally} for the same experiments are also given for the purpose of comparison. The overall performances of all algorithms are observed to be quite close, and the best average PSNR is given by SAPCA-BM3D at most noise levels. Nevertheless, the comparison between NCSR and NCSR-AGNN is more interesting, which shows that the proposed NCSR-AGNN algorithm yields a very similar performance to NCSR in denoising. A very slight improvement in average PSNR is obtained over NCSR at small noise levels, while this small advantage is lost at large noise levels. %One of the main reasons for this can be the fact that, contrary to the applications of superresolution and deblurring where the patch manifold has a regular geometric structure, the geometry of the patch manifold is deformed due to the noise in denoising. Especially as the noise level gets higher, the estimates of nearest neighborhoods may get affected significantly. Moreover, the observation that the performance gain is marginal suggests that the main factor that leads to an efficient denoising is the distinction of the basis vectors corresponding to noise from the basis vectors corresponding to the actual image content in the locally computed PCA bases. Calculating the PCA basis in a local neighborhood with respect to the Euclidean distance already achieves this purpose, and the advantage obtained by taking the manifold structure into account when forming the neighborhood contributes little to the overall reconstruction quality in denoising. 
One can observe that the performance of NCSR-AGNN is better on the Monarch and Fingerprint images. This may be an indication that in such images with strong and oscillatory high-frequency textures, the patch manifold must have a particular geometry that is easier to identify under noise and the consideration of the geometry in assigning the similarities may help improve the denoising performance.

%elimination of the noise in a small local neighborhood, in this case by eliminating the noise components of the PCA basis computed in a local Euclidean-distance neighborhood. The slight gain over NCSR achieved with NCSR-AGNN seems to indicate that the advantage obtained by taking the manifold structure into account is only secondary.

%Maybe explain why we cannot merge our method with SAPCA-BM3D or BM3D(A. Foi's remark: \textit{However, based on my intuition of your GOC algorithm, it does not seem straightforward to merge this with BM3D-SAPCA or BM3D, mostly because of the way indexing is handled, at least within our own implementations.} )}

%%%%%% FIGURE: USED IMAGES IN DENOISING

\begin{table*}[!t]%use {table*} to use 1 column
\centering
\caption{PSNR (in $\mathrm{d}\mathrm{B}$) results for the luminance components of denoised images for different denoising algorithms are reported in the following order: SAPCA-BM3D \cite{Katkovnik10from}; LSSC \cite{Mairal09non}; EPLL \cite{Zoran11from}; NCSR \cite{Dong13nonlocally}; and NCSR with proposed AGNN. }
\label{tbl:results_denoisingAGNN}
\begin{IEEEeqnarraybox}[\IEEEeqnarraystrutmode\IEEEeqnarraystrutsizeadd{2pt}{0pt}]{x/r/vx/r/x/r/x/r/x/r/x/r/x/r/x/r/x/r/x/r/x/r/x/r/x/r/x/r/x/r/x}
\IEEEeqnarraydblrulerowcut\\
&\hfill\raisebox{-8pt}[0pt][0pt]{\mbox{}}\hfill&&\IEEEeqnarraymulticol{29}{v}{}%
\IEEEeqnarraystrutsize{0pt}{0pt}\\
&&&&\hfill\mbox{Methods}\hfill&&\hfill\mbox{Lena}\hfill&&\hfill\mbox{Monarch}\hfill&&\hfill\mbox{Barbara}\hfill&&\hfill\mbox{Boat}\hfill&&\hfill\mbox{C. Man}\hfill&&\hfill\mbox{Couple}\hfill&&\hfill\mbox{F. Print}\hfill&&\hfill\mbox{Hill}\hfill&&\hfill\mbox{House}\hfill&&\hfill\mbox{Man}\hfill&&\hfill\mbox{Peppers}\hfill&&\hfill\mbox{Straw}\hfill&&\hfill\mbox{Average}\hfill&\IEEEeqnarraystrutsizeadd{0pt}{2pt}\\
\IEEEeqnarraydblrulerowcut\\
&\hfill\raisebox{-33pt}[0pt][0pt]{$\sigma=5$}\hfill&&\IEEEeqnarraymulticol{29}{v}{}%
\IEEEeqnarraystrutsize{0pt}{0pt}\\
&&&&\hfill\mbox{SAPCA-BM3D}\hfill&&\hfill\mbox{\textbf{38.86}}\hfill&&\hfill\mbox{\textbf{38.69}}\hfill&&\hfill\mbox{38.38}\hfill&&\hfill\mbox{\textbf{37.50}}\hfill&&\hfill\mbox{\textbf{38.54}}\hfill&&\hfill\mbox{\textbf{37.60}}\hfill&&\hfill\mbox{36.67}\hfill&&\hfill\mbox{\textbf{37.31}}\hfill&&\hfill\mbox{\textbf{40.13}}\hfill&&\hfill\mbox{\textbf{37.99}}\hfill&&\hfill\mbox{\textbf{38.30}}\hfill&&\hfill\mbox{35.81}\hfill&&\hfill\mbox{\textbf{37.98}}\hfill&\IEEEeqnarraystrutsizeadd{0pt}{2pt}\\
&&&&\hfill\mbox{LSSC}\hfill&&\hfill\mbox{38.68}\hfill&&\hfill\mbox{38.53}\hfill&&\hfill\mbox{\textbf{38.44}}\hfill&&\hfill\mbox{37.34}\hfill&&\hfill\mbox{38.24}\hfill&&\hfill\mbox{37.41}\hfill&&\hfill\mbox{36.71}\hfill&&\hfill\mbox{37.16}\hfill&&\hfill\mbox{40.00}\hfill&&\hfill\mbox{37.84}\hfill&&\hfill\mbox{38.15}\hfill&&\hfill\mbox{\textbf{35.92}}\hfill&&\hfill\mbox{37.87}\hfill&\IEEEeqnarraystrutsizeadd{0pt}{2pt}\\
&&&&\hfill\mbox{EPLL}\hfill&&\hfill\mbox{38.52}\hfill&&\hfill\mbox{38.22}\hfill&&\hfill\mbox{37.56}\hfill&&\hfill\mbox{36.78}\hfill&&\hfill\mbox{38.04}\hfill&&\hfill\mbox{37.32}\hfill&&\hfill\mbox{36.41}\hfill&&\hfill\mbox{37.00}\hfill&&\hfill\mbox{39.04}\hfill&&\hfill\mbox{37.67}\hfill&&\hfill\mbox{37.93}\hfill&&\hfill\mbox{35.36}\hfill&&\hfill\mbox{37.49}\hfill&\IEEEeqnarraystrutsizeadd{0pt}{2pt}\\
&&&&\hfill\mbox{NCSR}\hfill&&\hfill\mbox{38.70}\hfill&&\hfill\mbox{38.49}\hfill&&\hfill\mbox{38.36}\hfill&&\hfill\mbox{37.35}\hfill&&\hfill\mbox{38.17}\hfill&&\hfill\mbox{37.44}\hfill&&\hfill\mbox{36.81}\hfill&&\hfill\mbox{37.17}\hfill&&\hfill\mbox{39.91}\hfill&&\hfill\mbox{37.78}\hfill&&\hfill\mbox{38.06}\hfill&&\hfill\mbox{35.87}\hfill&&\hfill\mbox{37.84}\hfill&\IEEEeqnarraystrutsizeadd{0pt}{2pt}\\
&&&&\hfill\mbox{NCSR-AGNN}\hfill&&\hfill\mbox{38.74}\hfill&&\hfill\mbox{38.62}\hfill&&\hfill\mbox{38.32}\hfill&&\hfill\mbox{37.34}\hfill&&\hfill\mbox{38.19}\hfill&&\hfill\mbox{37.40}\hfill&&\hfill\mbox{\textbf{36.86}}\hfill&&\hfill\mbox{37.15}\hfill&&\hfill\mbox{40.06}\hfill&&\hfill\mbox{37.78}\hfill&&\hfill\mbox{38.09}\hfill&&\hfill\mbox{35.82}\hfill&&\hfill\mbox{37.86}\hfill&\IEEEeqnarraystrutsizeadd{0pt}{2pt}\\
\hline
&\hfill\raisebox{-33pt}[0pt][0pt]{$\sigma=10$}\hfill&&\IEEEeqnarraymulticol{29}{v}{}%
\IEEEeqnarraystrutsize{0pt}{0pt}\\
&&&&\hfill\mbox{SAPCA-BM3D}\hfill&&\hfill\mbox{\textbf{36.07}}\hfill&&\hfill\mbox{\textbf{34.74}}\hfill&&\hfill\mbox{\textbf{35.07}}\hfill&&\hfill\mbox{\textbf{34.10}}\hfill&&\hfill\mbox{\textbf{34.52}}\hfill&&\hfill\mbox{\textbf{34.13}}\hfill&&\hfill\mbox{32.65}\hfill&&\hfill\mbox{\textbf{33.84}}\hfill&&\hfill\mbox{\textbf{37.06}}\hfill&&\hfill\mbox{\textbf{34.18}}\hfill&&\hfill\mbox{\textbf{34.94}}\hfill&&\hfill\mbox{31.46}\hfill&&\hfill\mbox{\textbf{34.40}}\hfill&\IEEEeqnarraystrutsizeadd{0pt}{2pt}\\
&&&&\hfill\mbox{LSSC}\hfill&&\hfill\mbox{35.83}\hfill&&\hfill\mbox{34.48}\hfill&&\hfill\mbox{34.95}\hfill&&\hfill\mbox{33.99}\hfill&&\hfill\mbox{34.14}\hfill&&\hfill\mbox{33.96}\hfill&&\hfill\mbox{32.57}\hfill&&\hfill\mbox{33.68}\hfill&&\hfill\mbox{37.05}\hfill&&\hfill\mbox{34.03}\hfill&&\hfill\mbox{34.80}\hfill&&\hfill\mbox{31.39}\hfill&&\hfill\mbox{34.24}\hfill&\IEEEeqnarraystrutsizeadd{0pt}{2pt}\\
&&&&\hfill\mbox{EPLL}\hfill&&\hfill\mbox{35.56}\hfill&&\hfill\mbox{34.27}\hfill&&\hfill\mbox{33.59}\hfill&&\hfill\mbox{33.63}\hfill&&\hfill\mbox{33.94}\hfill&&\hfill\mbox{33.78}\hfill&&\hfill\mbox{32.13}\hfill&&\hfill\mbox{33.49}\hfill&&\hfill\mbox{35.81}\hfill&&\hfill\mbox{33.90}\hfill&&\hfill\mbox{34.51}\hfill&&\hfill\mbox{30.84}\hfill&&\hfill\mbox{33.79}\hfill&\IEEEeqnarraystrutsizeadd{0pt}{2pt}\\
&&&&\hfill\mbox{NCSR}\hfill&&\hfill\mbox{35.81}\hfill&&\hfill\mbox{34.57}\hfill&&\hfill\mbox{34.98}\hfill&&\hfill\mbox{33.90}\hfill&&\hfill\mbox{34.12}\hfill&&\hfill\mbox{33.94}\hfill&&\hfill\mbox{32.70}\hfill&&\hfill\mbox{33.69}\hfill&&\hfill\mbox{36.80}\hfill&&\hfill\mbox{33.96}\hfill&&\hfill\mbox{34.66}\hfill&&\hfill\mbox{\textbf{31.50}}\hfill&&\hfill\mbox{34.22}\hfill&\IEEEeqnarraystrutsizeadd{0pt}{2pt}\\
&&&&\hfill\mbox{NCSR-AGNN}\hfill&&\hfill\mbox{35.84}\hfill&&\hfill\mbox{34.66}\hfill&&\hfill\mbox{34.94}\hfill&&\hfill\mbox{33.87}\hfill&&\hfill\mbox{34.13}\hfill&&\hfill\mbox{33.90}\hfill&&\hfill\mbox{\textbf{32.72}}\hfill&&\hfill\mbox{33.66}\hfill&&\hfill\mbox{36.87}\hfill&&\hfill\mbox{33.95}\hfill&&\hfill\mbox{34.69}\hfill&&\hfill\mbox{31.46}\hfill&&\hfill\mbox{34.22}\hfill&\IEEEeqnarraystrutsizeadd{0pt}{2pt}\\
\hline
&\hfill\raisebox{-33pt}[0pt][0pt]{$\sigma=15$}\hfill&&\IEEEeqnarraymulticol{29}{v}{}%
\IEEEeqnarraystrutsize{0pt}{0pt}\\
&&&&\hfill\mbox{SAPCA-BM3D}\hfill&&\hfill\mbox{\textbf{34.43}}\hfill&&\hfill\mbox{\textbf{32.46}}\hfill&&\hfill\mbox{\textbf{33.27}}\hfill&&\hfill\mbox{\textbf{32.29}}\hfill&&\hfill\mbox{\textbf{32.31}}\hfill&&\hfill\mbox{\textbf{32.20}}\hfill&&\hfill\mbox{30.46}\hfill&&\hfill\mbox{\textbf{32.06}}\hfill&&\hfill\mbox{35.31}\hfill&&\hfill\mbox{\textbf{32.12}}\hfill&&\hfill\mbox{\textbf{33.01}}\hfill&&\hfill\mbox{29.13}\hfill&&\hfill\mbox{\textbf{32.42}}\hfill&\IEEEeqnarraystrutsizeadd{0pt}{2pt}\\
&&&&\hfill\mbox{LSSC}\hfill&&\hfill\mbox{34.14}\hfill&&\hfill\mbox{32.15}\hfill&&\hfill\mbox{32.96}\hfill&&\hfill\mbox{32.17}\hfill&&\hfill\mbox{31.96}\hfill&&\hfill\mbox{32.06}\hfill&&\hfill\mbox{30.31}\hfill&&\hfill\mbox{31.89}\hfill&&\hfill\mbox{\textbf{35.32}}\hfill&&\hfill\mbox{31.98}\hfill&&\hfill\mbox{32.87}\hfill&&\hfill\mbox{28.95}\hfill&&\hfill\mbox{32.23}\hfill&\IEEEeqnarraystrutsizeadd{0pt}{2pt}\\
&&&&\hfill\mbox{EPLL}\hfill&&\hfill\mbox{33.85}\hfill&&\hfill\mbox{32.04}\hfill&&\hfill\mbox{31.33}\hfill&&\hfill\mbox{31.89}\hfill&&\hfill\mbox{31.73}\hfill&&\hfill\mbox{31.83}\hfill&&\hfill\mbox{29.83}\hfill&&\hfill\mbox{31.67}\hfill&&\hfill\mbox{34.21}\hfill&&\hfill\mbox{31.89}\hfill&&\hfill\mbox{32.56}\hfill&&\hfill\mbox{28.50}\hfill&&\hfill\mbox{31.78}\hfill&\IEEEeqnarraystrutsizeadd{0pt}{2pt}\\
&&&&\hfill\mbox{NCSR}\hfill&&\hfill\mbox{34.09}\hfill&&\hfill\mbox{32.34}\hfill&&\hfill\mbox{33.02}\hfill&&\hfill\mbox{32.03}\hfill&&\hfill\mbox{31.99}\hfill&&\hfill\mbox{31.95}\hfill&&\hfill\mbox{30.46}\hfill&&\hfill\mbox{31.86}\hfill&&\hfill\mbox{35.11}\hfill&&\hfill\mbox{31.89}\hfill&&\hfill\mbox{32.70}\hfill&&\hfill\mbox{29.13}\hfill&&\hfill\mbox{32.21}\hfill&\IEEEeqnarraystrutsizeadd{0pt}{2pt}\\
&&&&\hfill\mbox{NCSR-AGNN}\hfill&&\hfill\mbox{34.11}\hfill&&\hfill\mbox{32.37}\hfill&&\hfill\mbox{32.98}\hfill&&\hfill\mbox{32.01}\hfill&&\hfill\mbox{32.00}\hfill&&\hfill\mbox{31.94}\hfill&&\hfill\mbox{\textbf{30.47}}\hfill&&\hfill\mbox{31.84}\hfill&&\hfill\mbox{35.14}\hfill&&\hfill\mbox{31.88}\hfill&&\hfill\mbox{32.73}\hfill&&\hfill\mbox{\textbf{29.14}}\hfill&&\hfill\mbox{32.22}\hfill&\IEEEeqnarraystrutsizeadd{0pt}{2pt}\\
\hline
&\hfill\raisebox{-33pt}[0pt][0pt]{$\sigma=20$}\hfill&&\IEEEeqnarraymulticol{29}{v}{}%
\IEEEeqnarraystrutsize{0pt}{0pt}\\
&&&&\hfill\mbox{SAPCA-BM3D}\hfill&&\hfill\mbox{\textbf{33.20}}\hfill&&\hfill\mbox{\textbf{30.92}}\hfill&&\hfill\mbox{\textbf{31.97}}\hfill&&\hfill\mbox{\textbf{31.02}}\hfill&&\hfill\mbox{\textbf{30.86}}\hfill&&\hfill\mbox{\textbf{30.83}}\hfill&&\hfill\mbox{28.97}\hfill&&\hfill\mbox{\textbf{30.85}}\hfill&&\hfill\mbox{34.03}\hfill&&\hfill\mbox{\textbf{30.73}}\hfill&&\hfill\mbox{\textbf{31.61}}\hfill&&\hfill\mbox{\textbf{27.52}}\hfill&&\hfill\mbox{\textbf{31.04}}\hfill&\IEEEeqnarraystrutsizeadd{0pt}{2pt}\\
&&&&\hfill\mbox{LSSC}\hfill&&\hfill\mbox{32.88}\hfill&&\hfill\mbox{30.58}\hfill&&\hfill\mbox{31.53}\hfill&&\hfill\mbox{30.87}\hfill&&\hfill\mbox{30.54}\hfill&&\hfill\mbox{30.70}\hfill&&\hfill\mbox{28.78}\hfill&&\hfill\mbox{30.71}\hfill&&\hfill\mbox{\textbf{34.16}}\hfill&&\hfill\mbox{30.61}\hfill&&\hfill\mbox{31.47}\hfill&&\hfill\mbox{27.36}\hfill&&\hfill\mbox{30.85}\hfill&\IEEEeqnarraystrutsizeadd{0pt}{2pt}\\
&&&&\hfill\mbox{EPLL}\hfill&&\hfill\mbox{32.60}\hfill&&\hfill\mbox{30.48}\hfill&&\hfill\mbox{29.75}\hfill&&\hfill\mbox{30.63}\hfill&&\hfill\mbox{30.28}\hfill&&\hfill\mbox{30.47}\hfill&&\hfill\mbox{28.29}\hfill&&\hfill\mbox{30.47}\hfill&&\hfill\mbox{33.08}\hfill&&\hfill\mbox{30.53}\hfill&&\hfill\mbox{31.18}\hfill&&\hfill\mbox{26.93}\hfill&&\hfill\mbox{30.39}\hfill&\IEEEeqnarraystrutsizeadd{0pt}{2pt}\\
&&&&\hfill\mbox{NCSR}\hfill&&\hfill\mbox{32.92}\hfill&&\hfill\mbox{30.69}\hfill&&\hfill\mbox{31.72}\hfill&&\hfill\mbox{30.74}\hfill&&\hfill\mbox{30.48}\hfill&&\hfill\mbox{30.56}\hfill&&\hfill\mbox{28.99}\hfill&&\hfill\mbox{30.61}\hfill&&\hfill\mbox{33.97}\hfill&&\hfill\mbox{30.52}\hfill&&\hfill\mbox{31.26}\hfill&&\hfill\mbox{27.50}\hfill&&\hfill\mbox{30.83}\hfill&\IEEEeqnarraystrutsizeadd{0pt}{2pt}\\
&&&&\hfill\mbox{NCSR-AGNN}\hfill&&\hfill\mbox{32.89}\hfill&&\hfill\mbox{30.72}\hfill&&\hfill\mbox{31.70}\hfill&&\hfill\mbox{30.73}\hfill&&\hfill\mbox{30.50}\hfill&&\hfill\mbox{30.55}\hfill&&\hfill\mbox{\textbf{29.01}}\hfill&&\hfill\mbox{30.52}\hfill&&\hfill\mbox{33.98}\hfill&&\hfill\mbox{30.51}\hfill&&\hfill\mbox{31.28}\hfill&&\hfill\mbox{27.50}\hfill&&\hfill\mbox{30.82}\hfill&\IEEEeqnarraystrutsizeadd{0pt}{2pt}\\
\hline
&\hfill\raisebox{-33pt}[0pt][0pt]{$\sigma=50$}\hfill&&\IEEEeqnarraymulticol{29}{v}{}%
\IEEEeqnarraystrutsize{0pt}{0pt}\\
&&&&\hfill\mbox{SAPCA-BM3D}\hfill&&\hfill\mbox{\textbf{29.07}}\hfill&&\hfill\mbox{\textbf{26.28}}\hfill&&\hfill\mbox{\textbf{27.51}}\hfill&&\hfill\mbox{\textbf{26.89}}\hfill&&\hfill\mbox{\textbf{26.59}}\hfill&&\hfill\mbox{\textbf{26.48}}\hfill&&\hfill\mbox{\textbf{24.53}}\hfill&&\hfill\mbox{\textbf{27.13}}\hfill&&\hfill\mbox{29.53}\hfill&&\hfill\mbox{\textbf{26.84}}\hfill&&\hfill\mbox{\textbf{26.94}}\hfill&&\hfill\mbox{\textbf{22.79}}\hfill&&\hfill\mbox{\textbf{26.71}}\hfill&\IEEEeqnarraystrutsizeadd{0pt}{2pt}\\
&&&&\hfill\mbox{LSSC}\hfill&&\hfill\mbox{28.95}\hfill&&\hfill\mbox{25.59}\hfill&&\hfill\mbox{27.13}\hfill&&\hfill\mbox{26.76}\hfill&&\hfill\mbox{26.36}\hfill&&\hfill\mbox{26.31}\hfill&&\hfill\mbox{24.21}\hfill&&\hfill\mbox{26.99}\hfill&&\hfill\mbox{\textbf{29.90}}\hfill&&\hfill\mbox{26.72}\hfill&&\hfill\mbox{26.87}\hfill&&\hfill\mbox{22.67}\hfill&&\hfill\mbox{26.54}\hfill&\IEEEeqnarraystrutsizeadd{0pt}{2pt}\\
&&&&\hfill\mbox{EPLL}\hfill&&\hfill\mbox{28.42}\hfill&&\hfill\mbox{25.67}\hfill&&\hfill\mbox{24.83}\hfill&&\hfill\mbox{26.64}\hfill&&\hfill\mbox{26.08}\hfill&&\hfill\mbox{26.22}\hfill&&\hfill\mbox{23.58}\hfill&&\hfill\mbox{26.91}\hfill&&\hfill\mbox{28.91}\hfill&&\hfill\mbox{26.63}\hfill&&\hfill\mbox{26.60}\hfill&&\hfill\mbox{22.00}\hfill&&\hfill\mbox{26.04}\hfill&\IEEEeqnarraystrutsizeadd{0pt}{2pt}\\
&&&&\hfill\mbox{NCSR}\hfill&&\hfill\mbox{28.89}\hfill&&\hfill\mbox{25.68}\hfill&&\hfill\mbox{27.10}\hfill&&\hfill\mbox{26.60}\hfill&&\hfill\mbox{26.16}\hfill&&\hfill\mbox{26.21}\hfill&&\hfill\mbox{\textbf{24.53}}\hfill&&\hfill\mbox{26.86}\hfill&&\hfill\mbox{29.63}\hfill&&\hfill\mbox{26.60}\hfill&&\hfill\mbox{26.53}\hfill&&\hfill\mbox{22.48}\hfill&&\hfill\mbox{26.44}\hfill&\IEEEeqnarraystrutsizeadd{0pt}{2pt}\\
&&&&\hfill\mbox{NCSR-AGNN}\hfill&&\hfill\mbox{28.90}\hfill&&\hfill\mbox{25.69}\hfill&&\hfill\mbox{27.08}\hfill&&\hfill\mbox{26.57}\hfill&&\hfill\mbox{26.12}\hfill&&\hfill\mbox{26.19}\hfill&&\hfill\mbox{24.50}\hfill&&\hfill\mbox{26.80}\hfill&&\hfill\mbox{29.63}\hfill&&\hfill\mbox{26.59}\hfill&&\hfill\mbox{26.54}\hfill&&\hfill\mbox{22.46}\hfill&&\hfill\mbox{26.42}\hfill&\IEEEeqnarraystrutsizeadd{0pt}{2pt}\\
\hline
&\hfill\raisebox{-33pt}[0pt][0pt]{$\sigma=100$}\hfill&&\IEEEeqnarraymulticol{29}{v}{}%
\IEEEeqnarraystrutsize{0pt}{0pt}\\
&&&&\hfill\mbox{SAPCA-BM3D}\hfill&&\hfill\mbox{25.37}\hfill&&\hfill\mbox{\textbf{22.31}}\hfill&&\hfill\mbox{23.05}\hfill&&\hfill\mbox{23.71}\hfill&&\hfill\mbox{22.91}\hfill&&\hfill\mbox{23.19}\hfill&&\hfill\mbox{21.07}\hfill&&\hfill\mbox{24.10}\hfill&&\hfill\mbox{25.20}\hfill&&\hfill\mbox{23.86}\hfill&&\hfill\mbox{23.05}\hfill&&\hfill\mbox{19.42}\hfill&&\hfill\mbox{23.10}\hfill&\IEEEeqnarraystrutsizeadd{0pt}{2pt}\\
&&&&\hfill\mbox{LSSC}\hfill&&\hfill\mbox{\textbf{25.96}}\hfill&&\hfill\mbox{21.82}\hfill&&\hfill\mbox{\textbf{23.56}}\hfill&&\hfill\mbox{\textbf{23.94}}\hfill&&\hfill\mbox{\textbf{23.14}}\hfill&&\hfill\mbox{\textbf{23.34}}\hfill&&\hfill\mbox{21.18}\hfill&&\hfill\mbox{24.30}\hfill&&\hfill\mbox{25.63}\hfill&&\hfill\mbox{\textbf{24.00}}\hfill&&\hfill\mbox{\textbf{23.14}}\hfill&&\hfill\mbox{\textbf{19.50}}\hfill&&\hfill\mbox{\textbf{23.29}}\hfill&\IEEEeqnarraystrutsizeadd{0pt}{2pt}\\
&&&&\hfill\mbox{EPLL}\hfill&&\hfill\mbox{25.30}\hfill&&\hfill\mbox{22.04}\hfill&&\hfill\mbox{22.10}\hfill&&\hfill\mbox{23.78}\hfill&&\hfill\mbox{22.87}\hfill&&\hfill\mbox{\textbf{23.34}}\hfill&&\hfill\mbox{19.80}\hfill&&\hfill\mbox{\textbf{24.37}}\hfill&&\hfill\mbox{25.44}\hfill&&\hfill\mbox{23.96}\hfill&&\hfill\mbox{22.93}\hfill&&\hfill\mbox{18.95}\hfill&&\hfill\mbox{22.91}\hfill&\IEEEeqnarraystrutsizeadd{0pt}{2pt}\\
&&&&\hfill\mbox{NCSR}\hfill&&\hfill\mbox{25.66}\hfill&&\hfill\mbox{22.05}\hfill&&\hfill\mbox{23.30}\hfill&&\hfill\mbox{23.64}\hfill&&\hfill\mbox{22.89}\hfill&&\hfill\mbox{23.22}\hfill&&\hfill\mbox{\textbf{21.29}}\hfill&&\hfill\mbox{24.13}\hfill&&\hfill\mbox{\textbf{25.65}}\hfill&&\hfill\mbox{23.97}\hfill&&\hfill\mbox{22.64}\hfill&&\hfill\mbox{19.23}\hfill&&\hfill\mbox{23.14}\hfill&\IEEEeqnarraystrutsizeadd{0pt}{2pt}\\
&&&&\hfill\mbox{NCSR-AGNN}\hfill&&\hfill\mbox{25.65}\hfill&&\hfill\mbox{22.09}\hfill&&\hfill\mbox{23.20}\hfill&&\hfill\mbox{23.53}\hfill&&\hfill\mbox{22.87}\hfill&&\hfill\mbox{23.20}\hfill&&\hfill\mbox{21.19}\hfill&&\hfill\mbox{24.10}\hfill&&\hfill\mbox{25.62}\hfill&&\hfill\mbox{23.95}\hfill&&\hfill\mbox{22.64}\hfill&&\hfill\mbox{19.27}\hfill&&\hfill\mbox{23.11}\hfill&\IEEEeqnarraystrutsizeadd{0pt}{2pt}\\
\IEEEeqnarraydblrulerowcut\\
\end{IEEEeqnarraybox}
\end{table*}

\section{Conclusion}
\label{sec:conclusion}

In this paper, we have focused on the problem of selecting local subsets of training data samples that can be used for learning local models for image reconstruction. This study has been motivated by the observation that the Euclidean distance may not always be a good dissimilarity measure for comparing data samples lying on a manifold. We have proposed two methods for such data subset selection which take into account the geometry of the data assumed to lie on a manifold.  Although the addressed problem has close links with manifold clustering, it differs by the fact that the goal here is not to obtain a partitioning of data, but instead select a local subset of training data that can be used for learning a good model for sparse reconstruction of a given input test sample. 
%Methods have also been described for setting the parameters of the proposed algorithms based on the local manifold geometry. 
The performance of the methods has been demonstrated in a super-resolution application leading to a novel single-image super-resolution algorithm which outperforms reference methods, as well as in deblurring and denoising applications.

% or


%\appendices
%\section{Proof of the First Theorem}

%\section{}

\section*{Acknowledgment}
The first author would like to thank CAPES Brazilian agency for the financial support (PDSE scholarship 18385-12-5). The authors also thank J\'er\'emy Aghaei Mazaheri for the helpful discussions.

\ifCLASSOPTIONcaptionsoff
  \newpage
\fi



\bibliographystyle{IEEEtran}
%\bibliography{IEEEabrv,C:/Users/juferrei/Pesquisas/Doutorado/Tese/library/library}
\bibliography{IEEEabrv,library}

\end{document}